\documentclass[final,5p,times,twocolumn]{elsarticle}

 \usepackage{afterpage}

\usepackage{amssymb}
\usepackage{url}
\usepackage{acro}
\usepackage{pdflscape,booktabs}

\usepackage{lipsum}
\usepackage{array}
\usepackage{geometry}
\usepackage{multirow}
\usepackage{array}
\newcolumntype{C}[1]{>{\centering\arraybackslash}p{#1}}

\usepackage{amssymb}
\usepackage{booktabs}
\usepackage{xcolor}

\usepackage{soul}
\usepackage{minitoc}

\usepackage{hyperref}
\usepackage{xcolor}
\definecolor{medium-blue}{rgb}{0,0,1}
\hypersetup{colorlinks, urlcolor={medium-blue}}
\usepackage{todonotes}

\usepackage{pifont}

\DeclareUnicodeCharacter{3B1}{\ensuremath{\alpha}}

\journal{Artificial Intelligence Review}

\usepackage{graphicx}
\usepackage{caption}

\DeclareCaptionFormat{myformat}{\bfseries #1 \mdseries#3}
\begin{document}

\begin{frontmatter}

\title{Generative AI for Synthetic Data Across Multiple Medical Modalities: A Systematic Review of Recent Developments and Challenges }







            
            

\author[first,second,third]{Mahmoud Ibrahim \textsuperscript{*}}       
\ead{mahmoud.ibrahim@vito.be}  
\cortext[cor1]{Corresponding author}  

\author[fourth]{Yasmina Al Khalil 
}
\author[fourth]{Sina Amirrajab
}

\author[first,second]{Chang Sun}
\author[fourth]{Marcel Breeuwer}
\author[fourth]{Josien Pluim}
\author[third]{Bart Elen}
\author[third]{Gökhan Ertaylan}

\author[first,second]{Michel Dumontier}

\affiliation[first]{organization={Institute of Data Science, Faculty of Science and Engineering, Maastricht University},
            city={Maastricht},
            country={The Netherlands}}

\affiliation[second]{organization={Department of Advanced Computing Sciences, Faculty of Science and Engineering, Maastricht University},
            city={Maastricht},
            country={The Netherlands}}
            
\affiliation[third]{organization={VITO},
        country={Belgium}}
        
\affiliation[fourth]{organization={Department of Biomedical Engineering, Eindhoven University of Technology
 },
            city={Eindhoven},
            country={The Netherlands}}


\begin{abstract}

This paper presents a comprehensive systematic review of generative models (GANs, VAEs, DMs, and LLMs) used to synthesize various medical data types, including imaging (dermoscopic, mammographic, ultrasound, CT, MRI, and X-ray), text, time-series, and tabular data (EHR). Unlike previous narrowly focused reviews, our study encompasses a broad array of medical data modalities and explores various generative models. Our search strategy queries databases such as Scopus, PubMed, and ArXiv, focusing on recent works from January 2021 to November 2023, excluding reviews and perspectives. This period emphasizes recent advancements beyond GANs, which have been extensively covered previously.

The survey reveals insights from three key aspects: (1) Synthesis applications and purpose of synthesis, (2) generation techniques, and (3) evaluation methods. It highlights clinically valid synthesis applications, demonstrating the potential of synthetic data to tackle diverse clinical requirements. While conditional models incorporating class labels, segmentation masks and image translations are prevalent, there is a gap in utilizing prior clinical knowledge and patient-specific context, suggesting a need for more personalized synthesis approaches and emphasizing the importance of tailoring generative approaches to the unique characteristics of medical data. Additionally, there is a significant gap in using synthetic data beyond augmentation, such as for validation and evaluation of downstream medical AI models. The survey uncovers that the lack of standardized evaluation methodologies tailored to medical images is a barrier to clinical application, underscoring the need for in-depth evaluation approaches, benchmarking, and comparative studies to promote openness and collaboration.

\end{abstract}


\end{frontmatter}

\onecolumn
\tableofcontents
\twocolumn

\section{Introduction}




The advent of Artificial Intelligence (AI) and Machine Learning (ML) has revolutionized numerous fields, including healthcare. These technologies have the potential to transform medical research and clinical practice, offering new avenues for diagnosis, treatment, and patient care. However, the application of AI and ML in healthcare depends on the availability of large and high-quality datasets, with diverse modalities and acquisition properties. In many instances, such datasets are not readily available due to  privacy concerns, restricted sharing policies, complex acquisition techniques, expensive annotation costs, as well as limited diversity in real world data. This has led to the emergence of synthetic data generation, a promising solution that leverages generative models to create artificial data that mimics real-world datasets.

Synthetic data could serve various critical purposes in data science and machine learning, mainly facilitating data sharing while protecting privacy, augmenting existing datasets, and promoting fairness and equity in AI applications \cite{rajotteSyntheticDataEnabler2022}.
Generative models are computational algorithms capable of learning and capturing complex data distributions, enabling the generation of new samples that closely resemble real data. By leveraging techniques such as Generative Adversarial Networks (GANs), Variational Autoencoders (VAEs), Diffusion Models (DMs), and Large Language Models (LLMs), researchers can create synthetic medical data across various modalities, including imaging, text, time-series, and tabular data. 



Our survey aims to provide a holistic understanding of the applications of these generative models in generating medical synthetic data. We delve into three key aspects: \textit{the purpose of synthesis, generation techniques, and evaluation methods}. We highlight the potential of synthetic data in addressing various clinical needs and identify gaps in current practices, such as the need for more personalized synthesis approaches and standardized evaluation methodologies. Moreover, we emphasize the importance of tailoring generative approaches to the unique characteristics of medical data and call for more in-depth evaluation approaches relevant to clinical applications. Our study encourages benchmarking and comparative studies to promote openness and collaboration in this field.

In essence, this survey paper serves as a valuable resource for researchers and practitioners interested in leveraging generative models for synthesizing medical data. By shedding light on the current practices, potential, and challenges in the field of synthetic medical data generation, we hope to spur further research and innovation in this critical area of healthcare AI and ML.

The structure of this paper is further organized into different chapters. Section \ref{sec-syndata_generation} provides an overview of the synthesis applications, generative models, and evaluation methods that are common across the different data types. The concepts explained in this section are essential to understand the detailed results in section \ref{sec:results}, where the findings from the surveyed papers are presented and divided into four sections, each focusing on a specific type of medical data: Electronic Health Records (EHR) in section \ref{sec:ehr}, physiological signals in section \ref{sec:signals}, medical images in section \ref{sec:images}, and medical text in section \ref{sec:text}. This organization allows for a detailed exploration of the use of generative models for each data type, providing readers with a comprehensive understanding of the current state of synthetic medical data generation. Section \ref{sec:discussion} reveal the insights and conclusions collected from the surveyed papers. Section \ref{sec:Recommendations} provides recommendations to be taken into account for future research, and \ref{sec:conclsuion} concludes the paper. The reader is referred to the table of contents for a smooth navigation of the paper. 

\subsection{Related work}



\begin{figure}
    \centering
    \includegraphics[width=1\linewidth]{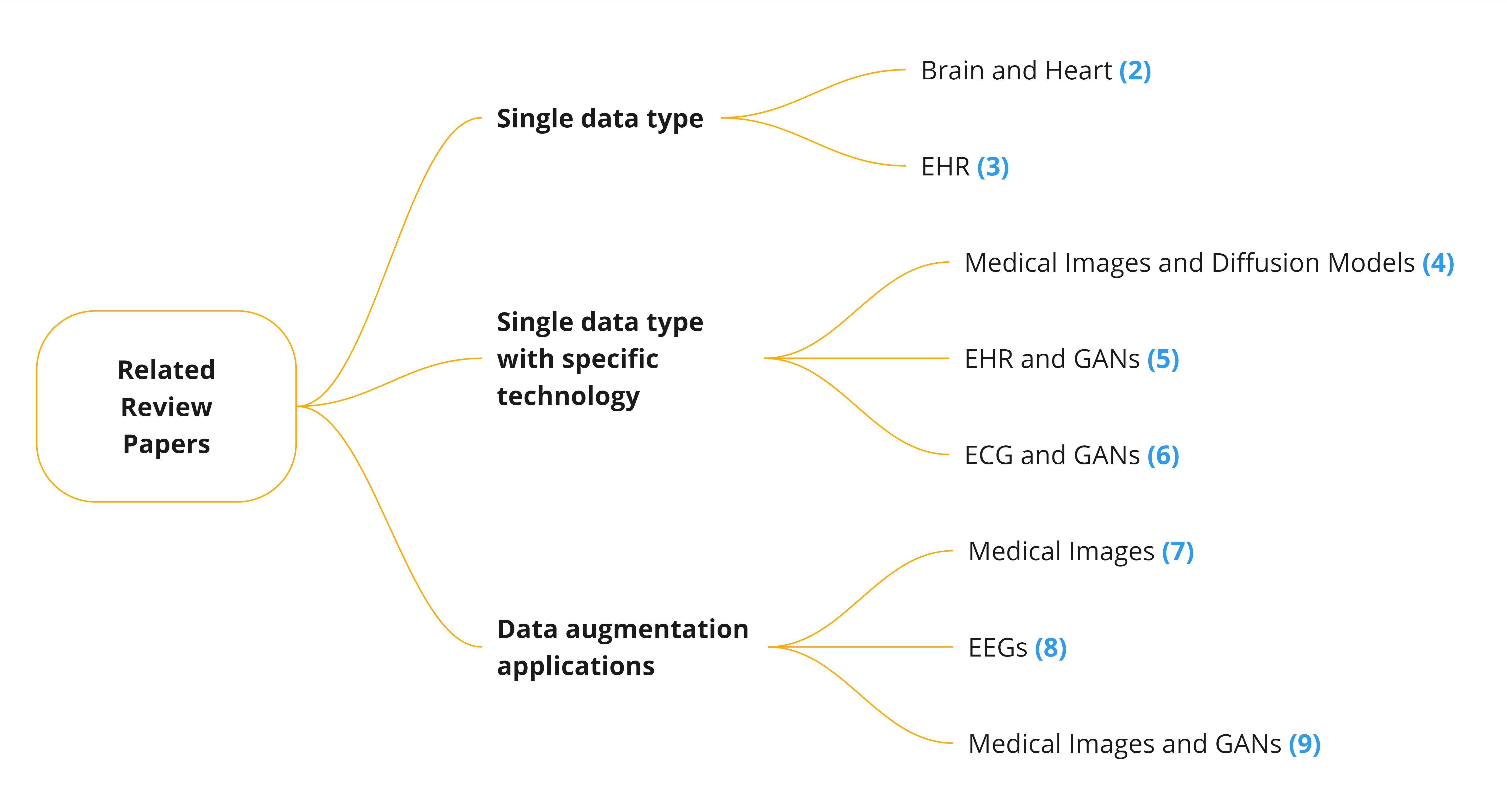}
    \caption{Specialization trends in related survey papers- The papers are either focused on a single data type, or combine a single data type with one specific technology, or are exclusively dedicated to data augmentation applications.}

    \label{fig:related-paper}
\end{figure}
Despite the growing interest in generative models for medical data synthesis, the review of existing survey papers in the field reveals a notable trend: a tendency towards narrow specialization, as seen in Fig.\ref{fig:related-paper}. These papers often focus on a single data type, such as brain and heart imaging \cite{s4} or EHR \cite{s8}, or they combine a single data type with one specific technology, like medical images with diffusion models \cite{s2} , EHR with GANs \cite{s6}, or Electrocardiography (ECG) with GANs \cite{s10}. Additionally, there is a subset of papers dedicated exclusively to data augmentation applications, which similarly specialize in either a single data type \cite{s1,s-11} or a combination of one modality and one technology \cite{s3}.

This pattern indicates a gap in the literature where comprehensive, multi-faceted analyses providing holistic overviews are less prevalent. In contrast, the proposed survey paper seeks to address this gap by covering a broader spectrum. It spans multiple data types, encompassing medical imaging, tabular EHRs, physiological signals, and clinical text notes. Furthermore, it explores various generative models, extending beyond GANs which have been extensively covered in literature, to include diffusion models and language models. The survey also emphasizes the aspect of conditional generation, which is not focused on in similar work. Additionally, we have considered a specific timeline for our review, consciously excluding older papers that have already been covered in various surveys. This approach aims to provide a more holistic understanding of the application of generative models in medical research, moving away from the trend of focusing on singular aspects. Such a comprehensive analysis could significantly enrich the medical field by offering insights that are potentially overlooked by more narrowly focused studies.

\subsection{Review methods and protocol}

Our review methodology and protocol aimed to address several key objectives and research questions. Firstly, we sought to explore the applications and purpose of synthesis beyond data augmentation in medical research. Additionally, we aimed to identify the latest generative models utilized for generating synthetic medical data. Lastly, we aimed to investigate the common protocols for evaluating synthetic data and the trade-offs between the different evaluation dimensions.

Our review is scoped to cover a broad spectrum of modalities, including tabular data (specifically EHRs), physiological signals (primarily ECGs and Electroencephalography (EEGs)), clinical text notes, and a variety of medical images such as dermatoscopic images, mammographic images, Utrasound (US) scans, Computed Tomography (CT) scans, Magnetic Resonance Imaging (MRI) scans, Optical Coherence Tomography (OCT) scans, and X-rays. We consider an array of generative models, including GANs, VAEs, DMs, and LLMs.

Our search strategy involves querying several databases, namely Scopus, PubMed, and ArXiv. We repeated our search query for each modality, using terms related to “synthetic data generation,” “conditional generation,” “generative models,” and other related keywords. To ensure the relevance and timeliness of our findings, we limit our search to papers and preprints published within the time frame of January 2021 to November 2023. Furthermore, we exclude review papers from our search to maintain our focus on original research and primary studies. This approach ensures a comprehensive and up-to-date review of the field.





\subsection{Search results}

\begin{figure*}
    \centering
    \includegraphics[width=0.7\linewidth]{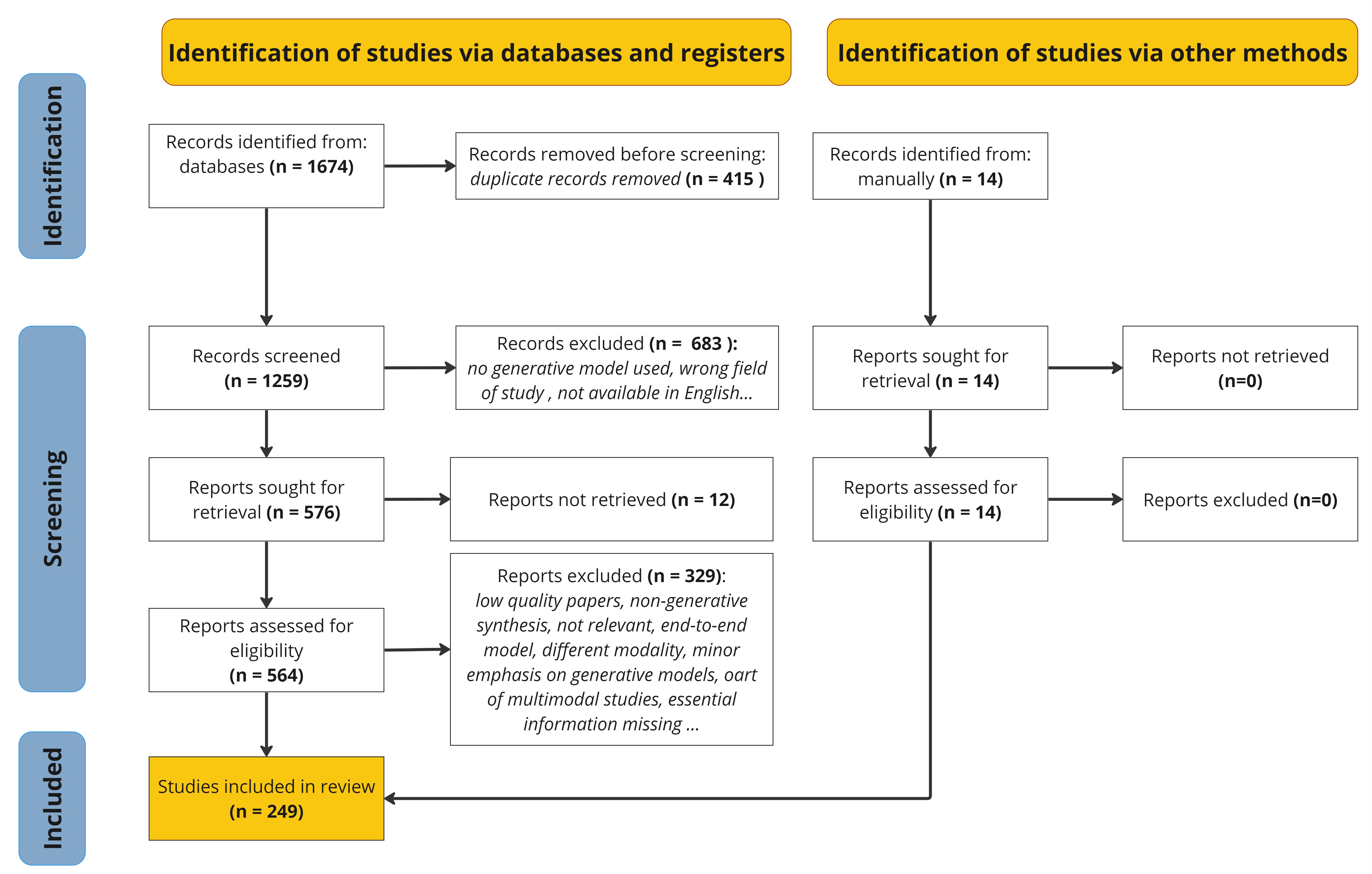}
    \caption{PRISMA flowchart of the literature screening process}
    \label{fig:literature}
\end{figure*}

\begin{figure}
    \centering
    \includegraphics[width=0.9\linewidth]{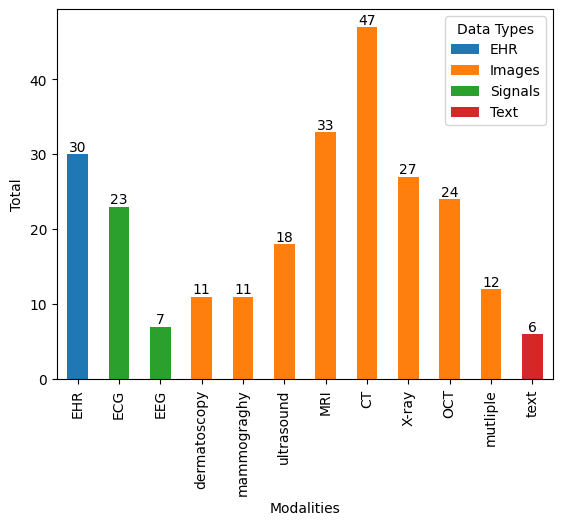}
    \caption{An overview of the number of surveyed papers per modality}
    \label{fig:results}
\end{figure}

The details of the literature screening processes according to the Preferred Reporting Items for Systematic reviews and Meta-Analyses (PRISMA) guidelines \cite{2021prisma} are shown in Fig.\ref{fig:literature}. Overall, 249 papers are included in the review, distributed over the different modalities as shown in Fig.\ref{fig:results}.

\section{Synthesis application, generative technologies and evaluations}
\label{sec-syndata_generation}
This section presents a strong foundational understanding of key concepts gathered from various surveyed papers concerning the overarching themes of (i) synthesis applications, (ii) generative technologies, and (iii) evaluation methodologies in medical data synthesis. The objective is to lay the groundwork by explaining fundamental concepts and approaches that are commonly encountered in the survey. This aids the reader in building the necessary background to understand the detailed and more specific findings presented in section \ref{sec:results}.  
First, we discusses different synthesis applications in section \ref{subsec-Synthesis Applications}, followed by an exploration of common generative models \ref{subsec-Generation}. Finally, in section \ref{subsec-evaluation}, we delve into the diverse evaluation methods employed in the surveyed studies, offering a comprehensive view of the research landscape in this field.

In our paper, we introduce specific definitions for the terms "data types" and "modalities". We classify EHR, images, text, and signals as distinct data types. Within each data type, we define various modalities. For example, within medical images, modalities include CT, MRI, and others. Additionally, within each modality, there exist different modalities; for example, within MRI, modalities such as T1 and T2 weights are distinguished.

\subsection{Synthesis applications and purpose of synthesis} \label{subsec-Synthesis Applications}

\begin{figure*}[h!]
    \centering
    \includegraphics[width=\linewidth]{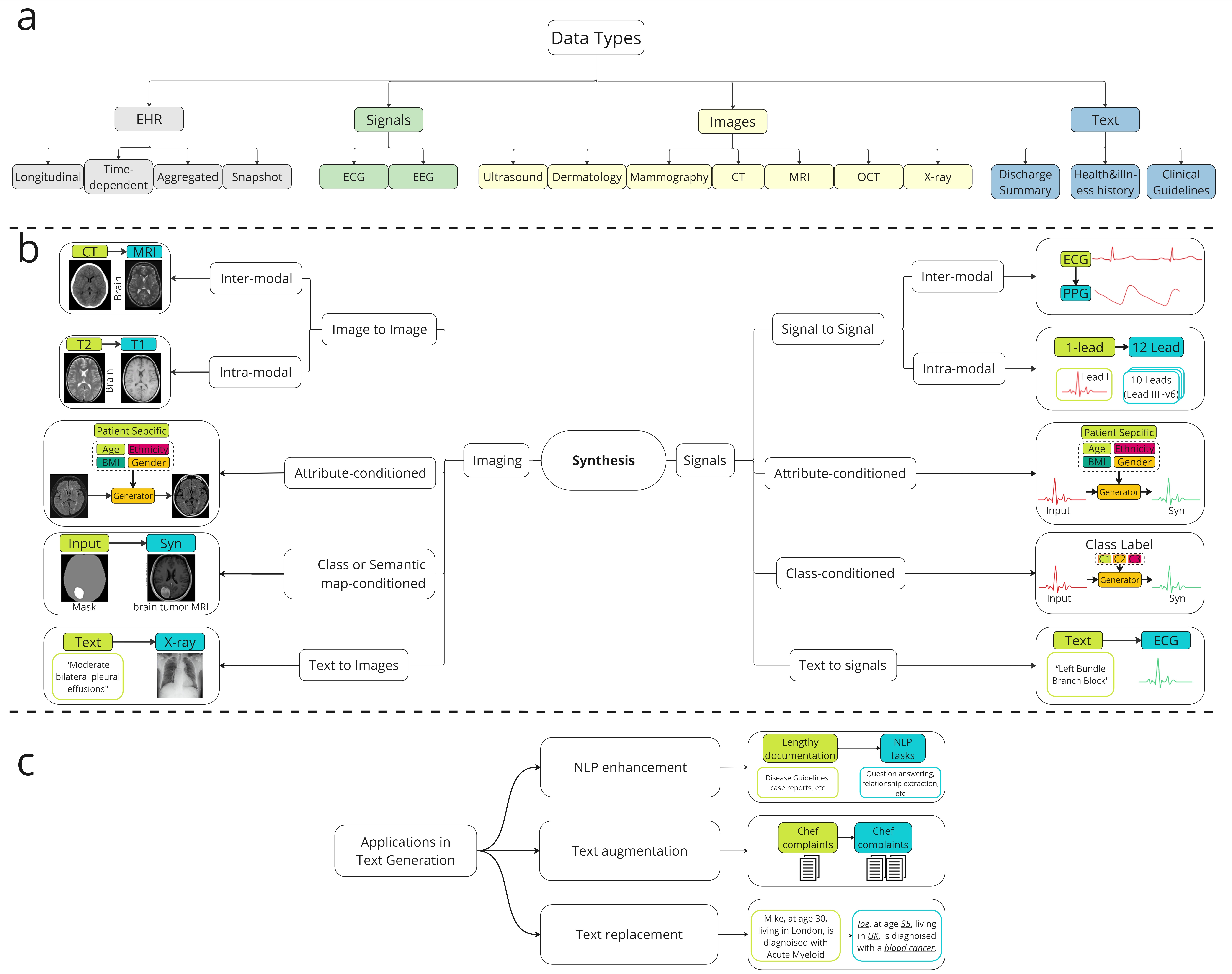}
    \caption{(a) The different data types and modalities covered in the survey. The different EHR formats also represent the synthesis applications of EHR.  (b) Overview of synthesis applications of the imaging and signals data types. The left side focuses on imaging data, while the right side focuses on signals. An example of inter-modal synthesis involves the transformation of a brain CT scan to an MRI in imaging or an ECG signal into a synthetic PPG signal in the signals data type, which helps in scenarios where certain imaging or signal modalities might be unavailable. Intra-modal synthesis involves the conversion of a brain MRI T2 sequence into a T1 sequence or transforming a single lead into a 10 lead synthetic ECG signal. Attribute-conditioned synthesis shows a patient-specific brain MRI or an ECG signal being generated that matches attributes like age, BMI, ethnicity, and gender. Class or semantic map-conditioned synthesis shows a synthetic brain MRI with tumor being generated using a binary mask of a brain tumor or a synthetic ECG signal being generated based on class labels, such as C1 and C2. This can be useful for generating labeled datasets for training. The figure also illustrates text-guided synthesis, such as generating a synthetic chest X-ray based on a textual description like "moderate bilateral pleural effusions" or a synthetic ECG signal based on textual descriptions like "Left Bundle Branch Block". (c) Synthesis applications of the text data type}
    \label{fig:synthesis-applications}
\end{figure*}

\begin{figure}[h!]
    \centering
    \includegraphics[width=\columnwidth]{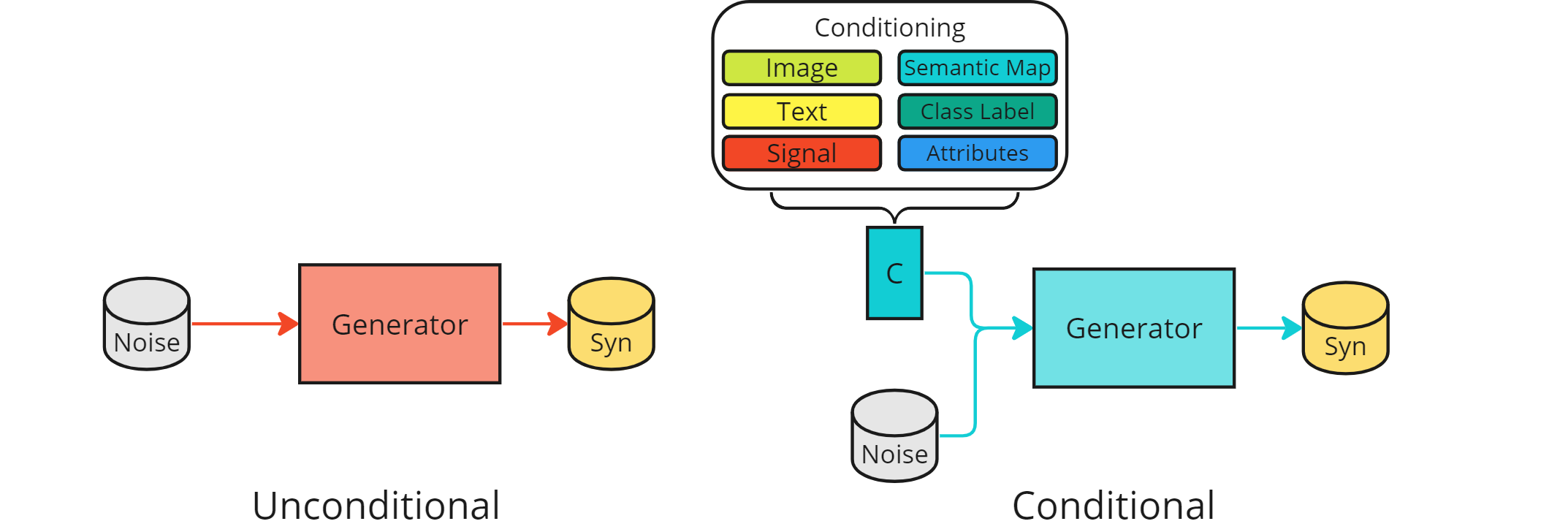}
    \caption{Unconditional (left) vs conditional generation (right)}
    \label{fig:cond}
\end{figure}

    Generative models are a powerful tool in machine learning and can be broadly categorized into two types: unconditional and conditional. Unconditional models take a random variable as input, allowing for the application of unconditional synthesis. On the other hand, conditional generative models introduce an additional layer of control by incorporating external information or context during the generation process that serve as additional guidance for the model. These can include images, text, semantic maps, class labels, attributes, and signals, as demonstrated in Fig.\ref{fig:cond}.  This added control allows conditional models to be used in a variety of synthesis applications specific to each data type shown in Fig.\ref{fig:synthesis-applications}.

    For the EHR data type, the synthesis depends on the EHR format, which includes (i) {Longitudinal EHR}, consisting of medical codes from various patient visits; (ii) {Aggregated EHR}, which consists of longitudinal visits of the patients condensed into a single row; (iii) {Time-dependent EHR}, which consists of time-series readings from one patient's visit; and (iv) {Snapshot EHR} consisting a single snapshot of a patient's EHR focusing on selected attributes.
    
    Regarding the physiological signals and imaging data types, prominent applications include (i) {Inter-modal translation} which involves converting data from one modality to another. For example, in the realm of images, this could involve translating CT scans to MR images. In the signals data type, this could mean converting ECG readings to PPG signals.
    (ii) {Intra-modal translation} which involves translating data within the same modality. For instance, in images, this could mean translating across different MRI contrasts: T1-weighted images to T2-weighted images. In signals , this could involve converting single-lead ECG inputs into a complete 12-lead set.
    (iii) {Class or semantic map-based synthesis} that involves generating data conditioned on a certain class label or a certain segmentation mask.
    (iv) {Attribute-based Synthesis} where the generated data is conditioned to correspond with specific subject characteristics and patient demographics, such as age, sex, and Body Mass Index (BMI). These approaches suggest a higher level of personalization in the synthetic data, potentially leading to more accurate representations of patient-specific medical data.
    (v) {Text-based synthesis} demonstrates innovative approach for integrating descriptive text into synthetic medical data generation by conditioning on clinical text reports for example. This could involve text to image, text to signal, or text to table synthesis.

    For the medical text synthesis, we categorize the applications based on their generation purposes into (i) {Natural Language Processing (NLP) enhancement}, where the generated synthetic text data is used to improve the performance of NLP tasks.  A typical NLP task in medical text generation includes name entity recognition, information / concept extraction, relation extraction, semantic similarity, summarization, and question answering. For example, the Name Entity Recognition (NER) performance on real clinical notes are limited by the insufficient amount and completeness of the dataset. By adding generated synthetic medical text can improve the NER task in this scenario; (ii) {text augmentation}, where the generated text is used to augment the existing text such as generating discharge summaries, patients reports, clinical notes and case reports of certain diseases especially when the real-world clinical notes are limited and sensitive to share or use for research; (iii) {text de-identification} by replacing personal identifiable and sensitive information about individuals in order to preserve patients privacy in the clinical text. These applications have a focus on particular attributes (such as names, address, or diagnosis) in the text .

\subsection{Generation techniques}  \label{subsec-Generation}
\begin{figure}[h!]
    \centering
    \includegraphics[width=\columnwidth]{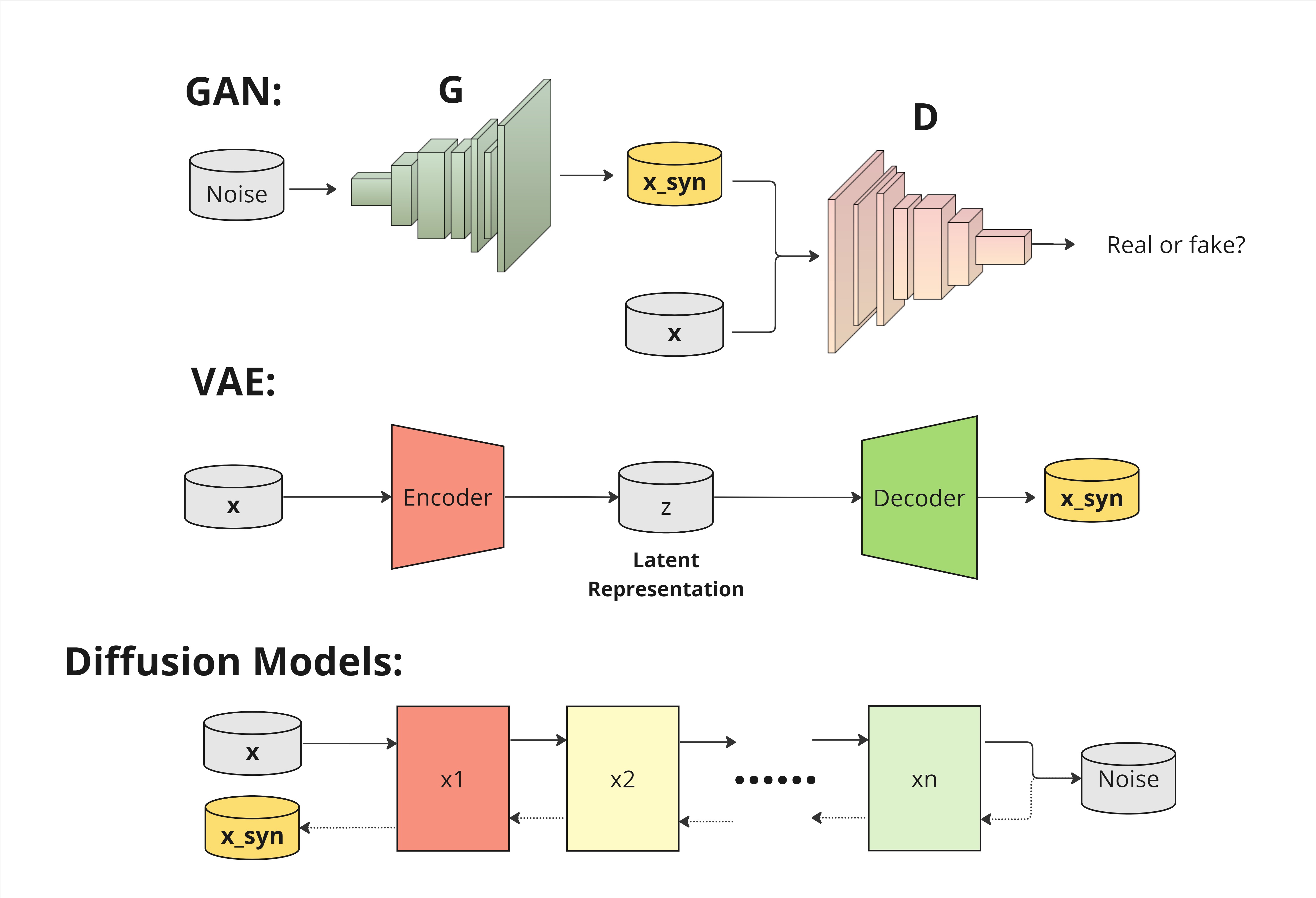}
    \caption{Generative Models. GANs follow adversarial training to implicitly model the real distribution. VAEs employs variational inference techniques \cite{vi,vi2} to approximate real distribution, while diffusion models gradually add noise to the input and reverse it back}
    \label{fig:generative-models}
\end{figure}

Our examination of various papers revealed a range of techniques utilized for generating synthetic medical data, with a primary focus on GANs \cite{goodfellow2014generative}, VAEs \cite{vae} and recent advancements in diffusion models \cite{diffusion} and language models. See Fig.\ref{fig:generative-models} for the basic working mechanisms of the different generative models.

Firstly, we discuss the differences in how various generative models operate across different datatypes: EHR, medical images, medical text, and physiological signals.
Next, we highlight notable state-of-the-art GANs frequently referenced in the literature.
Following this, we explore diffusion models, including denoising diffusion probabilistic models and latent diffusion models, alongside discussions on different text embedding methods.
Finally, we examine language models, concluding our overview of the diverse techniques employed in generating synthetic medical data.

\subsubsection{Differences in generative models working mechanism per datatype}

\begin{itemize}
\item  \textbf{Handling discrete variables:} GANs, initially introduced for generating realistic continuous images, face challenges when applied to discrete data. Similarly, many diffusion models are Gaussian processes, operating in continuous spaces and not the discrete space. To address these issues, various techniques have been introduced. (Variational) autoencoders are commonly employed before GANs and diffusion models to condense high-dimensional and heterogeneous features into latent representations. Notably, some diffusion models like Tabular Denoising Diffusion Probabilistic Models (tabDDPM) \cite{tabddpm} adopt a mixed sequence diffusion approach \cite{ehr-58,s-12}, employing Gaussian diffusion for continuous variables and multinomial diffusion \cite{multinomialdiffusion} for discrete variables. Others treat discrete variables similar to real-valued sequences but with further post-processing of the model output, transforming continuous variables into discrete.
\item  \textbf{Handling time aspect:} Notably, conventional vanilla GANs and diffusion models lack the ability to generate time-series data, making them unsuitable for directly creating time-dependent EHR data or signals. As a remedy, specialized sequential Deep Learning (DL) models like Recurrent Neural Networks (RNNs), Gated Recurrent Units (GRUs), and Long Short-term Memory(LSTMs) are employed within the architectures of the GANs or diffusion models.
\end{itemize}

\subsubsection{GANs} 

\begin{figure*}[h!]
    \centering
    \includegraphics[width=1\linewidth]{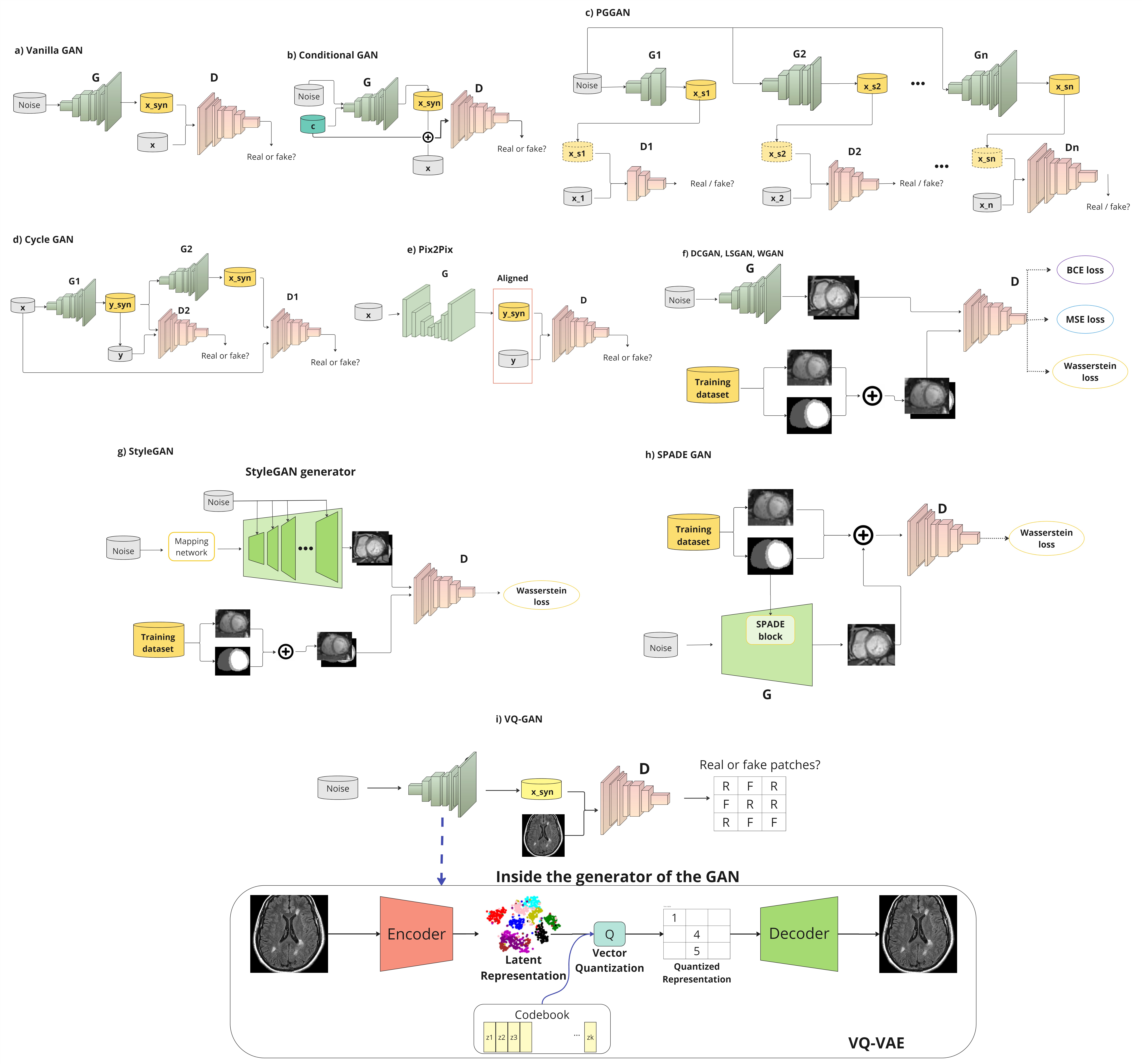}
    \caption{Comprehensive overview of various GAN architectures, each tailored for specific applications and improvements in synthetic data generation. The {Vanilla GAN} (a) illustrates the basic GAN structure with a generator creating samples from noise and a discriminator evaluating them. The {Conditional GAN} (b) integrates conditional variables to guide the generation process, enabling the production of more targeted outputs. The {PGGAN} (c) employs a series of generator and discriminator pairs (G1, D1 to Gn, Dn) that progressively increase the resolution and detail of the generated samples. {CycleGAN} (d) features a dual-generator system for effective style transfer, where each generator learns to translate images between two distinct domains, and the outputs are cycled back as inputs to maintain consistency. The {Pix2Pix} (e) uses aligned pairs of images for precise image-to-image translation tasks, converting segmented images to realistic photos using a trained generator. The Generator G is usually an encoder-decoder net or U-Net with skip connection.  {DCGAN}, {LSGAN}, and {WGAN} (f), use different loss functions like BCE, MSE, and Wasserstein to refine the discriminator's accuracy. {StyleGAN} (g) features a sophisticated generator with a mapping network that inputs noise, refines the generation through style blocks, and uses a discriminator with Wasserstein loss. {SPADEGAN} (h ) utilizes spatially-adaptive normalization to modulate the synthesis based on the input segmentation map, with a discriminator employing Wasserstein loss. {VQ-GAN} (i) merges GAN and VAE elements by encoding inputs into a latent space, quantizing them for efficient representation, and then reconstructing the output to improve image quality. {x} and {x\_syn}, as well as {y} and {y\_syn} denote input and synthesized data pairs,  while {c} represents the conditional variable. Similarly, {x\_1} and {x\_s1}, {x\_2} and {x\_s2}, {x\_n} and {x\_sn} represent real and synthesized data pairs. }
    
    \label{fig:gans}
\end{figure*}


Adversarial approaches, such as GANs, involve a generator and discriminator, trained to outperform each other, hence the term "adversarial". The generator network adapts its distribution and generates a new, reliable distribution, while the discriminator network learns to differentiate between real and augmented data. The work in \cite{goodfellow2014generative} pioneered GANs for generating realistic images, particularly dealing with continuous data, often referred to as the {vanilla GAN} (Fig.\ref{fig:gans}a). However, such models have been extended to different domains, such as audio and EEG signal generation \cite{donahue2018synthesizing, raoof2021study}. \cite{radford2015unsupervised} proposed the {Deep Convolutional GAN (DCGAN)}, that combines the original GAN with convolutional neural networks for better and more stable training. \cite{arjovsky2017wasserstein} proposed the {Wasserstein GAN (WGAN)}, while \cite{mao2017least} proposed the {Least Squares GAN (LSGAN)}, to address challenges like mode collapse and vanishing gradients typically encountered in GANs (see Fig.\ref{fig:gans}f). WGAN replaces binary classification with the Wasserstein distance. The training of WGAN was enhanced with the introduction of WGAN-GP in \cite{gulrajani2017improved}, which integrates a gradient penalty technique. {Progressive Growing GAN (PGGAN)}\cite{karras2017progressive} is an extension of GAN training, ensuring stability for generators producing large, high-quality images. Traditional GANs struggle with stability when working with larger sizes, attempting to balance both structure and fine details. This challenge worsens as resolutions increase, often required for medical image generation, leading to training failures, while memory constraints on Graphics Processing Units (GPUs) further necessitate reducing batch size, introducing instability. As illustrated in Fig.\ref{fig:gans}c, PGGAN addresses these issues by incrementally increasing model size during training, starting small and gradually adding layers until achieving the desired image size. 

GANs enable the generation of diverse image, video, or audio data from random input sampled from a normal distribution \cite{alqahtani2021applications}. However, vanilla, unconditional GANs lack control over the generated samples' appearance or class. To address this, {Conditional Generative Adversarial Networks (cGANs)} are introduced, which allow for conditional generation by incorporating semantic input or a condition (c), like image class, into the generation process \cite{mirza2014conditional}, as shown in Fig.\ref{fig:gans}b. Both the generator and discriminator in cGANs are conditioned on auxiliary information, enabling the model to learn complex mappings from diverse contextual inputs to corresponding outputs.

Conditional GANs have popularized the use of image conditioning in image-to-image translation tasks. For such tasks, models like {Pix2Pix} \cite{pix2pix} (see Fig.\ref{fig:gans}e) and its high-resolution counterpart {Pix2PixHD} \cite{pix2pixHD} rely on supervised training with paired datasets. Pix2PixHD addresses some limitations of Pix2Pix, specifically catering to high-resolution image translation. {CycleGAN} \cite{cyclegan} offers an alternative by excelling in unsupervised image translation, eliminating the need for paired datasets. CycleGAN utilizes cycle consistency, allowing generated images from one generator to serve as input for the other, with the output matching the original image, enabling consistency in both directions, as shown in Fig.\ref{fig:gans}d. This capability proves valuable in scenarios where obtaining paired training data is challenging. While CycleGAN shines in unsupervised learning, it is not optimal for high-resolution image-to-image translation. Consequently, paired approaches are often preferred for medical data generation \cite{wang2021review}.

Other notable GAN-based approaches frequently employed for medical data generation, include {StyleGAN} \cite{karras2019style}, {SPADE GAN} \cite{park2019semantic}, and {VQ-GAN} \cite{vqgan}. Style Generative Adversarial Network (StyleGAN) \cite{karras2019style}, depicted in Fig.\ref{fig:gans}g, represents a significant extension to the GAN architecture. It introduces changes like a mapping network for latent space, allowing control over style at various points in the generator model, and the incorporation of noise for variation. The StyleGAN generator and discriminator models are trained using the PGGAN method. This model not only produces high-quality, high-resolution images but also provides control over style at different detail levels through manipulation of style vectors and noise. A cutting-edge image translation model, akin to Pix2Pix, is SPADE GAN \cite{park2019semantic}, which introduces spatially-adaptive (SPADE) normalization. Unlike previous models, which employed segmentation maps only at the input layer, SPADE GAN ensures the segmentation map is inputted across all intermediate layers, preserving its information throughout the model's depth. (see Fig.\ref{fig:gans}h).

The VAE introduces a framework for compressing data into a lower-dimensional space \cite{vae}. The Vector-Quantized Variational Autoencoder (VQ-VAE) model enhances representation by mapping inputs to a continuous space and then discretizing these into tokens, facilitated by an encoder, decoder, and a learnable codebook \cite{vqvae}. The Vector-Quantized Generative Adversarial Network (VQ-GAN) fuses VAE and GAN models with vector quantization (VQ) to create data, including images, from noise, and employs VQ to elevate the quality by turning outputs into discrete values. This technique not only improves the definition and edges of generated data but also provides examples like images with more precise and clearer contours compared to outputs from conventional GANs, benefiting from the combined strengths of GANs and VQ's detailed representation \cite{vqgan}.

\begin{figure*}
    \centering
    \includegraphics[width=1\linewidth]{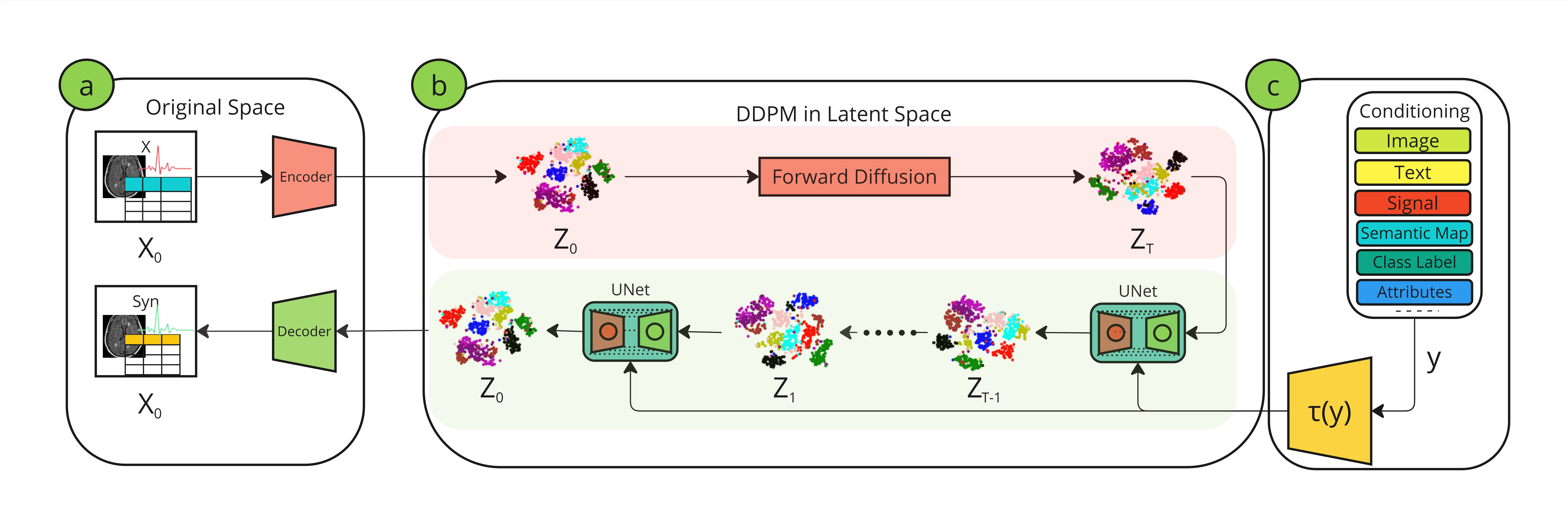}
    \caption{Diffusion Models. \textbf{a,b,c} constitute the components of the Stable diffusion pipeline. \textbf{a}: (Variational)AutoEncoder component for compressing the high-dimensional inputs $X_{0}$ into the lower-dimensional latent representations $Z_{0}$ and mapping them back to the pixel space. $X_{0}$ can be an image, a signal, or a table with heterogeneous features. \textbf{b}: DDPM process consisting of a forward diffusion process and a backward one. $Z_{0}$ is  destructed by adding noise iteratively. $Z_{0}$ is then reconstructed from $Z_{T}$ iteratively using a neural network. Within the pipeline of Latent Diffusion Model, the  DDPM process is performed in the latent space. Originally in the DDPM, the diffusion is performed in the original space. \textbf{c}: Optionally, an embedded conditioning $\tau(y)$ can be added, where y  could be other images, radiological reports, semantic maps, class labels... In SD, the embedding function $\tau$ is based on CLIP}



    \label{fig:diffusion}
\end{figure*}

\subsubsection{Diffusion models}

The original concept of the {diffusion model}, introduced by Diffusion Probabilistic Models (DPMs) \cite{diffusion}, draws inspiration from non-equilibrium statistical physics. The key idea is to systematically and iteratively destroy structure in a data distribution through a forward diffusion process. Subsequently, the reverse diffusion process is learned and applied to restore the structure in the data.

\paragraph{DDPM}The initial practical implementation of the diffusion model in the context of images was presented by \cite{ddpm}, introducing {Denoising Diffusion Probabilistic Models (DDPM)}. This approach destroys data by iteratively adding Gaussian noise to the image according to the Markov chain. To learn the reverse process, a deep neural network is needed to recover the data. Fig.\ref{fig:diffusion}b illustrates the working mechanism of the DDPM. The current best practice for image diffusion models is to use U-Net \cite{unet} architectures for denoising. However, these architectures are tailored for image generation tasks, and may not be a viable option for other data types, especially when the training data is limited. Moreover, U-Net architectures struggle to retain temporal dynamic information and lack flexibility in handling varying input sequence lengths, thus proving inadequate for managing sequential information. In addressing this limitation, some studies \cite{ehr-58} resort to Bidirectional Recurrent Neural Networks (BRNNs) implemented with either LSTM or GRU units.  A significant drawback of diffusion models is that they operate sequentially on the entire image during both training and inference. Consequently, they demand substantial computational resources and time, making both the training to be slow as well generating an image after training. As a result, \cite{ddim} speed up image generation by redefining the diffusion process as a non-Markovian one, which allows for skipping steps in the denoising process, without requiring all past states to be visited before the current state. 

\paragraph{Latent Diffusion Models- The Stable Diffusion Pipeline} To address computational inefficiencies of DDPMs, {Latent Diffusion Model (LDM)} \cite{ldm} was introduced initially for images, which uses encoders to compress images from their original size in the pixel space into a smaller representation in the latent space, as illustrated in Fig.\ref{fig:diffusion}a. Therefore, in latent diffusion, the diffusion process is performed on the latent representations rather than the original images (Fig.\ref{fig:diffusion}b), allowing LDMs to model long-range dependencies within the data and enabling training on limited computational resources while retaining their quality and flexibility. {Stable diffusion}, a foundation model built based on LDM, introduces text conditioning to the model for additional control over the generation process and consists of three main components: (i) the VAE for compression (Fig.\ref{fig:diffusion}a), (ii) a U-net based diffusion process in the latent space (Fig.\ref{fig:diffusion}b), and (iii) a conditioning mechanism that embeds a prompt describing the image using a Contrastive Language-Image Pre-training (CLIP) text encoder \cite{clip} (Fig.\ref{fig:diffusion}c). CLIP creates a numeric representation (embedding vector) of the prompt, mapping both the text and images into the same representational space, enabling comparison and similarity quantification. In addition to CLIP, other text encoders such as pre-trained T5X model \cite{t5x}, medBERT encoder \cite{medbert}, a Byte Pair Encoding (BPE) tokenizer \cite{bpe}, and Self-alignment pretraining for BERT (SAPBERT) \cite{sapbert}  are commonly used in different publications to enable text conditioning of the models.

The Stable Diffusion model is widely utilized in recent publications, where the pre-trained foundation model is fine-tuned on various medical modalities without the need to initiate training from scratch. Moreover, the idea of latent diffusion models, which involves performing the diffusion process in the latent space, has been extended to both the EHR and signals data types. In these extensions, a low-dimensional representation of the high-dimensional features, including discrete ones, is learned using an auto encoder before the diffusion process.

\subsubsection{Language models:} 
\begin{figure}
    \centering
    \includegraphics[width=0.8\columnwidth]{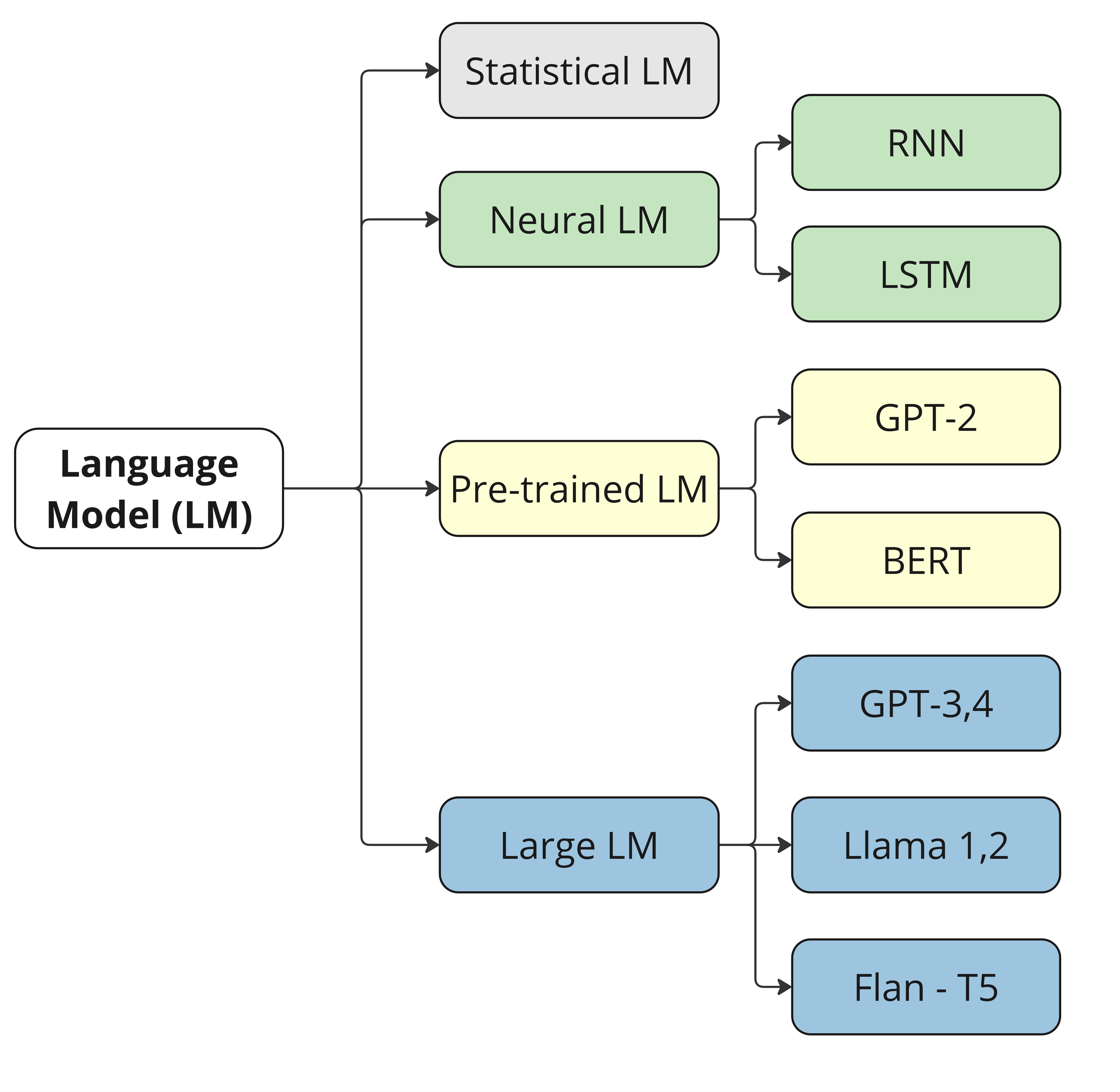}
    \caption{Language models clusters}
    \label{fig:lmcluster}
\end{figure}

A fundamental aspect of language models is enhancing the linguistic capabilities of machines to understand the probability of sequences of words, enabling them to predict future or missing words in sentences \cite{zhao2023survey}. The development of language models (shown in Fig.\ref{fig:lmcluster}) started from Statistical language models which were grounded in probability theories with a restriction to predicting the next word in a sequence based on a fixed number of previous words. However, these models face significant challenges in accurately predicting the next word due to the exponential growth in possible word sequences as the sentence length increases. 

Furthermore, more comprehensive processing methods - Neural Language Models - have been developed utilizing neural networks, such as RNNs and LSTM models. These advanced models excel in recognizing longer-distance relationships (compared to statistical language models) and contextual subtleties in texts, employing word embeddings to grasp semantic similarities among words, thus facilitating the generation of novel word combinations. RNNs and LSTM models have the advantages of creating semantically and syntactically correct content and capturing correlations across sentences. Rather than taking words and text as input, \cite{lee2018natural} generate synthetic chief complaints from EHR data. The generated complaints maintain certain epidemiological details and the relationships between the chief complaints and the discharge diagnosis code. However, both models have limitations in the text length that can be understood and processed and suffer from generating more complex and hierarchical text.

Subsequently, pre-trained language models such as BERT (Bidirectional Encoder Representations from Transformers) \cite{devlin2018bert} and GPT-2 (Generative Pretrained Transformer 2) \cite{radford2019language}, were proposed and garnered significant attention from language processing research due to its outstanding performance and efficiency. BERT is well-known for its architecture design which consists of pre-training deep bidirectional representations from unlabeled text and analyzing text data from both left to right and right to left. BERT is pre-trained on the BooksCorpus and English Wikipedia (3300M words in total). To generate medical text with specific domain knowledge, extended BERT models such as Bio-BERT \cite{lee2020biobert} and Clinical-BERT \cite{alsentzer2019publicly} were developed by training on additional medical corpus such as text from PubMed, clinical reports, doctors' notes. Similar to BERT, GPT-2 applied transformer-based architecture which uses a self-attention mechanism to weigh the importance of different words in single or multiple sentences. GPT-2 is trained on WebText data (40GB text) and focuses on unidirectional (left-to-right) text processing and generation. 

In recent years, large language models such as GPT-4, Llama (Large Language Model Meta AI) \cite{touvron2023llama}, and FLAN-T5 (Fine-tuned LAnguage Net with a Text-To-Text Transfer Transformer) \cite{chung2022scaling}, have significantly enhanced the capabilities of prior language models, showcasing remarkable proficiency across a spectrum of downstream applications. Owing to its high efficiency and scalability, the transformer model has become a foundation for most large language models including GPT-4 and Llama. GPT-4 is the latest updated version of GPT models which can understand larger and more complex text and generate more coherent and contextually relevant text compared to previous GPTs, while Llama (1 and 2) enhanced training efficiency and model scalability and accessibility for broader users. Different from GPT-4 and Llama, FLAN-T5 is based on the Text-to-Text Transfer Transformer model which transforms all the natural language processing tasks to text-to-text problems and trains on various tasks presented in natural language instructions. FLAN-T5 has shown outstanding performance in understanding and executing a set of tasks and has the capability to adapt to varied instructions and tasks.

General language models are designed to deal with a variety of tasks related to natural language understanding and generation such as text generation, translation, question answering, summarization, named entity recognition, text classification, and many others. To tailor them for clinical and medical texts, the language models are trained on medical literature (such as PubMed), patients records, case reports, clinical notes to handle the specialized and complex medical data. The main tasks for medical language models are augmenting clinical text data for downstream tasks (such as disease classification), generating summarization from lengthy clinical documentation, identifying personal health information, and generating anonymized data to replace sensitive personal information, and question answers to support clinical decision making process.

\subsection{Evaluation} \label{subsec-evaluation}
The evaluation of synthetic data is a multifaceted task, encompassing several dimensions that collectively determine the quality and usability of the synthetic data. We categorize the evaluation as utility, fidelity, diversity, qualitative assessment, clinical validation, and privacy. Each dimension presents unique challenges and considerations, and their comprehensive evaluation is crucial to ensure the effectiveness of synthetic data. In this survey, we analyze each of these evaluation approaches, aiming to provide a holistic view of the current synthetic data evaluation practices. The remaining of this section gives an overview of all evaluation categories considered in this paper, offering insights into their significance and the common metrics employed.


\textbf{Evaluation of utility} asses the use of the synthetic data for specific tasks, such as providing additional data for improving a downstream medical AI model. Several settings exist in which the synthetic data can be used for a downstream task. The synthetic data can be used to fully train the downstream model(replacing the need to access real data), or to augment real training data. Moreover, the synthetic data can be used for validating or testing the models. These settings can be employed in different downstream applications such as supervised classification, prediction, supervised regressions, image segmentation, even reinforcement learning, with the specific metrics of each application. 


    

\textbf{Evaluation of privacy} is a critical factor that ensures the synthetic dataset cannot be exploited to reveal sensitive information about individuals in the original dataset. When it comes to EHR data, privacy evaluation typically includes testing the synthetic data for the following risks:

\begin{itemize}
    \item \textbf{Membership inference risk:} Risk against membership inference attack, where an adversary attempts to determine whether a specific data record was used in the training of a machine learning model.
    \item \textbf{Attribute inference risk:} Risk against attribute inference attack, where an adversary wants to infer private attributes of individuals based on the available data or model's predictions. 
    \item \textbf{Nearest neighbor adversarial accuracy risk:} Risk that is based on the Nearest Neighbor Adversarial Accuracy (NNAA) metric \cite{ehr-59}. This risk assesses the privacy risk of synthetic data by analyzing how closely the nearest neighbor (the most similar record) in the synthetic dataset resembles any real individual. A high similarity raises the risk of re-identification.   

\end{itemize}
    Table \ref{tab:privacy_risks} in the appendix displays the usage of these metrics in the surveyed papers. 
    
    Researchers usually apply Differential Privacy (DP) \cite{DifferentialPrivacy} mechanisms to mitigate privacy risks. DP is a rigorous and mathematically grounded framework designed to protect the privacy of individuals in the dataset. By introducing randomness into data queries or the data generation process, differential privacy ensures that the synthetic data doesn't compromise the privacy of any individual in the original dataset. It's often considered the gold standard for privacy protection in data analysis.

\textbf{Evaluation of fidelity} assesses how well the synthetic data reflects the real-world characteristics and features of the original dataset, both at the individual sample level and across the entire data distribution. High fidelity indicates that the synthetic data closely resembles the original data in terms of appearance and statistical properties. Notably, fidelity metrics differ across different data types due to their inherent structural and inherent differences. Table \ref{tab:fidelity-metrics} summarizes fidelity metrics for different types of medical data. Tables \ref{app-imaging} and \ref{app-ehr} in the appendix display the different fidelity metrics as long as the papers using them.  In most of the cases, the fidelity metrics used are originally designed for different types of data like natural images, not taking into account the characteristics and statistics of typical medical image modalities. For instance, FID (Fréchet Inception Distance) and IS (Inception Score) are widely used to evaluate the similarity between synthetic and real images, but they may not accurately reflect the subtle variations and noise patterns inherent in medical data, as these metrics are underpinned by a feature extractor pre-trained on the ImageNet dataset, which is more relevant for natural images.

\textbf{Evaluation of diversity} measures the extent to which synthetic data captures the full variability of the real data. 
Following \cite{faithful}, we consider diversity and fidelity as distinct yet complementary metrics in evaluating synthetic data. This ensures that the generative models are not only evaluated for reproducing average or common scenarios but also generating the broad spectrum of real-world variations.
Table \ref{tab:diversity-metrics} summarizes diversity metrics for different types of medical data.


\textbf{Clinical evaluation} involves assessing the synthetic data's correct representation of anatomical details consistent with real-world clinical scenarios, its ability to replicate key clinical features and characteristics, and its adherence to established medical guidelines and practices. These anatomical and biological details may be overlooked by fidelity metrics, and thus a clinically relevant evaluation is pivotal in introducing synthetic images into the medical field. Clinical validation also aim to verify that synthetic data can effectively support medical decision-making, patient care, and clinical outcomes, ultimately enhancing its utility and trustworthiness in healthcare applications.

\textbf{Qualitative evaluation}  including human assessment, also known as a visual Turing Test, which involves medical experts reviewing the synthetic data to try to distinguish it from the real data. In some cases,  qualitative evaluation only involves data visualization using techniques such as t-distributed Stochastic Neighbor Embedding (t-sne) and Principal Component Analysis (PCA), to compare the distribution of real and synthetic data visually.


\begin{table*}[ht]
\footnotesize
\centering
\begin{tabular}{cll}
\hline
\textbf{Modality}                   & \multicolumn{1}{c}{\textbf{Purpose of evaluation}}            & \multicolumn{1}{c}{\textbf{Fidelity Metric}}              \\ \hline
\multirow{13}{*}{EHR}               & \multirow{5}{*}{Dimension wise distributional similarity}     & Bernoulli   Success probability                           \\
                                    &                                                               & Chi-square test                                           \\
                                    &                                                               & Kullback-Leibler   divergence (KLD)                       \\
                                    &                                                               & Kolmogorov–Smirnov Test                                   \\
                                    &                                                               & Pearson test                                              \\ \cline{2-3} 
                                    & \multirow{4}{*}{Joint Distribution Similarity}                & Jensen-Shannon Divergence                                 \\
                                    &                                                               & Maximum-Mean   Discrepancy (MMD)                          \\
                                    &                                                               & Wasserstein   distance                                    \\
                                    &                                                               & Propensity score                                          \\ \cline{2-3} 
                                    & \multirow{3}{*}{Inter-dimensional   Relationship Similarity}  & Pearson pairwise correlation                              \\
                                    &                                                               & Pairwise   correlation difference\\
                                    &                                                               & Dimension-wise prediction                                 \\ \cline{2-3} 
                                    & \multirow{2}{*}{Latent   Distribution similarity}& Log-cluster metric                                        \\
 & &Latent space representation\\
 \cline{2-3}& specific-language model&Perplexity\\
 & \multirow{2}{*}{specific-time-series}&Autocorrelation function\\
 & &Patient trajectories\\ \hline
\multirow{7}{*}{Imaging Data}       & \multirow{7}{*}{Pixel-wise similarity   (image based)}& \\
                                    &                                                               & (Root/Normalized)   Mean squared error (MSE, RMSE, NRMSE) \\
                                    &                                                               & Structural similarity index   (SSIM)                      \\
                                    &                                                               & Multi Scale SSIM (MS SSIM)                                \\
                                    &                                                               & Peak signal-to-noise ratio   (PSNR)                       \\
 & &Universal Quality Index (UQI)\\
 & &Contrast Noise Ratio  (CNR)\\ \cline{2-3} 
                                    & \multirow{7}{*}{Feature-wise similarity (distrubution based)}& Fréchet inception distance (FID)                          \\
                                    &                                                               & Kernel Inception Distance (KID)
\\
 & &Inception score (IS)                                      \\
 & &Maximum Mean Discrepancy (MMD)
\\
 & &Learned Perceptual Image Patch Similarity (LPIPS)
\\
 & &Feature Similarity Index (FSIM)
\\
 & &feature distribution similarity (FDS)
\\
 \cline{2-3} 
& \multirow{3}{*}{Image/Text Alignment}&Bilingual Evaluation Understudy (BLEU)
\\
                                     & &Recall-Oriented Understudy for Gisting Evaluation (ROUGE)
\\
 & &Contrastive Language–Image Pre-training score (CLIP) 
\\ \hline
\multirow{3}{*}{Time-series   Data} & \multirow{2}{*}{Temporal   sequences similarity}&                                                           Dynamic Time   Warping (DTW)                                  
\\
 & &Time Warp Edit Distance (TWED)\\ \cline{2-3} 
                                    & \multirow{2}{*}{Temporal   Correlation}                       & Autocorrelation                                           \\
                                    &                                                               & Cross-correlation (CC)                                    \\ \hline
\multirow{4}{*}{Medical Text}       &  Similarity between machine and human translation             & Bilingual Evaluation Understudy (BLEU)                    \\ \cline{2-3} 
                                    &  \multirow{4}{*}{Similarity between real and generated text}& Cosine similarities                                           \\
                                     &                                                               & Jaccard Similarity                      \\
                                     &                                                               & ROUGE-N recall                     \\
                                    &                                                               & G2 - Test         \\ \hline
\end{tabular}
\caption{Fidelity Metrics}
\label{tab:fidelity-metrics}
\end{table*}

\begin{table*}[ht]
\footnotesize
\centering
\begin{tabular}{@{}ccc@{}}

\toprule
\textbf{Modality} & \textbf{Diversity Metric} & \textbf{Usage} \\ \midrule
\multirow{5}{*}{EHR} & Category coverage & \cite{s-34} \\
 & Generated Disease types & \cite{ehr-22} \\
 & Required sample number to   generate all diseases (RN) & \cite{ehr-22} \\
 & Normalized Distance & \cite{ehr-22} \\
 & Medical concept abundance & \cite{s-11} \\ \midrule
Signals & Diversity of Samples Score & \cite{r-250} \\ \midrule
\multirow{7}{*}{Imaging} & MS-SSIM & \cite{pinaya_brain_2022,multi-8,x-04} \\
 & Nearest SSIM difference & \cite{multi-1} \\
 & Visualization(Diversity) & \cite{multi-14} \\
 & Diversity score (DS) & \cite{ct-04} \\
 & LPIPS & \cite{ct-l-14,ct-l-16,oct-04} \\
 & Hammind Distance & \cite{oct-04} \\
 & Euclindean Distance & \cite{oct-14} \\ \bottomrule
\end{tabular}
\caption{Diversity Metrics}
\label{tab:diversity-metrics}
\end{table*}
\section{Results} \label{sec:results}
In our survey, we've meticulously distinguished between various modalities within the realm of medical data: EHR, Physiological Signals, Medical Images, and Clinical Notes, even though some might consider all these modalities as part of a patient’s EHR. Each of these categories represents a distinct dimension of patient health data, each contributing unique insights and challenges.

EHR encompass a variety of formats, including longitudinal EHR, time-dependent EHR, aggregated EHR, and snapshot EHR. These different formats capture the patient's medical history over time, ranging from specific episodes of care to comprehensive summaries.

Physiological Signals, on the other hand, capture real-time physiological data such as ECG, EEG, and various other signals. These signals provide crucial information about a patient's physiological state and are often used in monitoring and diagnosing medical conditions.

Medical Images are vital, and include modalities such as Ultrasound, Mammography, Dermoscopy, MRI, CT, X-Ray, OCT, as well as combinations of multiple modalities. Each imaging modality offers unique insights into different aspects of the patient's anatomy and pathology.

Clinical Notes provide detailed textual descriptions of patient observations, diagnoses, treatments, and other relevant information. These notes offer valuable context and insight into the patient's condition, treatment history, and healthcare journey.

This section transitions from general foundational knowledge to a more focused examination of findings obtained from the surveyed papers, organized according to different modalities. For each modality, we present a table summarizing various aspects of the reviewed papers, including the synthesis applications, models utilized for generation, and the adopted evaluation framework. The table content encompasses details such as the type of synthesis applications, specifics of the models employed, evaluation metrics, and code availability. We aimed to maintain a consistent table structure across modalities to facilitate a better understanding of methodologies and outcomes in medical data synthesis across diverse domains, although not always feasible. This structured approach enhances comprehension and comparison across different modalities in the field.

\subsection{EHR}
\label{sec:ehr}

\begin{figure*}[t]
    \centering
    \includegraphics[width=0.75\linewidth]{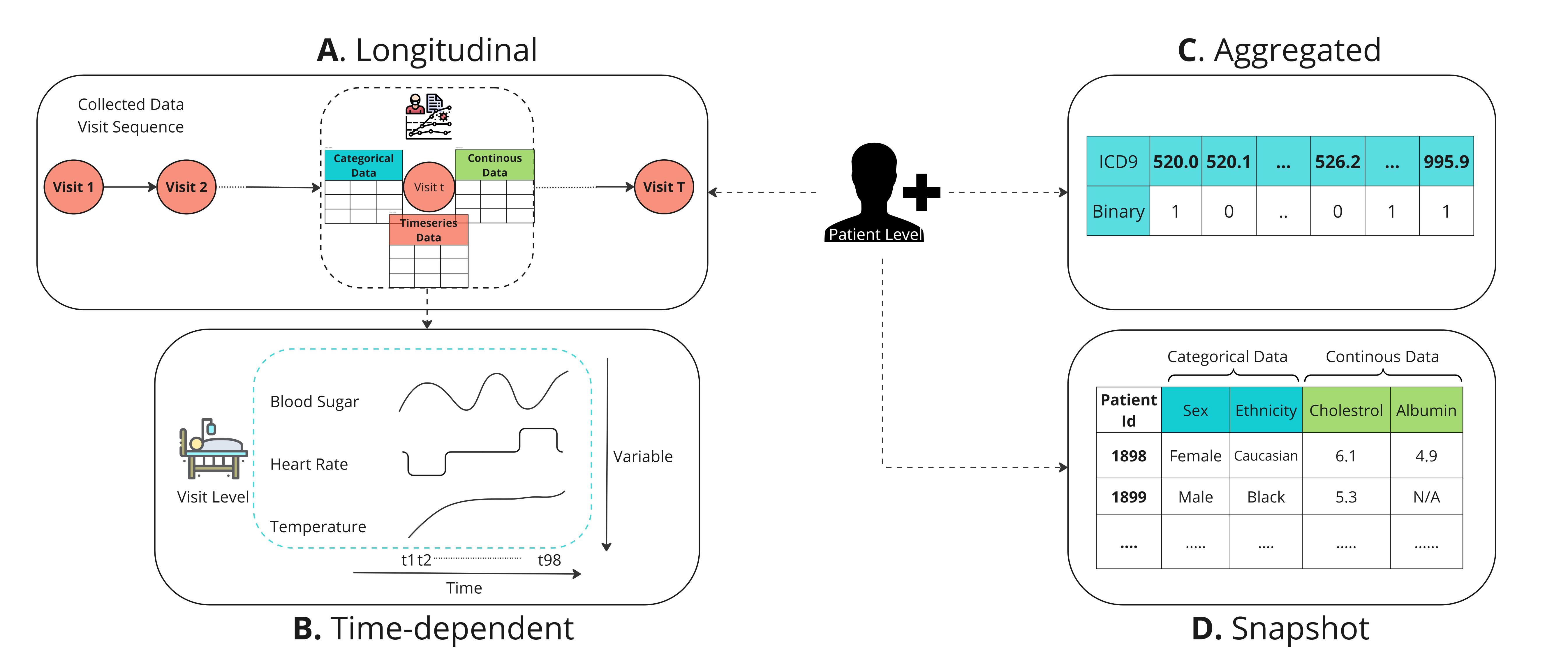}
    \caption{Different formats that EHR data exist in literature. \textbf{A. Longitudinal:} The most comprehensive yet least accessible due to privacy concerns, longitudinal EHRs track multiple attributes over the span of a patient’s life. A patient record $R_n$ consists of a sequence of visits of a patient to the healthcare provider $R_n = V_1, V_2, \ldots, V_T$, where each visit consists of a varying number of different data types (medical codes, lab results, diagnoses, clinical notes). In majority of cases, Each visit $V_t$ is a $K$-dimensional binary vector, where $V_T[k] = 1$ if the $k$-th code for patient $n$ was observed at visit $t$ and $0$ otherwise. $K$ denotes the cardinality of the set of possible codes.An Electronic Health Record (EHR) consists of a set of patient records $D = \{R_1, \ldots, R_n, \ldots, R_N\}$, where $N$ denotes the number of patients in the set. In this form, the temporal correlations between the visits should be maintained. \textbf{B. Time-dependent EHR:} This format captures regular, time-specific Data from monitoring devices, recorded at regular time intervals, focusing on physiological measurements such as Heart rate, respiratory rate,SpO2. \textbf{C. Aggregate:} This format flattens the diagnosis, medications, and procedure codes in the longitudinal data of a patient into aggregated data points represented in a single row with binary or count aggregations. The temporal and sequential aspect of the data is lost with this kind of processing, offering a cumulative overview instead \textbf{D. Snapshot or Cross-subsectional:} This is the most common format for public medical data, presenting a single static snapshot of a patient’s EHR focusing on selected attributes. }
    \label{fig:ehr-format}
\end{figure*}

EHRs are inherently heterogeneous, encompassing various types of data, including demographic data, medical codes assigned to diagnoses and procedures such as the International Classification of Diseases (ICD) codes, and time-series vital signs, among others. EHR typically consists of both static (e.g. demographics) and sequential numerical features (e.g. blood pressure), as well as multi-categorical features (e.g. medical codes). Given this complexity and heterogeneity, researchers and practitioners often differentiate four formats of EHR when developing generative models and generating synthetic data. These formats include longitudinal, time-dependent, aggregated, and snapshot, each with its own challenges and complexities, from modeling temporal dependencies and inter-dependencies between features to generating high-dimensional data and mixed-type features. Fig.\ref{fig:ehr-format} demonstrates different EHR data formats.

In this paper, we categorize different generative models based on their ability to handle the complexities of EHR data. We achieve this through a comprehensive evaluation that considers a diverse set of properties critical for generating realistic and heterogeneous EHR data. These properties include: capturing temporal relationships between events, generating data conditional on specific patient characteristics, creating data for uncommon diseases or underrepresented populations, effectively handling the high-dimensionality of EHR data, modeling missing data patterns, and generating both static and temporal features jointly. This comprehensive evaluation provides a thorough assessment of the current state-of-the-art in synthetic data generation for EHRs.

We start with an overview of early work in EHR synthesis and move to the different formats (longitudinal, time dependent, aggregated, and snapshot). A summary of the models is presented in Table \ref{tab:ehr}, indicating the technology used for generation, type of features and different properties that the generative models are capable of, as well as the evaluation framework.

\begin{table*}[h!]
\centering
\tiny
\begin{tabular}{ccclcccccccccccccccc}
\hline
\multirow{3}{*}{\textbf{Synthesis}} & \multirow{3}{*}{\textbf{Type}} & \multirow{3}{*}{\textbf{Features}} & \multirow{3}{*}{\textbf{Technology}} & \multirow{3}{*}{\textbf{Paper}} & \multirow{3}{*}{\textbf{DT}} & \multicolumn{6}{c}{\textbf{Features}} & \multicolumn{7}{c}{Evaluation} & \multirow{3}{*}{\textbf{Code}} \\ \cline{7-19}
 &  &  &  &  &  & \multirow{2}{*}{T} & \multirow{2}{*}{C} & \multirow{2}{*}{R} & \multirow{2}{*}{H} & \multirow{2}{*}{M} & \multirow{2}{*}{J} & \multicolumn{2}{c}{U} & \multirow{2}{*}{F} & \multirow{2}{*}{D} & \multirow{2}{*}{Q} & \multirow{2}{*}{C} & \multirow{2}{*}{P} &  \\ \cline{13-14}
 &  &  &  &  &  &  &  &  &  &  &  & Train & Test &  &  &  &  &  &  \\ \hline
\multirow{8}{*}{Longitudinal} & \multirow{3}{*}{GAN} & Dis & WGAN & \cite{s-36} & Pred & \checkmark (V) &  &  & \checkmark &  &  & S & R & \checkmark &  & \checkmark &  & \checkmark & \textbf{} \\
 &  & \multirow{2}{*}{Dis} & \multirow{2}{*}{WGAN-GP} & \multirow{2}{*}{\cite{ehr-22}} & \multirow{2}{*}{Pred} & \multirow{2}{*}{\checkmark} & \multirow{2}{*}{\checkmark} & \multirow{2}{*}{\checkmark} & \multirow{2}{*}{\checkmark} & \multirow{2}{*}{} & \multirow{2}{*}{} & \multicolumn{2}{c}{\multirow{2}{*}{**}} & \multirow{2}{*}{\checkmark} & \multirow{2}{*}{\checkmark} & \multirow{2}{*}{} & \multirow{2}{*}{} & \multirow{2}{*}{} & \multirow{2}{*}{\checkmark} \\
 &  &  &  &  &  &  &  &  &  &  &  & \multicolumn{2}{c}{} &  &  &  &  &  &  \\
 & VAE & Dis & EHR VAE & \cite{ehr-15} & Pred & \checkmark & \checkmark & - & \checkmark &  &  & S,A & R & \checkmark & \checkmark & \checkmark &  & \checkmark &  \\
 & DM & Dis & DDPM & \cite{ehr-57} & Pred & \checkmark &  &  & \checkmark &  &  & \checkmark &  &  &  &  &  &  & \textbf{} \\
 & \multirow{3}{*}{LM} & Dis & Prompt based LM & \cite{ehr-69} & Pred & \checkmark & \checkmark &  & \checkmark &  & \checkmark & S,A & R & \checkmark(Perp) &  &  &  & \checkmark & \checkmark \\
 &  & \multirow{2}{*}{Mix} & \multirow{2}{*}{\begin{tabular}[l]{@{}l@{}}Hierarchical\\ auto-regressive LM\end{tabular}} & \multirow{2}{*}{\cite{ehr-38}} & \multirow{2}{*}{Pred} & \multirow{2}{*}{\checkmark} & \multirow{2}{*}{\checkmark} & \multirow{2}{*}{\checkmark} & \multirow{2}{*}{\checkmark\checkmark} & \multirow{2}{*}{} & \multirow{2}{*}{\checkmark} & \multirow{2}{*}{S,A} & \multirow{2}{*}{R} & \multirow{2}{*}{\checkmark(Perp)} & \multirow{2}{*}{} & \multirow{2}{*}{} & \multirow{2}{*}{} & \multirow{2}{*}{\checkmark} & \multirow{2}{*}{\checkmark} \\
 &  &  &  &  &  &  &  &  &  &  &  &  &  &  &  &  &  &  &  \\ \hline
\multirow{16}{*}{Timeseries} & \multirow{3}{*}{GAN} & \multirow{16}{*}{Mix} & \multirow{3}{*}{WGAN} & \multirow{3}{*}{\cite{ehr-13}} & \multirow{3}{*}{Class,Pred} & \multirow{3}{*}{\checkmark} & \multirow{3}{*}{\checkmark} & \multirow{3}{*}{} & \multirow{3}{*}{\checkmark} & \multirow{3}{*}{} & \multirow{3}{*}{} & \multirow{3}{*}{S,R} & \multirow{3}{*}{S,R} & \multirow{3}{*}{} & \multirow{3}{*}{} & \multirow{3}{*}{} & \multirow{3}{*}{} & \multirow{2}{*}{\checkmark} & \multirow{3}{*}{} \\
 &  &  &  &  &  &  &  &  &  &  &  &  &  &  &  &  &  &  &  \\
 &  &  &  &  &  &  &  &  &  &  &  &  &  &  &  &  &  &  &  \\
 & VAE &  & Dynamical VAE & \cite{ehr-25} & Class & \checkmark & \checkmark & \textbf{\checkmark} & \textbf{} & \textbf{\checkmark} & \textbf{\checkmark} & S & R & \textbf{} & \textbf{} & \textbf{} & \textbf{} & \checkmark & \checkmark \\
 & \multirow{7}{*}{\begin{tabular}[c]{@{}c@{}}Hybrid :\\ VAE\\ +\\ GAN\end{tabular}} &  & \multirow{2}{*}{AE \& WGAN-GP} & \multirow{2}{*}{\cite{s-18}} & \multirow{2}{*}{Class} & \multirow{2}{*}{\checkmark} & \multirow{2}{*}{\textbf{}} & \multirow{2}{*}{\textbf{}} & \multirow{2}{*}{\textbf{}} & \multirow{2}{*}{\textbf{\checkmark}} & \multirow{2}{*}{\textbf{\checkmark}} & \multirow{2}{*}{S} & \multirow{2}{*}{R} & \multirow{2}{*}{\checkmark} & \multirow{2}{*}{\textbf{}} & \multirow{2}{*}{\textbf{*}} & \multirow{2}{*}{\textbf{}} & \multirow{2}{*}{\checkmark} & \multirow{2}{*}{} \\
 &  &  &  &  &  &  &  &  &  &  &  &  &  &  &  &  &  &  &  \\
 &  &  & AE \& WGAN-GP & \cite{ehr-21} & Pred & \checkmark & \checkmark & \textbf{} & \textbf{} & \textbf{} & \textbf{} & S & R & \textbf{} & \textbf{} & \textbf{} & \textbf{} & \checkmark &  \\
 &  &  & \multirow{3}{*}{VAE \& LSTM GANs} & \multirow{3}{*}{\cite{s-10}} & \multirow{3}{*}{Class} & \multirow{3}{*}{\checkmark} & \multirow{3}{*}{\checkmark} & \multirow{3}{*}{\textbf{-}} & \multirow{3}{*}{\textbf{}} & \multirow{3}{*}{\textbf{}} & \multirow{3}{*}{\textbf{}} & \multirow{3}{*}{S,A} & \multirow{3}{*}{R} & \multirow{3}{*}{\checkmark(time)} & \multirow{3}{*}{\textbf{}} & \multirow{3}{*}{\textbf{}} & \multirow{3}{*}{\textbf{}} & \multirow{3}{*}{\checkmark} & \multirow{3}{*}{\checkmark} \\
 &  &  &  &  &  &  &  &  &  &  &  &  &  &  &  &  &  &  &  \\
 &  &  &  &  &  &  &  &  &  &  &  &  &  &  &  &  &  &  &  \\
 &  &  & \begin{tabular}[l]{@{}l@{}}Combined Time-GAN\\  \& ADS-GAN\end{tabular} & \cite{TIME-ADS-GAN} & Pred & \checkmark &  &  &  &  & \textbf{\checkmark} & S & R & \checkmark & \checkmark & \textbf{*} & \textbf{} & \checkmark($\epsilon$) &  \\
 & \multirow{5}{*}{DM} &  & DDPM & \cite{s-34} & RL & \checkmark & \textbf{} & \textbf{} & \textbf{} & \textbf{} & \textbf{\checkmark} & S & R & \checkmark & \textbf{\checkmark} & \textbf{} & \textbf{} & \checkmark &  \\
 
 &  &  & \multirow{4}{*}{ \begin{tabular}[l]{@{}l@{}}mixed sequence\\DDPM\end{tabular}
} & \multirow{4}{*}{\cite{ehr-58}} & \multirow{4}{*}{Pred} & \multirow{4}{*}{\checkmark} & \multirow{4}{*}{\textbf{}} & \multirow{4}{*}{\textbf{}} & \multirow{4}{*}{\textbf{}} & \multirow{4}{*}{\textbf{\checkmark}} & \multirow{4}{*}{\textbf{\checkmark}} & \multirow{4}{*}{S,A} & \multirow{4}{*}{R} & \multirow{4}{*}{\checkmark} & \multirow{4}{*}{\textbf{}} & \multirow{4}{*}{\textbf{*}} & \multirow{4}{*}{\textbf{}} & \multirow{4}{*}{\checkmark} & \multirow{4}{*}{***} \\
 &  &  &  &  &  &  &  &  &  &  &  &  &  &  &  &  &  &  &  \\
 &  &  &  &  &  &  &  &  &  &  &  &  &  &  &  &  &  &  &  \\
 &  &  &  &  &  &  &  &  &  &  &  &  &  &  &  &  &  &  &  \\ \hline
\multirow{7}{*}{\begin{tabular}[c]{@{}c@{}}Aggregate,\\ Signals, \\ \& Snapshot\end{tabular}} & \multirow{7}{*}{\begin{tabular}[c]{@{}c@{}}Hybrid:\\ AE+GAN\end{tabular}} & \multirow{7}{*}{Dis} & \multirow{7}{*}{\begin{tabular}[l]{@{}l@{}}convolutional AE\\  \&  GAN\end{tabular}} & \multirow{7}{*}{\cite{ehr-16}} & \multirow{7}{*}{None} & \multirow{7}{*}{\checkmark} & \multirow{7}{*}{\checkmark} & \multirow{7}{*}{\textbf{}} & \multirow{7}{*}{\checkmark} & \multirow{7}{*}{\textbf{}} & \multirow{7}{*}{\textbf{}} & \multirow{7}{*}{} & \multirow{7}{*}{} & \multirow{7}{*}{\checkmark} & \multirow{7}{*}{\textbf{}} & \multirow{7}{*}{\textbf{}} & \multirow{7}{*}{\textbf{}} & \multirow{7}{*}{\textbf{\checkmark(RDP)}} & \multirow{7}{*}{\checkmark} \\
 &  &  &  &  &  &  &  &  &  &  &  &  &  &  &  &  &  &  &  \\
 &  &  &  &  &  &  &  &  &  &  &  &  &  &  &  &  &  &  &  \\
 &  &  &  &  &  &  &  &  &  &  &  &  &  &  &  &  &  &  &  \\
 &  &  &  &  &  &  &  &  &  &  &  &  &  &  &  &  &  &  &  \\
 &  &  &  &  &  &  &  &  &  &  &  &  &  &  &  &  &  &  &  \\
 &  &  &  &  &  &  &  &  &  &  &  &  &  &  &  &  &  &  &  \\ \hline
\multirow{4}{*}{Aggregate} & GAN & Mix & EMR-WGAN based & \cite{s-41} & None & \textbf{} & \checkmark & \textbf{-} & \checkmark & \textbf{} & \textbf{} &  &  & \checkmark & \textbf{} & \textbf{} & \textbf{} & \checkmark & \textbf{} \\
 & \multirow{3}{*}{DM} & \multirow{2}{*}{Mix} & \multirow{2}{*}{AE followed by SDE} & \multirow{2}{*}{\cite{s-24}} & \multirow{2}{*}{Class} & \multirow{2}{*}{} & \multirow{2}{*}{} & \multirow{2}{*}{} & \multirow{2}{*}{\checkmark} & \multirow{2}{*}{} & \multirow{2}{*}{} & \multicolumn{2}{c}{\multirow{2}{*}{\textbf{SRA}}} & \multirow{2}{*}{\checkmark} & \multirow{2}{*}{\checkmark} & \multirow{2}{*}{\checkmark} & \multirow{2}{*}{} & \multirow{2}{*}{} & \multirow{2}{*}{\checkmark} \\
 &  &  &  &  &  &  &  &  &  &  &  & \multicolumn{2}{c}{} &  &  &  &  &  &  \\
 &  & Dis & SDE & \cite{s-11} & Class & \textbf{} & \textbf{} & \textbf{} & \checkmark & \textbf{} & \textbf{} &  &  & \checkmark & \textbf{\checkmark} & \textbf{} & \textbf{} & \checkmark & \checkmark \\
\multirow{6}{*}{\begin{tabular}[c]{@{}c@{}}Aggregate, \\ \& Snapshot\end{tabular}} & \multirow{6}{*}{DM} & \multirow{2}{*}{D*,C} & \multirow{2}{*}{DDPM} & \multirow{2}{*}{\cite{s-15}} & \multirow{2}{*}{Class} & \multirow{2}{*}{} & \multirow{2}{*}{\checkmark} & \multirow{2}{*}{} & \multirow{2}{*}{\checkmark} & \multirow{2}{*}{} & \multirow{2}{*}{} & \multirow{2}{*}{S} & \multirow{2}{*}{R} & \multirow{2}{*}{\checkmark} & \multirow{2}{*}{} & \multirow{2}{*}{} & \multirow{2}{*}{} & \multirow{2}{*}{} & \multirow{2}{*}{} \\
 &  &  &  &  &  &  &  &  &  &  &  &  &  &  &  &  &  &  &  \\
 &  & \multirow{4}{*}{D*,M} & \multirow{4}{*}{\begin{tabular}[l]{@{}l@{}}mixed sequence\\DDPM\end{tabular}} & \multirow{4}{*}{\cite{s-12}} & \multirow{4}{*}{Class} & \multirow{4}{*}{\textbf{}} & \multirow{4}{*}{\textbf{}} & \multirow{4}{*}{\textbf{}} & \multirow{4}{*}{\checkmark} & \multirow{4}{*}{\textbf{}} & \multirow{4}{*}{\textbf{}} & \multirow{4}{*}{S,A} & \multirow{4}{*}{R} & \multirow{4}{*}{\checkmark} & \multirow{4}{*}{\textbf{}} & \multirow{4}{*}{\textbf{}} & \multirow{4}{*}{\textbf{}} & \multirow{4}{*}{\checkmark} & \multirow{4}{*}{\checkmark} \\
 &  &  &  &  &  &  &  &  &  &  &  &  &  &  &  &  &  &  &  \\
 &  &  &  &  &  &  &  &  &  &  &  &  &  &  &  &  &  &  &  \\
 &  &  &  &  &  &  &  &  &  &  &  &  &  &  &  &  &  &  &  \\ \hline
\multirow{14}{*}{Snapshot} & \multirow{14}{*}{GAN} & \multirow{14}{*}{Mix} & WGAN-GP  based & \cite{ehr-59} & None & \textbf{} & \textbf{} & \textbf{} & \checkmark & \textbf{} & \textbf{} &  &  & \checkmark & \textbf{} & \textbf{*} & \textbf{} & \checkmark & \textbf{} \\
 &  &  & \multirow{4}{*}{\begin{tabular}[l]{@{}l@{}}ADS-GAN:\\ WGAN-GP based\end{tabular}} & \multirow{4}{*}{\cite{ehr-12}} & \multirow{4}{*}{Pred} & \multirow{4}{*}{\textbf{}} & \multirow{4}{*}{\textbf{*}} & \multirow{4}{*}{\textbf{}} & \multirow{4}{*}{\textbf{}} & \multirow{4}{*}{\textbf{}} & \multirow{4}{*}{\textbf{}} & \multirow{4}{*}{S} & \multirow{4}{*}{R} & \multirow{4}{*}{\checkmark} & \multirow{4}{*}{\textbf{}} & \multirow{4}{*}{\textbf{}} & \multirow{4}{*}{\textbf{}} & \multirow{4}{*}{\textbf{\checkmark ($\epsilon$)}} & \multirow{4}{*}{\textbf{}} \\
 &  &  &  &  &  &  &  &  &  &  &  &  &  &  &  &  &  &  &  \\
 &  &  &  &  &  &  &  &  &  &  &  &  &  &  &  &  &  &  &  \\
 &  &  &  &  &  &  &  &  &  &  &  &  &  &  &  &  &  &  &  \\
 &  &  & MLP  gen \& disc & \cite{ehr-26} & Class,Reg & \textbf{} & \textbf{} & \textbf{} & \textbf{} & \textbf{} & \textbf{} & S & R & \textbf{} & \textbf{} & \textbf{*} & \textbf{} & \checkmark & \textbf{} \\
 &  &  & \multirow{8}{*}{CTGAN} & \multirow{8}{*}{\begin{tabular}[c]{@{}c@{}}
\cite{ehr-24}, \\
\cite{ehr-50}, \\
\cite{ehr-47}, \\ 
\cite{ehr-52}, \\ 
\end{tabular}} & \multirow{8}{*}{Class} & \multirow{8}{*}{\textbf{}} & \multirow{8}{*}{\checkmark} & \multirow{8}{*}{\textbf{}} & \multirow{8}{*}{\textbf{}} & \multirow{8}{*}{\textbf{}} & \multirow{8}{*}{\textbf{}} & \multirow{8}{*}{S,A} & \multirow{8}{*}{R} & \multirow{8}{*}{\checkmark} & \multirow{8}{*}{\textbf{}} & \multirow{8}{*}{\textbf{}} & \multirow{8}{*}{\textbf{}} & \multirow{8}{*}{\textbf{}} & \multirow{8}{*}{\checkmark} \\
 &  &  &  &  &  &  &  &  &  &  &  &  &  &  &  &  &  &  &  \\
 &  &  &  &  &  &  &  &  &  &  &  &  &  &  &  &  &  &  &  \\
 &  &  &  &  &  &  &  &  &  &  &  &  &  &  &  &  &  &  &  \\
 &  &  &  &  &  &  &  &  &  &  &  &  &  &  &  &  &  &  &  \\
 &  &  &  &  &  &  &  &  &  &  &  &  &  &  &  &  &  &  &  \\
 &  &  &  &  &  &  &  &  &  &  &  &  &  &  &  &  &  &  &  \\
 &  &  &  &  &  &  &  &  &  &  &  &  &  &  &  &  &  &  &  \\ \hline
\end{tabular}
\caption{Overview of all summarized and discussed studies on EHR synthesis with information about the technology utilized, the downstream task (DT), the evaluation procedure and code availability. The table lists the key properties evaluated for generative models of EHRs, including (T) preservation of temporal correlations and dependencies, (C) ability for conditional generation, (R) generation of uncommon diseases or underrepresented sub-populations, (H) capability of generating high-dimensional formats, (M) modeling missing patterns, (J)joint generation of static and temporal features. In the privacy column of the evaluation, the check mark indicates the presence of privacy evaluation, while the text in parenthesis indicate the method that the model follow to preserve privacy. In majority of the cases, the privacy is inherent, meaning that it is assumed in the design of the model. Evaluation is split according to the evaluation metrics reported, covering utility (U), fidelity (F), diversity (D), qualitative assessment (Q), clinical utility (C) and privacy (P) evaluation. In evaluation of utility (U), R stands for real data, S stands for synthetic data, and A stands for augmented data. SRA stands for Synthetic ranking agreement. In evaluation of fidelity, (perp) stands for perplexity measures, whereas (time) stands for time-series metrics. * stands for qualitative evaluation consisting of visualization of some plots only. **  stands for pretraining on synthetic, then fine tuning on real. *** stands for availability of code for original method only. RDP stands for Renyi DP. Dis stands for discrete features. C stands for continuous features. }

\label{tab:ehr}
\end{table*}

\subsubsection{Generative models- Longitudinal format }

Table \ref{tab:ehr} summarizes several models that generate synthetic longitudinal EHR data, emphasizing their ability to handle the high dimensionality of medical datasets, mainly discrete features, such as those including ICD codes. These models are capable of preserving temporal correlations across patient visits, with  \cite{s-36}  additionally maintaining these correlations at the visit level. Despite their capabilities, none of the models explicitly address the patterns of missing data in their design.

Beside GANs, VAE, DM, and LM have been utilized to generate longitudinal EHR. \cite{s-36} approached the problem using a two-stage strategy. The initial stage sequentially estimates temporal patterns and expected patient state from visits, incorporating a self-attention layer and using RNNs. The subsequent stage generates data, conditioned on the expected patient state, utilizing WGAN and the previous EMR-WGAN \cite{s-17} work. The model in \cite{ehr-22} uses Wasserstein GAN with gradient penalty (WGAN-GP) training with a GRU generator to recursively generate sequences and a sequential discriminator that can simultaneously distinguish whether individual visits are real and whether the entire sequence are real. The model proposed is focused on the generation of uncommon diseases, which is particularly beneficial for downstream tasks involving these less frequent conditions. To generate uncommon diseases, the conditional vector is smoothed into a conditional matrix for all visits to avoid the disease appearing only in the first visit or in every visit due to the characteristics of RNN-based models. \cite{ehr-15} used hierarchically factorized conditional VAE to generate discrete EHR sequences specific to medical conditions of interest

The models in \cite{ehr-38} and \cite{ehr-69} stand out by following a language model approach, with \cite{ehr-38} using hierarchical auto-regressive language model, treating EHR data as natural language sequences, predicting the probability of the presence of a potential medical code given the patient’s medical history and the previous codes. This model is uniquely capable of synthesizing both discrete (medical codes) and continuous (average values of lab tests) features, with extremely high dimensional, emulating the heterogeneous character of EHRs. Moreover, it uses demographic and chronic disease phenotypes for conditional generation, which helps create balanced datasets for training downstream tasks on rare conditions.
The model in \cite{ehr-69} allows for prompt-based generation and offers the flexibility to modulate diversity using a variety of methods found in the natural language generation literature. This is in contrast to GANs and VAEs, which rely on sampling  random noise vectors to inject diversity, an approach that is uncontrollable and compromises the quality of the generated content. 

Models for synthesizing longitudinal EHR outlined in Table \ref{tab:ehr} are assumed to be designed with inherent privacy-preserving features within their architectural frameworks, and privacy evaluation is performed in most cases. While all models were evaluated for their utility in downstream tasks specifically for training downstream AI models or augmenting the training set, the diffusion model in \cite{ehr-57} lacked fidelity evaluation. For evaluating the fidelity of the data from the language models, metrics adopted from language literature are used such as perplexity measure \cite{ehr-69,ehr-38}. Notably, \cite{ehr-22} evaluates for the diversity of the generated disease types, checking for the presence and representation of all possible diseases and medical codes in the synthetic data.

\subsubsection{Generative models- Time-dependent format }

The capacity to preserve temporal dependencies and correlations among clinical features is a critical functionality for models generating synthetic time dependent EHR data. The challenge of high-dimensional data generation is less pronounced in time-dependent than longitudinal format since the extensive data typically associated with medical codes is not utilized. The WGAN based model \cite{ehr-13} is the exception, with the ability to generate high-dimensional, time-dependent records featuring 714 attributes. Conversely, \cite{s-10}, in its initial design, limited the dimensionality of variables to less than 100, potentially constraining its capacity to represent the complexity of medical data fully. Models like the ones in\cite{s-18, s-34, ehr-25, ehr-58} focus as well on accurately reflecting the heterogeneous nature of real-world EHR through a combination of static and temporal features.  \cite{ehr-25,s-18,ehr-58} are all proficient in handling the patterns of missing data, a critical aspect that \cite{ehr-21} opts to eliminate through data imputation, potentially losing valuable information inherent in the absence of data . Iconically, \cite{s-18} manages varying time-series sequence lengths.

The idea of combining an autoencoder with a GAN is widely used, mainly to establish a shared latent space representation encompassing mixed types of EHRs features, including both continuous and discrete-valued data, prior to the generation process. For instance, both \cite{ehr-21} and \cite{s-18} employ autoencoder networks followed by a WGAN-GP. While \cite{ehr-21} utilizes a LSTM based recurrent autoencoder, \cite{s-18} employs a sequential encoder-decoder network. Expanding upon this strategy, \cite{s-10} incorporates a dual LSTM-based VAE. It then employs bilaterally coupled LSTM GANs to ensure capturing temporal correlations within diverse EHR features.  TIME-ADS-GAN \cite{TIME-ADS-GAN} combines research on time-dependent synthetic data (TimeGAN \cite{timegan}) and  privacy preserving static synthetic data (ADS-GAN  \cite{ehr-12}), where the TimeGAN also utilize the latent space of the autoencoder for better synthesis.

 \cite{s-34} and  \cite{ehr-58} are novel approaches that utilize diffusion models to generate mixed feature data. While \cite{s-34} resorts to Gaussian diffusion for generating discrete sequences, treating them similarly to real-valued sequences but with further post-processing of the model output,  \cite{ehr-58} on the other hand simultaneously generates (without need for post-processing) time series discrete (using multinomial diffusion \cite{multinomialdiffusion}) and continuous (using Gaussian diffusion) data that preserve temporal dependencies because of the use of a BRNN instead of U-net in the backward diffusion step.

Multiple models in this category allow for conditional generation, including \cite{ehr-13,ehr-25,ehr-21,s-10}.  \cite{s-10} utilizes condition-specific labels (outcomes: ICU mortality, hospital mortality, 30-day re-admission, hospital discharge …) to improve model performance, although its conditional generation capabilities are limited and not extendable to patient-specific EHRs based on more granular level information. In contrast, \cite{ehr-25} excels in this area by enabling conditional generation that considers patient demographic data, effectively addressing the needs of underrepresented sub-populations and contributing to fairness in downstream machine learning applications.



The privacy of \cite{s-10} is empirically evaluated using membership inference attack and differential privacy. Differential privacy analysis is performed by evaluating the performance of the same downstream medical intervention prediction tasks used for utility evaluation but under various differential privacy budgets. Similarly, \cite{ehr-13,s-18,ehr-21,s-10,s-34,ehr-58} are evaluated  against various adversary privacy attacks. \cite{ehr-25} indirectly supports privacy by ensuring that synthetic data do not replicate the training data using memorization analysis (latent space nearest neighbor). While all models were evaluated for their utility, only \cite{s-34} conducted a diversity analysis to evaluate the coverage of all demographic features present in the real dataset as well as uniquely evaluating the utility of their model for the downstream task of reinforcement learning. However, a limitation in the evaluation of this form of time-dependent synthetic EHR arises from the scarcity of time-specific metrics. Instead, the authors opted to assess the temporal aspect of the data by evaluating the utility for temporal tasks. 

\subsubsection{Generative models- Aggregated format }



In the context of aggregated format generation for EHR data, all models demonstrate a range of capabilities, particularly in handling high-dimensional data. However, a significant limitation across all these models is their lack of consideration for missing data patterns. Additionally, they do not integrate both static features, like patient demographics, and temporal features, such as medical readings over time, in their generation process. This gap limits their ability to fully replicate the nuanced and comprehensive nature of real-world EHR data. Beyond the primary focus on aggregated format EHR, several models have also shown effectiveness in other EHR formats.  For example, \cite{ehr-16} have been effective with EEG and ECG data, demonstrating their ability to preserve temporal correlations and maintain the integrity of time-series patterns. Additionally, \cite{s-12,ehr-16,s-15} proved to be effective in generating snapshot format data as well.

Early works on GANs for EHRs were centered around producing structured, discrete high dimensional aggregated EHRs, including diagnosis and billing ICD codes \cite{s-9}. The initial GAN architecture that generated discrete EHR, medGAN \cite{s-13},  utilized an autoencoder to address the original GAN architecture’s inability to generate discrete features in combination with a mini-batch averaging to mitigate mode collapse. Following the success of medGAN in generating discrete data, medWGAN and medBGAN \cite{s-16} were introduced, replacing the standard GAN with a WGAN-GP \cite{gulrajani2017improved} and boundary-seeking GANs (BGAN) \cite{BGAN}, respectively. The authors notably contributed to enhancing the quality of the generated data, surpassing that of the original medGAN. In MC-medGAN \cite{s-35}, adaptations to medGAN were proposed to facilitate a superior representation of multi-categorical data . Other endeavors have concentrated on enhancing training stability, such as the approach proposed in Electronic Medical Record Wasserstein GAN (EMR-WGAN) \cite{s-17}. This approach bypasses the autoencoder architecture used in medGAN, allowing the generator to directly produce synthetic data. Additionally, the authors implemented a filtering strategy to enhance GAN training specifically for clinical concepts and medical codes with low prevalence.

More research has been conducted in this direction,expanding the scope to cover not only discrete feature types but heterogeneous type (continuous and discrete) as well. \cite{ehr-16} employed a convolutional AE and convolutional GAN, while the model in \cite{s-41} is based on EMR-WGAN \cite{s-17} and generate mixed feature types. Recent models utilize diffusion models for generating aggregated format EHR.\cite{s-24} was the first to do so, followed by \cite{s-11,s-15}. \cite{s-24} compress the high-dimensional mixed features into a low-dimensional representation using an autoencoder and perform Stochastic Differential Equation (SDE) diffusion in the latent space, while \cite{s-11} apply SDE directly to discrete values, treating them like continuous numbers. \cite{s-15}, on the other hand, use a DDPM without requiring a pre-trained autoencoder, utilizing a modified U-net using 1D convolutional layers to accompany the tabular type of data, while \cite{s-12} adopt the TabDDPM  model \cite{tabddpm} and follow a mixed diffusion approach to generate realistic mixed-type tabular EHRs.

\cite{ehr-16,s-41,s-15} support conditional generation, with \cite{s-41} allowing for conditioning on demographic features such as age and gender, while \cite{s-15} facilitates class-based conditioning  by incorporating the idea of a classifier-guided sampling process \cite{classifierfree}. Despite these capabilities, these models do not specifically focus on generating data for uncommon diseases or underrepresented conditions, potentially overlooking important aspects of diversity in healthcare.

Empirical privacy risk analysis against various privacy attacks was performed on \cite{s-41, s-24} and demonstrated low privacy disclosure risk. However, there's room for improvement in the privacy mechanisms of these models, such as the potential incorporation of differential privacy, rather than relying solely on inherent privacy traits of the generative model.  RDP-CGAN \cite{ehr-16} stands out by explicitly implementing Renyi DP \cite{renyiDP}, a relaxed variant of traditional differential privacy, into the conditional GAN architecture. Meanwhile, the use of TabDDPM in \cite{s-12}, despite proving its excellence in data quality, utility, and augmentation, faces challenges when evaluated for privacy, underscoring the often-present trade-off between privacy and utility in synthetic data generation.

\subsubsection{Generative models- Snapshot format }


As the snapshot format lack the time aspect, none of the models are designed to preserve temporal correlations within the datasets they handle. Additionally, \cite{ehr-59} stands out as the only model capable of handling high-dimensional data (handling 324 features). Missing data patterns are not addressed by any of the models, nor the joint static and temporal features representation.
The models in \cite{ehr-59,ehr-12,ehr-26,ehr-52} were all evaluated for privacy, with ADS-GAN  \cite{ehr-12} explicitly focusing on preserving privacy by minimizing patient identifiability . They mathematically define $\epsilon$-identifiability based on the probability of re-identification given the combination of all data on any individual patient. ADS-GAN use a modified conditional GAN framework where conditioning variables are not pre-determined but rather optimized from the real patient records, while minimizing patient identifiability. \cite{ehr-59} introduced an array of metrics to evaluate realism and privacy together called nearest neighbor adversarial accuracy (NNAA). \cite{ehr-26} later propose a privacy preserving GAN that uses these metrics to assess the privacy risk.


Snapshot format data usually have the same properties of tabular data: correlated features, mixed feature type, highly imbalanced categorical features and non-Gaussian distribution of features. These properties impose challenges on the synthetic data generation process. Conditional tabular GANs (CTGANs) \cite{ctgan} have been recognized for their effectiveness in generating tabular data, mitigating the challenges and difficulties of this task, and have been adapted to ensure privacy, while allowing for conditional generation.  These models have been widely used in snapshot EHR format synthesis.The introduction of DP-CTGAN \cite{ehr-24} offers a differentially private variant that incorporates random perturbation into the GAN's critic for privacy protection. The federated version, FDP-CTGAN, extends this approach to a distributed data setting, evaluated across multiple medical datasets. \cite{ehr-50} uses CTGANs for data augmentation purposes (with varying sizes of augmentation) on classification problems. However, the authors of the paper resort to other data balancing methods like SMOTE and ADASYN instead of using the CTGAN for data balancing as well.  The authors of \cite{ehr-47}, on the other hand, use CTGAN as well to address the issue of poor performance of underrepresented sub-populations. They train CTGAN for each sub-population separately and incorporate synthetic samples into the training set of each, consequently increasing the generalizability of predictive models towards underrepresented sub-populations. \cite{ehr-52} utilizes synthetic data generated using CTGAN as a proxy for scarce real-world EHR for the patient length-of-stay multi-class classification task.

\subsubsection{Diverse papers from diverse formats}
Following the success of the HealthGAN model in \cite{ehr-13,ehr-59}, \cite{ehr-70} extends the utility, privacy, and fidelity evaluation to include fairness evaluation as well. \cite{ehr-19} implemented federated learning to train separate GANs locally at each organization, and then combining them into a central GAN that can generate synthetic discrete EHR, such as ICD-9 codes. \cite{ehr-42} evaluates four different GAN architectures: Tabular GAN (TGAN), CTGAN, WGAN-GP, and conditional TableGAN (CTABGAN) on four tabular medical datasets, three of which are in the snapshot format, and one is an EEG dataset. The aim of this study is to investigate the applicability of tabular GANs to medical data and analyze whether the GAN architectures specific to tabular data outperform the more general WGAN-GP by evaluation of model quality, fidelity and privacy.


\subsection{Time-series sgnals}\label{sec:signals}

In this section, we explore the generation of synthetic data for physiological signals, such as ECG and EEG. These signals are crucial for monitoring patient health. Our investigation delves into the synthesis techniques and evaluations frameworks specific to these type of signals, addressing challenges and opportunities in generating realistic physiological signals.

\label{sec:signals}

\subsubsection{ECGs}

\begin{table*}[h!]
\scriptsize
\centering
\begin{tabular}{ccllllcccccccc}
\hline
\multirow{3}{*}{\textbf{Application}} & \multirow{3}{*}{\textbf{Type}} & \multirow{3}{*}{\textbf{Modality}} & \multirow{3}{*}{\textbf{Technology}} & \multirow{3}{*}{\textbf{Paper}} & \multirow{3}{*}{\textbf{DT}} & \multicolumn{7}{c}{\textbf{Evaluation}} & \multirow{3}{*}{\textbf{Code}} \\ \cline{7-13}
 &  &  &  &  &  & \multicolumn{2}{c}{U} & \multirow{2}{*}{F} & \multirow{2}{*}{D} & \multirow{2}{*}{Q} & \multirow{2}{*}{C} & \multirow{2}{*}{P} &  \\ \cline{7-8}
 &  &  &  &  &  & Train & Test &  &  &  &  &  &  \\ \hline
\multirow{8}{*}{Unconditional} & Other & 12-lead  & Bi-LSTM and CNN & \cite{r-228} & Class & S & R &  &  &  &  &  &  \\
 & \multirow{6}{*}{GAN} & single-lead  & \begin{tabular}[l]{@{}l@{}}Composite GAN:\\ LSTM-GAN and DCGAN\end{tabular} & \cite{r-252} & Class & A & R &  &  &  &  &  &  \\
 &  & single and 2-lead  & LSTM-based:TS-GAN & \cite{r-256} & Class & A & R & \checkmark &  & * &  &  &  \\
 &  & 12-lead  & LSTM-based:BiLSTM & \cite{r-74} & None &  &  & \checkmark &  &  &  &  &  \\
 &  & single-lead  & Multiple GANs & \cite{s-9} & Class & A & R & \checkmark &  & \checkmark & \checkmark &  & \checkmark \\
 &  & short (10s) 12-lead  & WaveGAN,Pulse2Pulse & \cite{r-65} & None &  &  &  &  &  &  &  & \checkmark \\
 &  & single-lead  & image-based:TTS-GAN & \cite{s-51} & None &  &  & \checkmark &  & * &  &  & \checkmark \\
 & DM & 2-lead  & image-based:DDPM & \cite{s-8} & Class & S & R & \checkmark &  &  &  &  & \checkmark \\ \hline
\multirow{6}{*}{Class or Label Cond. } & \multirow{2}{*}{DM} & 12-lead  & DSAT-ECG & \cite{r-233} & Class & R & R,S & \checkmark &  &  &  &  &  \\
 &  & 2-lead  & DDPM-based:DiffECG & \cite{r-229} & Class & A & R & \checkmark &  & \checkmark &  &  &  \\
 & \multirow{2}{*}{GAN,VAE} & single-lead  & CVAE,CWGAN & \cite{r-232} & Class & S,R & R,S &  &  &  &  &  &  \\
 &  &  & PHYSIOGAN & \cite{r-250} & Class & S & R & \checkmark & \checkmark &  &  &  & \checkmark \\
 & \multirow{2}{*}{GAN} & single-lead  & \begin{tabular}[l]{@{}l@{}}WGAN-GP-based:\\ AC-WGAN-GP\end{tabular} & \cite{r-64} & Class & A & R &  &  &  &  &  & \checkmark \\
 &  & beat generation & DCCGAN & \cite{r-245} & Class & S & R & \checkmark &  & \checkmark &  &  &  \\ \hline
Attribute Cond & VAE & 12-lead  & cVAE & \cite{r-77} & None &  &  & \checkmark &  &  &  &  &  \\ \hline
Text-guided & VAE,GAN & 12-lead  & Auto-TTE & \cite{s-44} & Class & S & R & \checkmark & \checkmark & \checkmark &  &  & \checkmark \\ \hline
\multirow{2}{*}{Intra-modal trans.} & \multirow{2}{*}{GAN} & single-lead to 12-lead & Bi-LSTM and CNN & \cite{r-235} & None &  &  & \checkmark &  &  &  &  & \checkmark \\
 &  & 2-lead to 12 LEAD & StarGAN v2 & \cite{r-236} & Class & A & R &  &  & * &  &  & ** \\ \hline
\multirow{2}{*}{Inter-modal trans.} & DM & PPG-to-ECG & RDDM & \cite{r-251} & Class & S & R & \checkmark &  &  & \checkmark &  & \checkmark \\
 & GAN,AE & \begin{tabular}[l]{@{}l@{}}PCG to ECG,\\ VCG to 12-lead transfer\end{tabular} & \begin{tabular}[l]{@{}l@{}}classical GAN,adversarial AE,\\ modality transfer GAN\end{tabular} & \cite{r-241} & Class & S,R & R,S & \checkmark &  & * &  &  &  \\ \hline
\multirow{3}{*}{Other Conditioning} & \multirow{2}{*}{GAN} & \begin{tabular}[l]{@{}l@{}}Conditioned on \\  clinical Knowledge\end{tabular} & \begin{tabular}[l]{@{}l@{}}WGAN-GP-based:\\ CardiacGen\end{tabular} & \cite{r-98} & Class & S,A & R & \checkmark &  &  & \checkmark &  & \checkmark \\
 &  & \begin{tabular}[l]{@{}l@{}}Conditioned on other\\ECG statements\end{tabular} & Conditional GAN & \cite{r-230} & Class & A & R &  &  & \checkmark &  &  &  \\
 & DM & \begin{tabular}[l]{@{}l@{}}Conditioned on other \\ECG statements \end{tabular} & SSSD-ECG & \cite{s-42} & Class & S,R & R,S &  &  & \checkmark &  &  & \checkmark \\ \hline
\end{tabular}
\caption{Overview of all summarized and discussed studies on \textbf{ECG} synthesis with information about the technology utilized, the downstream task (DT), the evaluation procedure and code availability. Evaluation is split according to the evaluation metrics reported, covering utility (U), fidelity (F), diversity (D), qualitative assessment (Q), clinical utility (C) and privacy (P) evaluation. In evaluation of utility (U), R stands for real data, S stands for synthetic data, and A stands for augmented data. * stands for qualitative evaluation consisting of visualization of some plots only.** stands for availability of code for original method only. Acronyms: Deep convolutional conditional GAN (DCCGAN), Region-Disentangled Diffusion Model (RDDM)}

\end{table*}

Various technologies are employed in the synthesis of ECG signals with many proposed methods, such as Pulse2Pulse, WaveGAN, BiLSTM GAN, and TS-GAN, tailored to ECG data specifics such as morphology (appearance), durations, and intervals. The models were assessed for their ability to generate various synthetic ECG formats, including 2-lead synthesis, 12-lead synthesis, single-lead synthesis. ECG signals are commonly recorded in various formats, depending on the number of leads used. Each lead measures the electrical activity of the heart from a different angle, providing unique information about heart function. Each type of ECG provides a different level of detail and is suited for different clinical needs and settings. The 12-lead ECG is the most informative for diagnostic purposes, while single-lead ECGs are useful for continuous, everyday heart rhythm monitoring. An ECG signal is divided into heartbeats (also known as cardiac cycles), and a typical normal ECG heartbeat consists of three waves, the P wave, the QRS complex which contain the R peak, and the T wave.

Atrial Fibrillation (AF) is a cardiac disorder where abnormal electrical impulses start firing in the atria, causing a faster heart rate, with two known clinical markers: The P waves are either absent or replaced by fibrillatory wave, and the distance between consecutive R peaks is irregular, causing a phenomena called Heart Rate Variability (HRV). To simulate the two AF markers, \cite{r-252} employs a dual-GAN approach, combining LSTM-GAN to generate the R-R interval time-series and DCGAN for generating signal morphology in terms of missing P waves or presence of abnormal fibrillatory waves before the QRS complex. This underscores the necessity of composite models to capture the intricate characteristics of the signal. 

Most existing approaches to multi-lead ECG synthesis generate ECG lead signals that are independent of each other. To mitigate this, \cite{r-74} introduces a 2 dimension bidirectional LSTM GAN , where the second dimension captures the physiological and spatial correlation among ECG leads of the same recording when synthesizing 12-lead ECGs.\cite{r-256} also presents a LSTM based GAN architecture, where the discriminator and generator are based on LSTMs, which demonstrates superiority over previous LSTM-based models like TimeGAN \cite{timegan} and C-RNN-GAN \cite{crnngan}. \cite{r-228} follows a hybrid approach combining LSTM and Convolutional Neural Network (CNN) layers, where the model utilizes bi-LSTM layers to extract temporal features and CNN layers to extract spatial features from ECG signals.

The GAN technologies are further diversified in \cite{r-65}, which develops two state-of-the-art GAN methods, Pulse2Pulse (inspired by the U-Net architecture) and WaveGAN, to generate synthetic 12-lead ECG signals. Here, the Pulse2Pulse GAN outperforms its counterpart, highlighting the importance of architecture choice in generating realistic ECGs.\cite{s-8} and \cite{s-51} share a common approach to treating time-series data akin to images, with \cite{s-8} employing a diffusion model for this purpose, while \cite{s-51} utilizes tts-gan, a transformer-based GAN architecture. Both studies pivot away from traditional time series modeling, instead adopting image processing paradigms for ECGs. 

ECGs are typically categorized based on heartbeat types, including normal beats (N), premature ventricular contraction beats (V), and fusion beats (F), with various models leveraging these class labels for conditioned generation tasks. \cite{r-64} developed a WGAN-GP inspired model that generates class-conditioned single-lead ECG data. The work in \cite{r-250} introduces an innovative hybrid model that combines the generative capabilities of GANs with VAEs. The resulting hybrid model consists of recurrent neural networks encoder and decoder, as well as a discriminator component to improve the quality of the class conditional generation. \cite{r-229} employs a DDPM based model for a range of tasks including ECG signal generation, partial signal completion, and full heartbeat forecasting. This model is capable of conditional generation across various classes, indicating a flexible approach to the creation of ECG data. 

Conditional VAEs play a role in conditional generation, as demonstrated by \cite{r-232} and \cite{r-77}. The work in \cite{r-232}  showcased the VAE model’s faster convergence and reduced complexity (fewer parameters) compared to traditional GAN models. \cite{r-77} also utilizes conditional VAEs to generate  12-lead ECG signals that are conditioned on specific subject characteristics, such as age, sex, and BMI, as well as image-derived data, which provides information on heart position and orientation from Cardiac MRI. This approach suggest a higher level of personalization in the synthetic data, potentially leading to more accurate representations of patient-specific ECG signals.

Structured State Space Model (SSSM) is a linear state space transition equation that proved to be a promising model to captures long-term dependencies in time series data \cite{SSS}. This is achieved by discretizing the input and output relations and representing them as a convolution operation. SSSD-ECG \cite{s-42} and DSAT-ECG \cite{r-233} combine SSSM with diffusion models to generate synthetic ECG signals. The reader is referred to \cite{alcaraz2023diffusionbased} for more information about the integration of SSSM with diffusion models. \cite{s-42} condition on representative referential ECG signals to guide the generation process, demonstrating superior performance over prior GAN-based models such as WaveGAN and Pulse2Pulse. \cite{r-233}, on the other hand, synthesize conditional 12-lead ECG based on 12 heart rhythm classes,  outperforming that of SSSD-ECG in \cite{s-42}  and earlier GAN works, including WaveGAN and Pulse2Pulse, in terms of fidelity and utility evaluation. 

Other than conditioning on class labels, \cite{r-98} relies on two WGAN-GP-based models conditioned on HRV characteristics and pulse morphological properties, underscoring the potential of incorporating domain knowledge into the generative process to generate synthetic but physiologically plausible cardiac signals.

Auto-TTE, presented in \cite{s-44}, introduces an auto-regressive model integrating text descriptors and raw ECG data for comprehensive 12-lead ECG synthesis, allowing for patient-specific input like age and gender, marking a significant leap in the field.The generation process in Auto-TTE involves two phases. Initially, raw ECG signals are converted into a sequence of discrete tokens using a VQ-VAE. These tokens, alongside text tokens processed by a BPE tokenizer and patient specific tokens like age and gender, are then fed into an autoregressive Transformer-decoder model. This model is trained to generate ECG tokens autoregressively, which are then decoded back into ECG signals using  Hifi-GAN decoder architecture, completing the synthesis process.

In another application, \cite{r-235} presents a model capable of translating single-lead ECG inputs into a complete 12-lead set, utilizing a GAN framework with a bidirectional grid LSTM generator and a CNN discriminator, showcasing another innovative instance of LSTM and CNN hybrid model architecture. \cite{r-236} also focuses on synthesizing twelve-lead ECGs from two asynchronous leads. Drawing inspiration from GAN-based image-to-image translation models, \cite{r-236} develops a model primarily influenced by StarGAN v2 and its adaptations in speech processing  to synthesize twelve-lead ECGs from two asynchronous leads. These translation capabilities are pivotal for enhancing the applicability of ECG data in clinical settings where different lead configurations are used.

The evaluation criteria of the synthetic data primarily concentrated on fidelity and utility, while diversity received limited attention in the cited studies \cite{r-250, s-44}. Generic fidelity metrics, including Frechet Distance, RMSE, and MMD, were prevalent. Signal-specific metrics such as dynamic time warping and time warp edit distance, that consider the temporal alignment of two time sequences, were introduced in only a handful of studies \cite{ s-9,s-8,r-233, r-245}. Some studies extended the evaluation by incorporating qualitative analysis from medical experts or by engaging in clinical validation. This involved comparing generated beats to templates and assessing the realism of HRV features in synthetic data. However, privacy was not explicitly mentioned in any of the papers, , indicating an area for potential future emphasis in model development and evaluation.

\subsubsection{EEGs and other signals}

Under the scope of unconditional synthesis, \cite{eeg-1} employs a latent diffusion model to generate 30-second windows of synthetic EEG signals, while \cite{eeg-5} uses a  generative model called Causal Recurrent Variational Autoencoder (CR-VAE) integrating Granger causality \cite{Granger} into a recurrent VAE framework to infer relationships between different time series variables. Given the success of language models, specifically GPT-2 \cite{radford2019language}, in different applications such as  patent claims \cite{patientclaim} and stock market analysis \cite{stock}, \cite{eeg-8} wants to explore the capabilities of language models in a brand new field of application: the generation of bio-synthetic signals. For this reason, \cite{eeg-8} trains GPT-2 language models for generating synthetic biological signals, where a model is trained for each class separately. 

\cite{eeg-6} uses Class conditional Wasserstein GAN (CWGAN) for generating synthetic EEG signals. However, instead of using the raw EEG signal directly, features extracted from the EEG are utilized due to the high dimensionality of the raw signal. \cite{eeg-2} uses DDPMs to generate realistic synthetic data for a variety of physiological signals (LFP, ECoG, EEG) and evaluates if dataset-specific features such as sharp wave-ripples and ross-channel couplings were reproduced in the generated data. The work in \cite{eeg-3} employs a diffusion model (DiffEEG) that uses the Short-time Fourier transform (STFT) spectrograms of the EEG signal segments as the condition for synthesis.

\begin{table*}[h!]
\centering
\scriptsize
\begin{tabular}{ccllllcccccccc}
\hline
\multirow{3}{*}{\textbf{Application}} & \multirow{3}{*}{\textbf{Type}} & \multirow{3}{*}{\textbf{Modality}} & \multirow{3}{*}{\textbf{Technology}} & \multirow{3}{*}{\textbf{paper}} & \multirow{3}{*}{\textbf{DT}} & \multicolumn{7}{c}{Evaluation} & \multirow{3}{*}{\textbf{Code}} \\ \cline{7-13}
 &  &  &  &  &  & \multicolumn{2}{c}{U} & \multirow{2}{*}{F} & \multirow{2}{*}{D} & \multirow{2}{*}{Q} & \multirow{2}{*}{C} & \multirow{2}{*}{P} &  \\ \cline{7-8}
 &  &  &  &  &  & \textbf{Train} & Test &  &  &  &  &  &  \\ \hline
\multirow{4}{*}{Unconditional} & DM & EEG & LDM & \cite{eeg-1} & None &  &  & \checkmark &  &  &  &  & \checkmark \\
 & VAE & EEG,.Fmri & CRVAE & \cite{eeg-5} & Predict next samples & S & R & \checkmark &  & * &  &  & \checkmark \\
 & GAN & stereo EEG & TGAN & \cite{eeg-7} & Classification & S & R &  &  & * &  &  &  \\
 &LLM & EEG,EMG & GPT2 & \cite{eeg-8} & Classification & R,S,A & R,S &  &  &  &  &  & \checkmark \\ \hline
\multirow{2}{*}{Class-conditional} & DM & EEG,LFP, ECoG, & DDPM & \cite{eeg-2} & Classification & S & R & \checkmark &  &  & \checkmark &  &  \\
 & GAN & EEG & Conditional WGAN & \cite{eeg-6} & Classification & A & R &  &  &  &  &  &  \\ \hline
Other Conditioning   & DM & STFT spectrograms & DiffEEG & \cite{eeg-3} & Classification & A & R & \checkmark &  &  &  &  &  \\ \hline
\end{tabular}
\caption{Overview of all summarized and discussed studies on \textbf{EEG} synthesis with information about the technology utilized, the downstream task (DT), the evaluation procedure and code availability. Evaluation is split according to the evaluation metrics reported, covering utility (U), fidelity (F), diversity (D), qualitative assessment (Q), clinical utility (C) and privacy (P) evaluation. In evaluation of utility (U), R stands for real data, S stands for synthetic data, and A stands for augmented data. * stands for qualitative evaluation consisting of visualization of some plots only.Acronyms: Large Language Model (LLM), causal recurrent VAE (CRVAE), temporal GAN (TGAN)}

\label{tab:EEG}
\end{table*}

\subsection{Medical imaging}
\label{sec:images}
In this section, we investigate the synthesis of medical image data across various modalities essential for diagnostic imaging. Medical imaging plays a crucial role in diagnosis, treatment planning, and monitoring of various medical conditions. Our exploration delves into the synthesis techniques tailored to each imaging modality, ranging from dermatoscopy (also known as dermoscopy) \ref{sec-derma}, mammography \ref{sec-mammo}, ultrasound \ref{sec-us}, to MRI \ref{sec-mri}, CT scans \ref{sec-ct}, X-rays \ref{sec-xray}, OCTs \ref{sec-oct}, and multiple modalities \ref{sec-multi}. Through the analysis of synthesis methods and evaluation strategies, we aim to enhance understanding of synthetic medical image data generation and its applications in healthcare research.

\subsubsection{Dermatoscopy } \label{sec-derma}

\begin{table*}[h!]
\scriptsize
\begin{tabular}{ccllllcccccccc}
\hline
\multirow{3}{*}{\textbf{Application}} & \multirow{3}{*}{\textbf{Type}} & \multirow{3}{*}{\textbf{Modality}} & \multirow{3}{*}{\textbf{Technology}} & \multirow{3}{*}{\textbf{Paper}} & \multirow{3}{*}{\textbf{DT}} & \multicolumn{7}{c}{\textbf{Evaluation}} & \multirow{3}{*}{\textbf{Code}} \\ \cline{7-13}
 &  &  &  &  &  & \multicolumn{2}{c}{U} & \multirow{2}{*}{F} & \multirow{2}{*}{D} & \multirow{2}{*}{Q} & \multirow{2}{*}{C} & \multirow{2}{*}{P} &  \\ \cline{7-8}
 &  &  &  &  &  & Train & Test &  &  &  &  &  &  \\ \hline
\multirow{4}{*}{Unconditional} & \multirow{4}{*}{GAN} &  & PGGAN & \cite{derm-11} & Classification & A & R & \checkmark &  &  &  &  &  \\
 &  & Rosacea & StyleGAN2-ADA & \cite{derm-8} & None &  &  & \checkmark &  & \checkmark &  &  & \checkmark \\
 &  &  & StyleGAN2 & \cite{derm-1} & Classification & S & R & \checkmark &  &  &  &  &  \\
 &  &  & SLA-StyleGAN & \cite{derm-5} & Classification & \multicolumn{2}{c}{\checkmark} & \checkmark &  &  &  &  &  \\ \hline
\multirow{3}{*}{Class or Label Cond.} & \multirow{3}{*}{GAN} &  & \multirow{2}{*}{StyleGAN2-ADA} & \cite{derm-13} & Classification & S,A & R & \checkmark &  &  &  &  & \checkmark \\
 &  &  &  & \cite{derm-7} & Classification & S,A & R & \checkmark & \checkmark & \checkmark &  & \checkmark & \checkmark \\
 &  &  & cGAN & \cite{derm-12} & Classification & A & R &  &  &  &  &  &  \\ \hline
\multirow{3}{*}{Text-guided} & \multirow{3}{*}{DM} &  &  \begin{tabular}[l]{@{}l@{}}LDM, Stable Diffusion,\\Fine tuned stable diffusion\end{tabular}   & \cite{derm-3} & Classification & \multicolumn{2}{c}{\checkmark} &  &  &  &  &  &  \\
 &  &  & LDM & \cite{derm-6} & Classification & R,S,A & R &  &  &  &  &  &  \\
 &  &  & DALL-E2 & \cite{derm-10} & Classification & \multicolumn{2}{c}{\checkmark} &  &  &  &  &  &  \\ \hline
Intra-modal trans. & GAN &Majority to Minority & CycleGAN & \cite{derm-4} & Classification & A & R & \checkmark &  &  &  &  &  \\ \hline
\end{tabular}
\caption{Overview of all summarized and discussed studies on \textbf{dermatology} synthesis with information about the technology utilized, the downstream task (DT), the evaluation procedure and code availability. Evaluation is split according to the evaluation metrics reported, covering utility (U), fidelity (F), diversity (D), qualitative assessment (Q), clinical utility (C) and privacy (P) evaluation. In evaluation of utility (U), R stands for real data, S stands for synthetic data, and A stands for augmented data }
\label{tab:derma}
\end{table*}

The research on generative models in the field of dermoscopic imaging involves GANs and diffusion models. GANs, including variants like PGGAN, StyleGAN2 and its adaptations , and CycleGAN, have been used for data augmentation, class imbalance correction, and improving classification performance across diverse open source datasets like ISIC, Fitzpatrick 17k, ISIC, and BCN10000. These datasets are valuable  resources for dermatology research, with Fitzpatrick 17k being annotated with Fitzpatrick skin types \cite{fitzpatrick;}, a six-point scale that categorizes skin based on color and reaction to sun exposure.
\cite{derm-11} utilized PGGAN to generate synthetic skin images and used them for data augmentation, outperforming common augmentation tools, like Unsupervised Data Augmentation for Consistency Training (UDA) \cite{unsupervised}, Random Eraser \cite{randomeraser}, and random data augmentation. Further GANs employed in the unconditional synthesis utilize adaptations of StyleGAN2 \cite{derm-8,derm-1,derm-5}. In \cite{derm-8}, the StyleGAN2 with adaptive discriminator augmentation (StyleGAN2-ADA) model is fine-tuned on a small limited Rosacea dataset. The synthetic data was qualitatively evaluated by specialist dermatologists.
In \cite{derm-1}, StyleGAN2 network together with the Differentiable Augmentation (DiffAugment) \cite{diffaugment} is used to balance the number of images per class and enhance the performance of the skin lesion images classifier.\cite{derm-5} considered an adaptation of the original StyleGAN, with modifications to enhance style control mechanisms, focusing on replicating specific features such as hair, color, and lesion shapes in the images. By introducing variability through noise, the model generated a diverse set of synthetic samples. 

StyleGAN2-ADA was also used in studies for conditional synthesis\cite{derm-13,derm-7}. \cite{derm-13} examined if the generated data makes the classification models more prone to biases, specifically those associated with specific artifacts (e.g.: hairs, frames, or rulers present in the image). The results show that while GANs tend to amplify prominent biases present in the training data, they can mitigate less common ones. The study also observed that models trained on a mix of real and synthetic data were more resilient to bias, maintaining consistent predictions even when biases were introduced into the data. \cite{derm-7} considered StyleGAN2-ADA in a federated learning setup. 

Recently, different variants of diffusion models such as DALL.E-2 and LDM-based Stable Diffusion are employed for generating high-quality synthetic images \cite{derm-3,derm-6,derm-10}, with a focus text-to-image synthesis applications.Specifically, \cite{derm-10} uses DALL.E-2 to generate synthetic images from different skin types using structured text prompts specifying the type of medical condition, the location of the condition, the skin color of the patient, as well as the gender. This work show that the generated synthetic data can improve the classification of skin disease especially for underrepresented Fitzpatrick skin types. \cite{derm-3} fine-tuned the Stable Diffusion on a small number of training images from Fitzpatrick 17k using DreamBooth \cite{dreambooth} method to generate images of skin diseases conditioned on text. The synthetic data was used for augmenting training data for an disease classification problem, and showed that it can mitigate class imbalance for under-represented skin-types and improve the classification result, outperforming basic image transformations (comprising random flipping, cropping, rotating, warping, and lighting changes).

\cite{derm-4} employed CycleGANs to tackle the challenge of class imbalance in skin disease classifiers by generating synthetic images. Given the seven classes of images in HAM10000, they introduced a novel training strategy for the generators, taking the  majority class images and pairing them with each of the 6 minority classes separately to facilitate the generation of minority class images, enhancing the classification accuracy of the downstream models. 

While majority of papers evaluate the fidelity and utility of the generated data, only two papers \cite{derm-8,derm-7}  conduct a qualitative assessment, with  \cite{derm-7} utilizing 4 experts, 2 of which are dermatologist and 2 are DL experts, while \cite{derm-8} employ three dermatologist, which raise questions about the validity and the statistical significance of such qualitative assessment. Moreover, only one paper \cite{derm-7} consider privacy aspect of the generative model and the generated data. To assess the authenticity of the generative model, the authors project real dataset samples into the latent space of the generator and measure the similarity between pairs of real and generated images. This same paper measures diversity of the generated images by calculating the recall of the visual Turing test.


\subsubsection{Mammography} \label{sec-mammo}

\begin{table*}[h!]
\scriptsize
\centering
\begin{tabular}{ccllllcccccccc}
\hline
\multirow{3}{*}{\textbf{Application}} & \multirow{3}{*}{\textbf{Type}} & \multirow{3}{*}{\textbf{Modality}} & \multirow{3}{*}{\textbf{Technology}} & \multirow{3}{*}{\textbf{Paper}} & \multirow{3}{*}{\textbf{DT}} & \multicolumn{7}{c}{\textbf{Evaluation}} & \multirow{3}{*}{\textbf{Code}} \\ \cline{7-13}
 &  &  &  &  &  & \multicolumn{2}{c}{U} & \multirow{2}{*}{F} & \multirow{2}{*}{D} & \multirow{2}{*}{Q} & \multirow{2}{*}{C} & \multirow{2}{*}{P} &  \\ \cline{7-8}
 &  &  &  &  &  & Train & Test &  &  &  &  &  &  \\ \hline
Unconditional & GAN & patch-based & DCGAN, WGAN-GP & \cite{mamm-14} &class & A & R &  &  &  &  &  &  \\ \hline
\multirow{2}{*}{Class or Label Cond. } & \multirow{2}{*}{GAN} & DFM & cGAN & \cite{mamm-1} & class & \multicolumn{2}{c}{\checkmark} &  &  &  &  &  &  \\
 &  & ROI-based & ROImammoGAN & \cite{mamm-12} & None &  &  & \checkmark &  &  &  &  &  \\ \hline
Text-guided & DM & FFDM & FineTuned StableDiffusion & \cite{mamm-4} & None &  &  & \checkmark &  & \checkmark &  &  & \checkmark \\ \hline
\multirow{7}{*}{Intra-modal trans.} & \multirow{7}{*}{GAN} & Low-to-high density FFDM & CycleGAN & \cite{mamm-6} & det & A & R & \checkmark &  & \checkmark &  &  & \checkmark \\
 &  & Single-view $\rightarrow$ Two-view & CR-GAN & \cite{mamm-7} & None &  &  & \checkmark &  &  &  &  & \checkmark \\
 &  & \multirow{2}{*}{Opposite breast synthesis} & \multirow{2}{*}{Pix2Pix} & \cite{mamm-8} & det & \multicolumn{2}{c}{\checkmark} & \checkmark &  &  &  &  &  \\
 &  &  &  & \cite{mamm-15} & det & R,S,A & R &  &  &  &  &  &  \\
 &  & Lesion mask & TMP-GAN & \cite{mamm-10} & det & \multicolumn{2}{c}{\checkmark} &  &  &  &  &  &  \\
 &  & DFM $\rightarrow$ FFDM & HRGAN,  based on CycleGAN & \cite{mamm-11} & Seg,calc & \multicolumn{2}{c}{\checkmark} &  &  &  &  &  &  \\
 &  & Seg masks $\rightarrow$ mass images & DCGAN, InfillingGAN & \cite{mamm-18} & det & \multicolumn{2}{c}{\checkmark} & \checkmark &  &  &  &  & * \\ \hline
\end{tabular}
\caption{Overview of all summarized and discussed studies on \textbf{mammography} synthesis with information about the technology utilized, the downstream task (DT), the evaluation procedure and code availability. Evaluation is split according to the evaluation metrics reported, covering utility (U), fidelity (F), diversity (D), qualitative assessment (Q), clinical utility (C) and privacy (P) evaluation. In evaluation of utility (U), R stands for real data, S stands for synthetic data, and A stands for augmented data. * stands for availability of code for original method only. Acronyms: Region of interest (ROI),detection (det), segmentation (seg), calcification (calc), classification (class) conditional GAN (cGAN), complete representation GAN (CR-GAN), texture-constrained multichannel progressive GAN (TMP-GAN), high resolution GAN (HRGAN) }
\label{tab:mammo}
\end{table*}

The development of AI software products to enhance breast screening outcomes relies heavily on access to well-curated images. Over time, mammography technology has undergone significant evolution, transitioning from Digitized Screen-film Mammograms (DFMs) to Full-Field Digital Mammography (FFDM) and Digital Breast Tomosynthesis (DBT), thus revolutionizing breast cancer screening practices. Initially, DFMs involved X-ray images captured on film and later digitized for analysis. However, the introduction of FFDM marked a crucial advancement by allowing direct digital acquisition of breast images, eliminating the need for film processing and enabling immediate image availability. FFDM offered numerous benefits, including improved image quality, faster acquisition times, and enhanced visualization capabilities for radiologists.
Building upon the advantages of FFDM, DBT emerged as a groundbreaking technique in breast imaging. DBT captures multiple X-ray images from different angles, reconstructing three-dimensional breast tissue images. This innovative approach overcomes the limitations of traditional mammography by reducing tissue overlap and providing clearer, more detailed images, especially in dense breast tissue, but at the expense of higher radiation dose.
Aside from imaging techniques, the mammographic view plays a crucial role in breast cancer detection. While two-view mammography (mediolateral-oblique and craniocaudal views) is the current standard for breast cancer screening, single-view mammography remains in use in certain regions, particularly for screening specific age groups of women. However, the effectiveness of detecting cancer is comparatively lower with single-view mammography than with the two-view approach, as the breast cancer may be visually obscured by overlapping mammary glands in two-dimensional mammograms \cite{single-view-mammo}. 
Moreover, breast density poses an additional challenge in mammography and breast cancer detection. Dense breasts can obscure or mimic masses, making cancer detection more challenging. Additionally, extremely dense breasts are associated with a higher risk of cancer compared to low-density breasts \cite{dense-breast}. 

Intra-model translation within the mammography modality emerges as a promising approach to address some of these challenges. One primary hurdle is the scarcity of labeled FFDMs. Despite FFDMs becoming the preferred screening method over DFMs, datasets predominantly remain in DFM format. To overcome this, \cite{mamm-11} proposes a method to convert DFMs into FFDMs. They enhance the CycleGAN with advanced techniques like Pair with Constraint (PWC) training and gradient map input for discriminators, aiming to bridge the gap between available datasets and the preferred FFDM format. To overcome the limitation of using single-view mammograms over two-view mammograms in cancer detection, \cite{mamm-7} develops a method to generate two-view mammograms from single-view images, enhancing cancer detection capabilities.  To mitigate the low detection accuracy of dense breasts, \cite{mamm-6} employs CycleGANs to enhance mass detection in dense breasts. They generate synthetic dense mammograms from low-density counterparts, augmenting training data to improve detection accuracy. Through these methods, researchers aim to enhance the effectiveness of DL algorithms in breast cancer screening.

Generative models also serve various purposes in synthetic mammogram generation, including unconditional, class conditioned, and text conditioned synthesis. Researchers utilize diverse public datasets containing different lesion types and cases (normal, benign, and malignant). Examples include CBIS-DDSM \cite{cbis} and MIAS\cite{mias} for DFMs, and OPTIMAM \cite{optimam} and INBREAST \cite{inbreast} for FFDMs. \cite{mamm-14} use DCGAN to generate non-healthy mammogram patches containing both malignant and benign lesion. To overcome mode collapse in DCGAN and to increase training stability, the authors substitute DCGAN’s binary cross-entropy loss with a Wasserstein distance based loss function, ending up with WGAN-GP based training. The generated lesions patches are used to augment training data for patch classification models in different clinical centers, simulating the idea of sharing generative models instead of private data.
\cite{mamm-1} and \cite{mamm-12} utilize GANs for class conditional synthesis, with  \cite{mamm-1} generating DFMs with diverse conditions (normal, benign, and malignant), while \cite{mamm-12} generate region of interest (ROI) based digital mammograms with different abnormalities. Fine-tuned stable diffusion model was used for text conditioned healthy FFDM generation, specifying the description of the mammogram including view, breast density, breast area and vendor, in addition to a stable diffusion in-painting model used to generate synthetic lesions in desired regions of the normal mammogram \cite{mamm-4}.  

Evaluation of synthetic mammography data mainly focuses on utility and fidelity, with limited consideration for clinical validation, privacy, diversity metrics, and comparative analyses between models.  Qualitative assessment is performed in two papers, with \cite{mamm-6} employing a reader study involving two breast radiologists and one surgical oncologist while \cite{mamm-4} involved only one radiologist to determine whether the synthetic images were distinguishable from the real ones. The limited number of experts involved in both studies raise concerns about the validity and statistical significance of the qualitative assessment.

\subsubsection{Ultrasound} \label{sec-us}

\begin{table*}[h!]
\scriptsize
\centering
\begin{tabular}{ccllllcccccccc}
\hline
\multirow{3}{*}{\textbf{Application}} & \multirow{3}{*}{\textbf{Type}} & \multirow{3}{*}{\textbf{Modality}} & \multirow{3}{*}{\textbf{Technology}} & \multirow{3}{*}{\textbf{Paper}} & \multirow{3}{*}{\textbf{DT}} & \multicolumn{7}{c}{\textbf{Evaluation}} & \multirow{3}{*}{\textbf{Code}} \\ \cline{7-13}
 &  &  &  &  &  & \multicolumn{2}{c}{U} & \multirow{2}{*}{F} & \multirow{2}{*}{D} & \multirow{2}{*}{Q} & \multirow{2}{*}{C} & \multirow{2}{*}{P} &  \\ \cline{7-8}
 &  &  &  &  &  & Train & Test &  &  &  &  &  &  \\ \hline
\multirow{6}{*}{Unconditional} & \textbf{H:}DM,GAN & brain fetal & DSR-GAN,TB-GAN & \cite{us-18} & None &  &  & \checkmark &  &  &  &  & \checkmark \\
 & \textbf{H:}GAN,VAE & thyroid & Improved α-WGAN-GP & \cite{us-6} & Seg & A & R & \checkmark &  &  &  &  &  \\
 & \multirow{4}{*}{GAN} &  fetal brain& Stylegan2-ada & \cite{us-13} & Class & \multicolumn{2}{c}{\checkmark} & \checkmark &  &  &  &  & \checkmark \\
 &  & breast & StyleGAN2 variants & \cite{us-11} & Class & A & R & \checkmark &  &  &  &  &  \\
 &  & liver & StackGAN & \cite{us-14} & Class & A & R & \checkmark &  & * &  &  &  \\
 &  & Breast & TripleGAN & \cite{us-16} & Class & \multicolumn{2}{c}{\checkmark} &  &  & \checkmark (2) &  &  &  \\ \hline
\multirow{3}{*}{Class or Label Cond.} & \multirow{2}{*}{GAN} & Muscles & GAN-CA & \cite{us-10} & Class & \multicolumn{2}{c}{\checkmark} &  &  &  &  &  & \checkmark \\
 &  & nodule mask $\rightarrow$ US & Phased GAN & \cite{us-4} & Class & A & R &  &  &  &  &  &  \\
 & DM & \begin{tabular}[l]{@{}l@{}}cardiac anatomical\\semantic label maps\end{tabular} & \begin{tabular}[l]{@{}l@{}}specifically semantic\\DDPM\end{tabular} 
 & \cite{us-7} & Seg & S & R &  &  &  &  &  & \checkmark \\ \hline
\multirow{4}{*}{Intra-modal trans.} & \multirow{4}{*}{GAN} & \multirow{2}{*}{B-mode $\rightarrow$ EUS} & U-net based gen & \cite{us-19} & Class & \multicolumn{2}{c}{\checkmark} & \checkmark &  & \checkmark(2) &  &  & \checkmark \\
 &  &  & AUE-net:Pix2Pix based & \cite{us-3} & None &  &  & \checkmark &  & * &  &  &  \\
 &  & FC sketch $\rightarrow$ FC views & PSFFGAN & \cite{us-2} & None &  &  & \checkmark &  & \checkmark(2) &  &  &  \\
 &  & \begin{tabular}[l]{@{}l@{}}lung, hip joint, and ovary-\\sketch guided\end{tabular}  & PGGAN & \cite{us-17} & Seg & A & R & \checkmark &  &  &  &  & \checkmark \\ \hline
\multirow{5}{*}{Inter-modal trans.} & VAE & T2 MR $\rightarrow$ iUS & MHVAE & \cite{us-23} & None &  &  & \checkmark &  &  &  &  & \checkmark \\
 & \multirow{4}{*}{GAN} & pMR   $\rightarrow$ iUS & ApGAN & \cite{us-22} & Class & A & R & \checkmark &  & \checkmark(5) & \checkmark &  &  \\
 &  & \multirow{2}{*}{CT $\rightarrow$ US} & MR-CycleGAN & \cite{us-20} & Seg & \multicolumn{2}{c}{\checkmark} &  &  &  &  &  &  \\
 &  &  & CycleGAN & \cite{us-21} & Seg & \multicolumn{2}{c}{\checkmark} & \checkmark &  &  &  &  &  \\
 &  & CT AMs $\rightarrow$ US & 3D Pix2Pix & \cite{us-1} & Seg & S,A & R &  &  & \checkmark &  &  &  \\ \hline
\end{tabular}

\caption{Overview of all summarized and discussed studies on \textbf{Ultrasound} synthesis with information about the technology utilized, the downstream task (DT), the evaluation procedure and code availability. Evaluation is split according to the evaluation metrics reported, covering utility (U), fidelity (F), diversity (D), qualitative assessment (Q), clinical utility (C) and privacy (P) evaluation. In the Type column, H stands for hybrid approach combining two different models. In evaluation of utility (U), R stands for real data, S stands for synthetic data, and A stands for augmented data. In qualitative evaluation, the number between parenthesis indicate the number of specialists employed, while * stands for visual inspection of visualization plots. Acronyms: generator (gen),Multi-Modal Hierarchical Latent Representation VAE (MHVAE), anatomy preserving GAN (ApGAN), Pseudo-Siamese Feature Fusion Generative Adversarial Network (PSFFGAN), GAN with coordinate attention mechanism (GAN-CA), Diffusion-Super-Resolution-GAN (DSR-GAN), Brightness mode (B-mode), intra-operative US (iUS),Elastography Ultrasound (EUS), four-chamber (FC), anatomical model (AM), preoperative magnetic resonance (pMR), Transformer-based-GAN (TB-GAN), StyleGAN2 variants: StyleGAN2 ADA, StyleGAN2 DiffAug}

\end{table*}


Ultrasound synthesis using generative models spans various organs, including fetal brain, thyroid, breast, liver, muscles, and more, and different ultrasound techniques, such as Brightness Mode (B-mode) ultrasound and Elastography Ultrasound (EUS). B-mode ultrasound, commonly known as conventional ultrasound, is a widely used imaging technique in medical diagnostics. It generates 2D grayscale images of soft tissues, offering detailed anatomical information in a non-invasive manner. Despite its safety and relative comfort for patients, B-mode ultrasound has limitations in accurately assessing tissue stiffness and distinguishing between various types of lesions or abnormalities. To address these limitations, EUS is employed as an advanced technique. It evaluates tissue stiffness or elasticity by measuring the response of tissue to compression or shear waves, providing additional functional information beyond traditional B-mode ultrasound. Elastography often displays tissue stiffness using a color scale. This makes it valuable for assessing tissue pathology and differentiating between benign and malignant lesions. However, EUS may require specialized equipment and training, and its accuracy can be affected by factors such as operator technique and tissue depth \cite{sigrist2017ultrasound}. 

To generate the synthetic ultrasound under different applications of synthesis, various generative models are employed.
For instance, in unconditional synthesis, a hybrid approach combining diffusion models and GANs was adopted in \cite{us-18} to synthesize fetal brain images. This method, known as DSR-GAN, comprises a DDPM followed by a Super-Resolution-GAN. Additionally, StyleSwin, a Transformer-based GAN (TB-GAN), was utilized in this study.The results indicated that the DSR-GAN outperformed the TB-GAN in terms of FID, showcasing the potential of using hybrid generative models  for improved image synthesis. Similarly, \cite{us-6} employed a hybrid approach, combining GANs and VAEs to generate synthetic thyroid ultrasound images. Similar to applications in dermoscopy imaging, different variants of StyleGAN2, like StyleGAN2-ADA and DiffAug, are employed in \cite{us-11} and \cite{us-13} to address the issue of limited training data. To generate class-conditioned images, \cite{us-13} train a generator for each class separately. In \cite{us-14}, a two-stage stacked GAN architecture was introduced for synthesizing realistic B-mode liver ultrasound images, with the second stage responsible for enhancing tissue details.

Models focusing on class and label-conditioned synthesis also exist, with \cite{us-10} utilizing different classes of the idiopathic inflammatory myopathies disease  and \cite{us-4,us-7} conditioning on semantic label maps. In \cite{us-10}, a GAN with a coordinate attention mechanism (GAN-CA) synthesizes muscle ultrasound images, focusing on disease features rather than muscle contour information by segmenting the data into subsets based on the muscle site. Meanwhile, \cite{us-7} generates cardiac synthetic data guided by cardiac anatomical semantic label maps. In breast ultrasound classification, some cases are considered to be hard, as they contain overlapping benign and malignant nodules with mixed characteristics of both. Addressing the challenge of classifying hard cases, \cite{us-4} introduces a phased GAN that consists of one generator and two discriminators to generate  complex cases. The first discriminator help the generator to generate normal samples, while the second discriminator help the generation of hard samples. The generated hard samples enhance the classifiers' ability to classify them.

Intra-modal translation is a common application, with techniques mainly employing GANs to translate sketches into images or B-mode to EUS. To make the structural details of generated images more realistic, \cite{us-17} introduce auxiliary sketch guidance into a cGAN, and adopt a PGGAN training strategy to generate high-resolution images. Inspired by the conditional GAN and by integrating the optimization strategy of WGAN-GP, \cite{us-2} propose a model to synthesizes fetal four-chamber (FC) views from given sketch images. The generator of the model consists of US image encoder, sketch image decoder, and a US image decoder, while the discriminator is based on PatchGAN \cite{patchgan}.
Generating EUS from conventional B-mode ultrasound, as in \cite{us-19,us-3},  offers a promising solution to address the challenges associated with capturing EUS directly. Because the aim of elasticity generation is to accurately evaluate the degree of elasticity of the tumor, \cite{us-19} utilize a local tumor discriminator to determine whether the tumor area is real or fake in addition to the global discriminator used to classify whether the input image is real or fake. Moreover, because the elastography image mainly relies on color to distinguish tissue elasticity , \cite{us-19} utilize  color re-balancing module to re-weight L1 loss during training based on the color rarity.\cite{us-3} employs an Pix2Pix based architecture combined with various attention modules and color loss features to generate EUS images from conventional ultrasound images.

Inter-modal translation plays a significant role in US synthesis, encompassing translations from MR  or CT  scans to US images. In some cases during surgery, obtaining high-quality intra-operative US (iUS) images with clear anatomical structures can be difficult due to various factors such as tissue deformation, blood flow, and surgical instruments. To address the lack of iUS data, researchers propose generating iUS images using Preoperative Magnetic Resonance (pMR) scans. These approach utilizes various techniques, including VAEs and popular GAN models such as CycleGAN and Pix2Pix. \cite{us-23} introduce Multi-Modal Hierarchical Latent Representation VAE (MHVAE), the first multi-modal VAE approach with a hierarchical latent representation for unified medical image synthesis, mainly focusing on translating MR to iUS. \cite{us-22}  proposed an anatomy preserving GAN (ApGAN) to generate intra-operative ultrasound (Sim-iUS) of liver from pMR. In \cite{us-1}, 3D Pix2Pix GANs were trained using high-resolution anatomical models derived from CT scans as input.

The evaluation of synthetic ultrasound images predominantly focuses on utility, fidelity assessment, and qualitative analysis, with a few studies incorporating clinical validation. For instance, in \cite{us-22}, clinical assessment encompassed various anatomical structures and focal areas to convince physicians of the authenticity of the synthetic data, including unique features like annular structures of the gallbladder or kidney and other relevant distances. As part of the qualitative assessment, several studies employed visual Turing tests, as seen in \cite{us-16,us-19,us-2,us-22,us-12}. These tests typically involved specialists evaluating the realism of the synthetic images, with a limitation that the majority of studies employ two specialists only, while \cite{us-22} employed five specialists for a more comprehensive assessment. Moreover, the fidelity and utility of the synthetic data of many proposed models were compared against state-of-the-art synthesis methods, including off-the-shelf models like CycleGAN, Pix2Pix, Interpretable Representation Learning by Information Maximizing Generative Adversarial Nets (InfoGAN) \cite{infogan}, GAN, and DCGAN, as well as traditional augmentation methods. Comparisons in most cases are limited by the fact that off-the-shelf models are not optimized for specific medical modalities.

\subsubsection{MRI} \label{sec-mri}
MRI has become a ubiquitous medical imaging tool owing to its remarkable capability to generate a plethora of contrast mechanisms in soft tissues using specific pulse sequences. While acquiring images of the same anatomy with different contrasts offer a comprehensive diagnostic information, acquisition constraints, scan time limitations, inherent noise and artifacts may compromise the integrity of acquired data. In these scenarios, the ability to synthesize missing or corrupted contrast through computational algorithms holds significant promise for enhancing diagnostic utility. 

Within the application of intra-modality translation, various generative models convert images from one appearance (source domain) to another (target domain). In many cases, variants of CycleGAN and Pix2Pix models are widely used for unpaired and paired translations, respectively. The authors show promising results for translating the contrast of one MRI sequence to another in both brain \cite{la_rosa_mprage_2021,qin_style_2022,zhu_dualmmp-gan_2022,fei_deep_2021, huo_brain_2022} and cardiac MRI \cite{dhaene_myocardial_2023}. \cite{chen_diverse_2021} proposes a GANs-based approach for inter-modal translation from Brain CT to MRI and cardiac MRI to CT to develop a segmentation model on the target domain without labels.

Several studies have tackled the issue of missing or corrupted contrasts in multi-contrast brain MRI by using different generative models for the task of contrast-to-contrast translation. The goal of the image synthesis is to generate a missing contrast from other available images, for instance generating a T2-weighted image from T1-weighted image. Most proposed methods rely on GANs with modifications to handle multiple image sources and combine complementary features from images of the same anatomy with different contrasts. \cite{yurt_mustgan_2021} propose a multi-stream GAN approach, named mustGAN that aggregates information across multiple source images and effectively utilizes both the shared and complementary features of multiple source images through a combination of multiple one-to-one streams and a joint many-to-one stream, resulting in superior performance in multi-contrast MRI synthesis compared to existing methods. Similarly, \cite{moshe_handling_2023} explored the use of the Pix2Pix GAN model for generating multi-modal brain MRI data, aimed at enhancing lesion segmentation and classification tasks, demonstrating that using generated images yielded significantly better performance compared to simply replacing missing images with blank or duplicate ones.

\cite{fan_tr-gan_2022} addresses the challenge of predicting Alzheimer's disease progression from 3D brain MRI data by proposing a Temporal Recurrent GAN (TR-GAN). The study aims to enable multi-session prediction and synthesis of missing longitudinal data, allowing for a more accurate observation of Alzheimer's disease progression. The TR-GAN model employs recurrent connections to generate future sessions with variable lengths and incorporates specific modules to encourage the generation of high-quality MRI data. The study demonstrates the effectiveness of TR-GAN in generating future MRI sessions and outperforms other GAN architectures in image quality evaluation metrics such as MSE, MS-SSIM, and PSNR.
\cite{huang_common_2022} introduces a 3D context-aware generative adversarial network (CoCa-GAN) for synthesizing multi-modality glioma MRI images. It maps input modalities into a shared feature space to synthesize the target modality and complete tumor segmentation. The method's contributions include utilizing a common feature space for encoding multi-modality information, employing separate encoders for decomposing each modality into high-level features, and implementing tumor attention for learning lesion-specific representations.

Label-conditioned generation, where the generative model uses segmentation masks as input, has shown promise for creating images in various MRI modalities. One of the key goals is to address medical data scarcity when training downstream DL models, for instance generating images with labels for the development of a DL-based organ segmentation. This approach offers two advantages: increased control over image generation by conditioning of the anatomy present in the labels and the ability to directly use synthetic data with ground truth labels to augment supervised segmentation tasks. Several frameworks based on the SPADE GAN \cite{park2019semantic} architecture have been developed for generating cardiac MRI images, specifically aimed at tackling data scarcity in multi-vendor settings and augmenting existing datasets for building generalizable segmentation networks \cite{amirrajab_label-informed_2022, al_khalil_usability_2023, lustermans_optimized_2022,thermos_controllable_2021,amirrajab_pathology_2022}. The authors found that incorporating anatomical variations, such as changes in heart shape during synthesis of cardiac MRI data and tailoring the generation to the downstream task significantly reduced segmentation failures \cite{al_khalil_reducing_2023} and led to the development of a more robust segmentation model \cite{lustermans_optimized_2022}. In \cite{thermos_controllable_2021}, a novel framework is introduced for disentangling images into spatial anatomy factors and corresponding imaging representations to enable controllable image synthesis. This framework allows combining anatomical factors to create new plausible heart anatomies in synthesized cardiac MR images. The authors propose the concept of a disentangled anatomy arithmetic GAN (DAA-GAN) in the generative model and assess the impact of data augmentation on classification and segmentation tasks. They demonstrate that synthetic data augmentation not only enhances learning on balanced data but also improves model performance on underrepresented classes in both disease classification and segmentation tasks.

While few studies have explored diffusion models, Med-DDPM \cite{dorjsembe_conditional_2023} and Brain-SPADE \cite{fernandez_can_2022} present methods for generating 3D semantic medical images using diffusion models. \cite{fernandez_can_2022} demonstrates that the segmentation models trained on synthetic data can perform comparable to those trained on real data, even generalizing to unseen data distributions. However, despite the realistic appearance of Med-DDPM's generated images, the study found that advanced data augmentation using nnU-Net \cite{isensee2021nnu} led to better results, suggesting a gap between real and synthetic data quality.

Researchers have also explored unconditional image generation for various modalities \cite{li_image_2021, dorjsembe_three-dimensional_2022,dar_investigating_2023}, as well as generation conditioned on additional information like age, gender, brain volume, and time frame \cite{pinaya_brain_2022,yoon_sadm_2023}. In \cite{pinaya_brain_2022}, an LDM is proposed to generate synthetic MRI images of the adult human brain, conditioned on factors such as age, gender, ventricular volume, and brain volume. The realism of the synthetic data is evaluated using FID score and diversity with the MS-SSIM for 1000 synthetic samples. The method's age conditional ability is verified in a brain age prediction task, and the volume conditional ability is analyzed by an external model, demonstrating a high correlation between the obtained volume and the inputted value. Conditional LDMs are also applied in \cite{r-173}  for multi-sequence and multi-contrast generation of prostate MR images, conditioned on both text and images. The synthesized quality is evaluated by a clinician to identify generated pathology, and a model for lesion identification is developed on the augmented data with synthetic examples.

Despite generating high quality images, few studies adequately evaluate the usefulness of synthetic data for downstream tasks like improving DL based segmentation or classification models. Many studies solely focus on qualitative evaluation of the synthesis results without usability studies \cite{fan_tr-gan_2022,hu_domain-adaptive_2022,  jiang_cola-diff_2023, meng_novel_2022, zhan_lr-cgan_2021}, highlighting the need for further research to assess the usability and clinical applicability of these generated images. Similarly, there is limited evaluation of image diversity and privacy preserving characteristics of the synthetic MRI data.

\begin{table*}[h!]
\scriptsize
\centering

\begin{tabular}{ccllllcccccccc}
\hline
\multirow{3}{*}{\textbf{Application}} & \multirow{3}{*}{\textbf{Type}} & \multirow{3}{*}{\textbf{Modality}} & \multirow{3}{*}{\textbf{Technology}} & \multirow{3}{*}{\textbf{Paper}} & \multirow{3}{*}{\textbf{DT}} & \multicolumn{7}{c}{\textbf{Evaluation}} & \multirow{3}{*}{\textbf{Code}} \\ \cline{7-13}
 &  &  &  &  &  & \multicolumn{2}{c}{U} & \multirow{2}{*}{F} & \multirow{2}{*}{D} & \multirow{2}{*}{Q} & \multirow{2}{*}{C} & \multirow{2}{*}{P} &  \\ \cline{7-8}
 &  &  &  &  &  & Train & Test &  &  &  &  &  &  \\ \hline
\multirow{3}{*}{Unconditional} & GANs & Brain   & Pix2Pix, WGANGP & \cite{li_image_2021} & Age est. & R, A & R &  &  &  &  &  & \checkmark \\
 & \multirow{2}{*}{DMs} & Brain   T1 & 3D-DDPM & \cite{dorjsembe_three-dimensional_2022} & None &  &  & \checkmark &  & \checkmark &  &  & \checkmark \\
 &  & Knee  , PCCTA & LDM & \cite{dar_investigating_2023} & None &  &  &  &  &  &  & \checkmark &  \\ \hline
\multirow{11}{*}{Class or Label Cond.} & \multirow{7}{*}{GANs} & Cardiac   cine, label & SPADE GAN & \cite{al_khalil_usability_2023} & Seg. & R, S, A & R &  &  &  &  &  & \checkmark \\
 &  & Cardiac   cine, label & SDNet & \cite{thermos_controllable_2021} & None &  &  & \checkmark &  &  &  &  &  \\
 &  & Breast   & Pix2PixHD & \cite{teixeira_adversarial_2021} & Seg. &  &  & \checkmark &  &  &  &  &  \\
 &  & Cardiac   LGE, label & SPADE GAN & \cite{lustermans_optimized_2022} & Seg. & R, A & R &  &  &  &  &  & \checkmark \\
 &  & Cardiac   cine, label & SPADE GAN & \cite{amirrajab_label-informed_2022} & Seg. & R, S, A & R & \checkmark &  & \checkmark &  &  & \checkmark \\
 &  & Cardiac   cine, label & SPADE GAN & \cite{al_khalil_reducing_2023} & Seg. & R, A & R &  &  & \checkmark & \checkmark &  & \checkmark \\
 & VAE & Brain lesion & Progressive VAE & \cite{huo_brain_2022} & Seg. & R, A & R & \checkmark &  & \checkmark  &  &  &  \\
 & VAEs, GANs & Cardiac   cine, label & VAE, SPADE GANs & \cite{amirrajab_pathology_2022} & Seg. & R, A & R &  &  & \checkmark & \checkmark &  & \checkmark \\
 & DMs & Brain   T1, label & MED-DDPM & \cite{dorjsembe_conditional_2023} & Seg. & R, S, A & R & \checkmark &  & \checkmark &  &  & \checkmark \\
 & DMs, GANs & Brain   T1, FLAIR, label & brainSPADE & \cite{fernandez_can_2022} & Seg. & R, S, A & R &  &  &  &  &  &  \\ \hline
\multirow{2}{*}{Attribute Cond} & DMs & Brain, Cardiac   & SADM & \cite{yoon_sadm_2023} & None &  &  & \checkmark &  &  &  &  & \checkmark \\
 & LDM & Brain   & CLDM & \cite{pinaya_brain_2022} & Class. &  &  & \checkmark & \checkmark &  &  &  &  \\ \hline
\multirow{18}{*}{Intra-modal trans.} & \multirow{10}{*}{GANs} & Brain   T1$\rightarrow$T2, T2$\rightarrow$T1 & ST-cGAN & \cite{qin_style_2022} & None &  &  & \checkmark &  &  &  &  &  \\
 &  & Brain   missing contrast & CoCa-GAN & \cite{huang_common_2022} & Seg. & R, A & R & \checkmark &  & \checkmark &  &  &  \\
 &  & Brain   missing contrast & LR-cGAN & \cite{zhan_lr-cgan_2021} & None &  &  & \checkmark &  &  &  &  &  \\
 &  & Brain   missing contrast & MustGAN & \cite{yurt_mustgan_2021} & None &  &  & \checkmark &  & \checkmark &  &  & \checkmark \\
 &  & Brain   missing sessions & TR-GAN & \cite{fan_tr-gan_2022} & Class. & R, S, A & R & \checkmark &  &  &  &  &  \\
 &  & Brain   different sequences & Pix2Pix & \cite{la_rosa_mprage_2021} & Seg. & R, A & R & \checkmark &  & \checkmark &  &  &  \\
 &  & Brain   missing contrast & DualMMP-GAN & \cite{zhu_dualmmp-gan_2022} & Seg. & R, S, A & R & \checkmark &  &  &  &  &  \\
 &  & Brain   missing contrast & Pix2Pix-GAN & \cite{moshe_handling_2023} & Class. & R, A & R & \checkmark &  & \checkmark &  &  &  \\
 &  & Brain   T1, T2, T2c$\rightarrow$FLAIR & encoder-GANs & \cite{fei_deep_2021} & None &  &  & \checkmark &  &  &  &  &  \\
 &  & Cardiac   cine$\rightarrow$tagged & CycleGAN & \cite{dhaene_myocardial_2023} & Seg. & R, S & R &  &  &  &  &  &  \\
 & VAE & Brain      FLAIR,T2$\rightarrow$T1 & VAE & \cite{hu_domain-adaptive_2022} & None &  &  & \checkmark &  &  &  &  & \checkmark \\
 & \multirow{4}{*}{DMs} & Brain   slice generation & DDPM & \cite{peng_generating_2023} & None &  &  & \checkmark &  &  &  &  & \checkmark \\
 &  & Cardiac   ED$\rightarrow$ES phases & DDPM & \cite{kim_diffusion_2022} & None &  &  & \checkmark &  &  &  &  & \checkmark \\
 &  & Brain   with tumor & DDPM & \cite{wolleb_swiss_2022} & None &  &  &  &  &  &  &  &  \\
 & \multirow{3}{*}{LDM} & Brain   missing contrast & CoLa-Diff , LDM & \cite{jiang_cola-diff_2023} & None &  &  & \checkmark &  & \checkmark &  &  &  \\
 &  & Brain SWI$\rightarrow$MRA, T1$\rightarrow$T2 & LDM & \cite{zhu_make--volume_2023} & None &  &  & \checkmark &  & \checkmark &  &  &  \\
 &  & Brain   missing contrast & LDM & \cite{meng_novel_2022} & None &  &  & \checkmark &  &  &  &  &  \\ \hline
Inter-modal trans. & GANs &\begin{tabular}[l]{@{}l@{}} Brain CT$\rightarrow$MRI,\\ Cardiac   MRI$\rightarrow$CT \end{tabular} & DDA-GAN & \cite{chen_diverse_2021} & Seg. & R, A & R &  &  &  &  &  &  \\
\hline
\end{tabular}

\caption{Overview of all summarized and discussed studies on \textbf{MRI} synthesis with information about the technology utilized, the downstream task, the evaluation procedure and code availability. Evaluation is split according to the evaluation metrics reported, covering utility (U), fidelity (F), diversity (D), visual quality (Q), clinical utility (C) and privacy (P) evaluation. Missing contrast generation can include any combination of available source(s) of contrast to target contrast, such as T1, T2 $\rightarrow$ FLAIR, T1, PD $\rightarrow$ T2, FLAIR, T2 $\rightarrow$T1, etc. }
\label{tab:MRI}
\end{table*}

\subsubsection{CT} \label{sec-ct}

Radiation therapy relies on CT as the primary imaging tool for treatment planning, offering precise tissue geometry visualization and electron density conversion crucial for dose calculations \cite{seco2006assessing}. Although MRI complements CT by providing superior soft-tissue contrast without ionizing radiation, it lacks the electron density data necessary for accurate dose calculations \cite{nyholm2009systematisation}. In addition, cone-beam CT is commonly utilized for patient positioning and monitoring before, during, or after dose delivery \cite{jaffray2012image}. Integrating information from both modalities involves registering MRI to CT, a process prone to systematic errors affecting treatment accuracy \cite{nyholm2009systematisation, ulin2010results}. MR-only radiotherapy seeks to overcome registration issues \cite{nyholm2014counterpoint}, yet the absence of tissue attenuation data in MRI requires methods to convert MR data for precise dose calculations equivalent to CT.
Similarly, various techniques aim to enhance CBCT quality by generating synthetic CT from alternative imaging modalities \cite{taasti2020developments}. Despite its significance in image-guided adaptive radiation therapy, CBCT faces limited usage due to scatter noise and truncated projections, leading to reconstruction challenges and artifacts like shading, streaking, and cupping \cite{zhu2009noise}. Consequently, CBCT is rarely employed for online plan adaptation. Conversion of CBCT to CT could enable precise dose computation and enhance the quality of image-guided adaptive radiation therapy for patients. In addition, efforts to reduce radiation dose in CT imaging have gained significant attention due to concerns about patient exposure \cite{antypas2011comprehensive}. Various methods have been explored, including adjustments to tube current, voltage, and x-ray intensity, as well as protocol revisions. However, these approaches often compromise image quality, impacting diagnostic accuracy. Consequently, recent focus has shifted towards developing low-dose CT (LDCT) restoration algorithms using diverse DL techniques.

As such, most proposed approaches focus on image-to-image translation for synthetic CT generation (Table \ref{tab:ct}), mostly from MR images. Some methods extend to generating CBCT or low-dose CT images. While these typically require extensive paired or unpaired MRI and CT data, unsupervised methods are still rare. However, the emergence of diffusion-based models has inspired interest in unsupervised and text-to-image approaches, offering potential for realistic tissue structure generation without the dependency on source images.

Generative methods play a crucial role in synthesizing realistic CT scans, frequently employed for data augmentation across various tasks like segmentation and classification. Using a PGGAN, \cite{ct-02} augments data for segmenting prostate and organs in 3D pelvic CT data. Building on this, \cite{ct-11} employs PGGANs for unsupervised synthesis of high-resolution CT scans, yielding outputs that are often indistinguishable from real CT scans, as demonstrated in a comprehensive Visual Turing test study. However, challenges persist in accurately generating smaller anatomical structures, indicating limitations in current methods due to homogeneous data training. Addressing this, \cite{ct-04} introduces a hybrid model using a DDPM with a Swin-transformer-based network for abdomen and pelvic CT scan generation, achieving state-of-the-art performance, in terms of synthetic image quality and diversity, compared to existing generative models such as WGAN, PGGAN, SGAN and DDPM. Challenges exist in extending this approach to 3D synthesis due to complexity and resource-intensive requirements. 

Numerous applications in CT imaging revolve around generating synthetic lung and chest CT images, such as \cite{ct-l-02} and \cite{ct-l-19}, utilizing a 3D CT-GAN, enabling a gradual generation of CT volumes by generating sub-component slices and slabs, reducing the GPU memory requirements. Interest in clinical applications of unsupervised and/or unconditional generative methods has expanded to include COVID-19 detection \cite{ct-l-17, ct-l-18, ct-l-20}. \cite{ct-l-17} presents a 3D cycleGAN framework tailored for generating non-contrast CT images from contrast-enhanced ones, leveraging a sizable dataset of more than 1000 COVID-19 scans. In \cite{ct-l-20}, the authors explore COVID-19 detection in CT scans using vision transformers, augmented with synthetic images from SAGAN-ResNet. \cite{ct-l-18} delves into the challenges of employing large diffusion models, proposing a lightweight variant with a U-Net architecture and step information injection, stressing the importance of specific operations in encoder-decoder models to maintain synthetic image quality.

Recent advancements in exploring conditional models have emphasized the injection of greater control into the generation process, resulting in the production of higher quality synthetic images. \cite{ct-03} leverages a VAE to compress image space, and then conditions a latent diffusion model on its output to synthesize CT images for the denoising of LDCT images. \cite{ct-38} introduces MedGen3D, a novel diffusion-based generative framework, which utilizes conditioning on mask sequences generated by a Multi-Condition Diffusion Probabilistic Model (MC-DPM), considering both the position or direction in which sequences are generates (forward or backward), as well as the sequence content. These conditions guide the synthesis of realistic medical images aligned with masks while maintaining spatial consistency across adjacent slices.

Conditioning on label maps has been thoroughly explored to guide the accurate anatomical synthesis of tissues in chest and brain \cite{ct-05}, liver \cite{ct-08}, as well as lung CT \cite{ct-l-01, ct-l-03, ct-l-04, ct-l-11, ct-l-16}. \cite{ct-05} benchmarks variations of PGGAN and StyleGAN for synthetic medical image generation, revealing insights into their performance under variable conditions, such as the variation in label combinations, sample size and spatial resolution. Reducing class numbers improves synthetic data quality, but caution is needed to avoid overfitting on rare classes, while refining label conditioning mechanisms is crucial for enhancing overall predictive performance.  \cite{ct-08} presents a 3D GAN framework for liver CT image synthesis using vascular segmentation masks and region-based weight-balancing with a stable Multiple Gradient Descent Algorithm (MGDA). While promising, challenges remain in accurately labeling vascular structures, underscoring the need for more explicit information for precise 3D liver image synthesis. Variations of the Pix2Pix model are commonly used for synthesizing lung images with diverse pathological cases \cite{ct-l-01}, generating free-form lesion images from tumor sketches using StyleGAN concepts for accurate style translation \cite{ct-l-16}, and lung nodule synthesis \cite{ct-l-03} for data augmentation, employing window-guided semantic learning to control the generation of smaller semantic features. \cite{ct-l-04} investigates InfoGAN for generating shape-controlled tumor images and evaluates their potential for histological classification of lung cancer with CT scans. InfoGAN outperforms WGAN in controlling tumor size and chest wall presence, improving image features. However, both GANs still fall short of fully capturing the entire distribution of real images, likely due to differences in generated tumor shapes. \cite{ct-l-11} explores Semantic Diffusion Models (SDM) for generating high-quality pulmonary CT images compared to other SOTA approaches, such as SPADE GANs, while prioritizing lung nodule detection. 

Text-to-image synthesis is gaining traction for diversifying data augmentation, offering text-controlled generation of synthetic images to address challenges like rare case simulation and privacy concerns by creating images without the need for sharing sensitive or identifiable data, typically sourced from real images. 
\cite{ct-l-13} utilizes radiology reports for high-resolution 3D medical image synthesis, starting with low-resolution synthesis guided by tokenized reports and up-scaled using a super-resolution module. \cite{ct-12} introduces a text-conditional CT generation method, combining a pre-trained language model, transformer-based 3D chest CT generation, and a super-resolution diffusion model. Limitations include the lack of benchmarks, reliance on 2D super-resolution, and considerations for model generalization, applicability to non-chest tissues, and clinical relevance evaluation.

\begin{table*}[h!]
\scriptsize
\begin{tabular}{ccllllcccccccc}
\hline
\multirow{3}{*}{\textbf{Application}} & \multirow{3}{*}{\textbf{Type}} & \multirow{3}{*}{\textbf{Modality}} & \multirow{3}{*}{\textbf{Technology}} & \multirow{3}{*}{\textbf{Paper}} & \multirow{3}{*}{\textbf{DT}} & \multicolumn{7}{c}{\textbf{Evaluation}} & \multirow{3}{*}{\textbf{Code}} \\ \cline{7-13}
 &  &  &  &  &  & \multicolumn{2}{c}{\textbf{U}} & \multirow{2}{*}{\textbf{F}} & \multirow{2}{*}{\textbf{D}} & \multirow{2}{*}{\textbf{Q}} & \multirow{2}{*}{\textbf{C}} & \multirow{2}{*}{\textbf{P}} &  \\ \cline{7-8}
 &  &  &  &  &  & Train & Test &  &  &  &  &  &  \\ \hline
\multirow{8}{*}{Unconditional} & \multirow{6}{*}{GAN} & Pelvic CT & PGGAN & \cite{ct-02} & Seg. & R,S & R &  &  &  & \checkmark &  &  \\
 &  & Whole body CT & PGGAN & \cite{ct-11} & None &  &  &  &  & \checkmark &  &  &  \\
 &  & \multirow{2}{*}{Lung CT} & CT-SGAN with BiLSTM & \cite{ct-l-02} & Det. & R,S,A & R & \checkmark &  &  &  &  &  \\
 &  &  & 3D CT-GAN & \cite{ct-l-19} & Class. & R,A & R & \checkmark &  & \checkmark &  &  &  \\
 &  & \multirow{2}{*}{COVID19} & SA-GAN ResNet & \cite{ct-l-20}& Det. & R,A & R &  &  &  &  &  & \checkmark \\
 &  &  & 3D patch-based cycle-GAN & \cite{ct-l-17}& Class. & R & R, S & \checkmark &  & \checkmark &  &  &  \\
 & \multirow{2}{*}{DM} & Abd. \& pelvic CT & MT-DDPM & \cite{ct-04} & Class. & R,S,A & R & \checkmark & \checkmark & \checkmark &  &  &  \\
 &  & COVID19 & U-Net DM & \cite{ct-l-18} & None &  &  & \checkmark &  &  &  &  &  \\ \hline
\multirow{7}{*}{Class or Label Cond.} & \multirow{6}{*}{GAN} & Brain CT & PGGAN, cpd-GAN & \cite{ct-05}& Class. & R,S & R & \checkmark &  & \checkmark &  & \checkmark & \checkmark \\
 &  & Liver CT & U-Net patch-GAN & \cite{ct-08} & None &  &  & \checkmark & \checkmark &  & \checkmark &  &  \\
 &  & \multirow{5}{*}{Lung CT} & Pix2Pix & \cite{ct-l-01} & None &  &  & \checkmark &  & \checkmark &  &  &  \\
 &  &  & Info GAN & \cite{ct-l-04} & Class.* & R,S & R & &  &  &  &  &  \\
 &  &  & Style Pix2Pix & \cite{ct-l-16} & Class. & R,S,A & R & \checkmark & \checkmark &  &  &  &  \\
 &  &  & MGGAN + Pix2Pix + WGSLN & \cite{ct-l-03} & Class. & R,A & R & \checkmark &  &  & \checkmark &  &  \\
 & DM &  & SDM & \cite{ct-l-11} & Det. & R,A & R & \checkmark &  &  &  &  &  \\ \hline
\multirow{2}{*}{Attribute Cond.} & \multirow{2}{*}{DM} & Abd. \& head CT & CLDM & \cite{ct-03} & None &  &  & \checkmark &  &  &  &  &  \\
 &  & Thoracic CT & Cond. DPM + Semantic DM & \cite{ct-38} & Seg.* & R,S,A & R & \checkmark & \checkmark &  &  &  &  \\ \hline
\multirow{2}{*}{Text-guided} & \multirow{1}{*}{DM}  & Lung CT & BERT + cond. DDPM & \cite{ct-l-13} & None &  &  & \checkmark &  &  & \checkmark &  & \checkmark \\
 & Hybrid & Chest CT & LLM, ViT, transf.-MaskGT & \cite{ct-12} & Class. & R,S,A & R & \checkmark &  &  &  &  & \checkmark \\ \hline
\multirow{10}{*}{Intra-modal trans.} & \multirow{8}{*}{GAN} & CBCT $\rightarrow$ CT & R2ACNN & \cite{ct-10} & None &  &  & \checkmark &  &  & \checkmark &  &  \\
 &  & CT $\rightarrow $CTA & DCT-GAN & \cite{ct-13} & None &  &  & \checkmark &  &  &  &  & \checkmark \\
 &  & NDCT $\rightarrow$ LDCT & GAN-NETL & \cite{ct-01} & None &  &  & \checkmark &  &  & \checkmark &  & \checkmark \\
 &  & dCT $\rightarrow$ CT & Pix2Pix & \cite{ct-20} & None &  &  & \checkmark &  &  & \checkmark &  & \checkmark \\
 &  & CBCT $\rightarrow$ CT & Attention GAN & \cite{ct-30} & None &  &  & \checkmark &  &  & \checkmark &  &  \\
 &  & \multirow{3}{*}{COVID19} & StarGANv2 & \cite{ct-l-06} & Seg. & R,A & R &  &  &  &  &  &  \\
 &  &  & sRD-GAN & \cite{ct-l-14} & None &  &  & \checkmark & \checkmark &  \checkmark & \checkmark &  &  \\
 &  &  & Cycle GAN & \cite{ct-l-15} & Seg. & R,S & R &  &  &  & \checkmark &  &  \\
 & DM & CBCT $\rightarrow$ CT & Cond. DDPM & \cite{ct-16} & None &  &  & \checkmark &  &  & \checkmark &  &  \\
 & Transf. & insp. $\rightarrow$ exp. CT & SWIN & \cite{ct-l-12} & None &  &  & \checkmark & \checkmark &  & \checkmark &  &  \\ \hline
\multirow{18}{*}{Inter-modal trans.} & \multirow{13}{*}{GAN} & \multirow{11}{*}{MR $\rightarrow$ CT} & CTF GAN (Pix2Pix) & \cite{ct-17} & Seg. & A & R & \checkmark &  &  &  &  &  \\
 &  &  & 3D Pix2Pix & \cite{ct-22} & None &  &  & &  &  & \checkmark &  &  \\
 &  &  & Pix2Pix + StyleGAN & \cite{ct-23} & None &  &  & \checkmark &  &  &  &  &  \\
 &  &  & Multi-cycle GAN & \cite{ct-27} & None &  &  & \checkmark &  &  &  &  &  \\
 &  &  & U-Net cycleGAN & \cite{ct-29} & None &  &  &  \checkmark &  &  & \checkmark &  &  \\
 &  &  & Cycle GAN + attention & \cite{ct-32} & None &  &  & \checkmark  & & \checkmark  &  &  &  \\
 &  &  & Cycle GAN & \cite{ct-39}& None &  &  & \checkmark &  & \checkmark &  &  &  \\
 &  &  & 3D Cycle GAN & \cite{ct-40} & None &  &  &  &  &  & \checkmark &  &  \\
 &  &  & Pix2Pix & \cite{ct-41} & None &  &  & \checkmark &  &  & \checkmark &  &  \\
 &  &  & Cycle GAN, structGAN, Flow CGAN & \cite{ct-42} & None &  &  &  &  &  & \checkmark &  &  \\
 &  &  & U-Net-based WGAN & \cite{ct-l-05} & None &  &  & \checkmark &  &  & \checkmark &  &  \\
 &  & PET/MRI $\rightarrow$ CT & Patch GAN & \cite{ct-28} & None &  &  & \checkmark & \checkmark &  &  &  &  \\
 &  & US $\rightarrow$ CT & cGAN + SRGAN & \cite{ct-33} & None &  &  & \checkmark &  &  & \checkmark &  &  \\
 & \multirow{3}{*}{DM} & \multirow{5}{*}{MR $\rightarrow$ CT} & SynDiff & \cite{ct-06} & None &  &  & \checkmark &  &  &  &  & \checkmark \\
 &  &  & DDPM + SDE & \cite{ct-18} & None &  &  & \checkmark & &  &  &  &  \\
 &  &  & DDMM & \cite{ct-19} & None &  &  & \checkmark &  &  &  &  &  \\
 & \multirow{2}{*}{Hybrid} &  & Swin-VNET & \cite{ct-24} & None &  &  & \checkmark &  &  & \checkmark &  &  \\
 &  &  & RTCGAN-CNN/transf. & \cite{ct-26} & None &  &  & \checkmark &  & \checkmark & \checkmark &  &  \\ \hline
\end{tabular}
\caption{Overview of all summarized and discussed studies on \textbf{CT} synthesis with information about the technology utilized, the downstream task (DT), the evaluation procedure and code availability. Evaluation is split according to the evaluation metrics reported, covering utility (U), whereby automated machine learning models are trained with real (R), synthetic (S) or augmented with synthetic data (A), and tested on real or synthetic data, fidelity (F), diversity (D), visual quality (Q), clinical utility (C) and privacy (P) evaluation. * stands for studies that have utilized synthesized images for pre-training.  }
\label{tab:ct}
\end{table*}

Applications of generative models for LDCT synthesis include \cite{ct-01}, which introduces a GAN with Noise Encoding Transfer Learning (GAN-NETL) to generate paired Normal-dose CT (NDCT) and LDCT datasets, addressing variations in LDCT noise across scanners. For MRI-only radiotherapy, \cite{ct-17} proposes joint synthesis and segmentation using a Pix2Pix GAN, transferring semantic information for anatomical structure correction in synthetic CT images. Additionally, \cite{ct-22} extends the Pix2Pix generator for 3D CT synthesis from T1-weighted MR images within a patch-based conditional GAN, showing high correlation with real CT.

To tackle the lack of paired data, unpaired synthesis with cycle GANs has become popular across various tasks, including generating synthetic abdominal and pelvic CT \cite{ct-02, ct-04}, prostate \cite{ct-39}, lumbar spine \cite{ct-40} in the realm of inter-modal translation, and COVID-19 \cite{ct-l-16,ct-l-06, ct-l-15} in intra-modal translation applications. To address structural consistency issues between MRI and CT images inherent in cycle GAN-based synthesis for inter-modal translation, \cite{ct-27} propose a multi-cycle GAN with a pseudo-cycle consistent module and domain control module. This ensures high-quality and realistic CT image generation. Similarly, \cite{ct-42} introduce augmented cycle consistent GANs, injecting structural information and optical flow consistency constraints. \cite{ct-32} present a paired-unpaired unsupervised attention-guided GAN, combining Wasserstein GAN adversarial loss with content and L1 losses, capturing fine structures and improving global consistency. Additionally, to mitigate spatial inconsistencies often generated by 2D cycle GANs in synthetic 3D CT images, \cite{ct-29} employ a double U-Net cycle GAN for 2.5D synthesis.

Cone beam CT (CBCT) images are crucial for image-guided therapy but often suffer from low contrast and artifacts, impacting adaptive radiotherapy dose calculations. Synthetic CT addresses these challenges, with \cite{ct-30} using an attention-guided GAN for unpaired low-dose CBCT to CT synthesis, emphasizing artifact correction due to X-ray scatter and respiratory movements. Similarly, \cite{ct-10} introduces a sequence-aware contrastive generative network to enhance CBCT image quality. 
Other approaches include \cite{ct-13} using a cascaded GAN for synthesizing plain CT images into CTA images, \cite{ct-14} applying styleGAN2 with a tailored loss function for breast CT images, and \cite{ct-23} manipulating styleGAN2's latent space for diverse CT image generation.

Intra- and inter-modality translation models, like \cite{ct-06}, propose a cycle-consistent architecture for diffusion-based synthesis between CT and MRI scans. Similarly, \cite{ct-18} explores diffusion and score-matching models for CT-MRI image conversion, showcasing superior image quality with faithful anatomical details and avoiding over-smoothing and artifacts. Additionally, \cite{ct-19} introduces a denoising diffusion model for cross-modal medical image synthesis, allowing for adjustment of CT projection numbers and enhancing CT image fidelity. 
Meanwhile, \cite{ct-24} proposes a method for generating CT scans from MRI data, employing a transformer-based DDPM with a shifter-window transformer network for the diffusion process, facilitating the synthesis of synthetic CT from MRI.

While many studies use fidelity metrics, comprehensive evaluations of newly proposed models, especially those employing diffusion or transformer-based approaches, are lacking  \cite{ct-l-18, ct-03, ct-38, ct-l-11, ct-18, ct-24}. The practical utility of generated images in downstream tasks is often overlooked, raising concerns about clinical applicability. A crucial need for realistic synthetic medical images is highlighted, emphasizing the importance of a visual Turing test with radiologists \cite{ct-04, ct-l-01, ct-32}. However, these tests often involve few participants and limited samples, introducing biases, and challenges, particularly in assessing 2D slices representing 3D data.

Examining model limitations, studies like \cite{ct-l-19,ct-l-01} reveal the presence of artificial features and challenges in the realistic synthesis of certain anatomical regions, such as the thoracoabdominal junction. \cite{ct-l-19} suggests modifications to the visual Turing test and stresses involving multiple radiologists for robust evaluations. Insights from \cite{ct-05} indicate challenges in discriminating between real and synthetic images at lower spatial resolutions, underlining the importance of image quality and label accuracy in realistic synthesis. Interestingly, classification accuracy improves with higher spatial resolution due to detailed features and visual artifacts, aligning with similar findings. These considerations emphasize ongoing challenges in evaluating synthetic medical images, highlighting the need for a standardized approach in assessing generative models in clinical contexts.

Metrics like FID and MMD, though effective for natural images, may overlook crucial anatomical details in medical images, as evidenced by the small gap between methods. A clinically relevant evaluation is thus pivotal in introducing synthetic images into the medical field. CT synthetic images are primarily assessed by calculating differences in Hounsfield units per organ \cite{ct-02, ct-l-03, ct-10, ct-30, ct-16, ct-28, ct-33, ct-39, ct-24,ct-26}, crucial for reflecting tissue density on CT scans. Given CT's significance in radiation oncology, dose distribution comparison is common \cite{ct-10, ct-20, ct-30, ct-22, ct-39, ct-41, ct-l-05, ct-33}. Additional studies evaluate organ contour quality and target volumes for planning and dose calculation \cite{ct-08, ct-30,ct-41, ct-33,  ct-26}. A subset of studies delves into clinically-relevant anatomical measurements from synthetic images, showcasing diverse applications. \cite{ct-l-12} calculates biomarkers for air trapping and emphysema, \cite{ct-22} compares skull properties for transcranial ultrasound procedures, and \cite{ct-40} assesses errors in lumbar spine synthetic CT images. \cite{ct-l-13} conducts an in-depth analysis of diffusion-based text-guided image synthesis, evaluating anatomical preservation in lung vessels and airways. This method, utilizing specific input prompts from real radiology reports, enhances the evaluation of text-based models, considering clinical indices like pleural effusion and cardiothoracic ratio. Additionally, \cite{ct-28} examines attenuation correction in PET images from synthetic CT, compared to current MRI methods. 

These methods offer insights into generative model effectiveness for clinical use, surpassing traditional metrics. However, integrating clinical information into the generation process remains uncommon, and evaluating privacy preservation with synthetic data is seldom done. Notably, \cite{ct-05} stands out as one of the few studies employing nearest neighbor matching and cosine distance to demonstrate that GAN models may inherently offer privacy through stochastic gradient descent. Nevertheless, the study emphasizes the imperative need to formulate and quantify privacy guarantees.

Lastly, a crucial approach in \cite{ct-15} uniquely advocates for employing synthetic data in AI model validation, utilizing it as a part of the training process to prevent overfitting and facilitate model selection. The study introduces synthetic tumors in an extensive validation dataset, incorporated into a continual learning framework. Models trained and validated on dynamically expanding synthetic data consistently outperform those relying solely on real-world data for tumor segmentation. The findings highlight the efficacy of synthetic data in mitigating overfitting and supporting early cancer detection, suggesting its potential for large-scale testing, especially in scenarios where obtaining real data is challenging, as is often the case in clinical settings.

\subsubsection{X-Ray}  \label{sec-xray}

Common applications of X-ray image synthesis (Table \ref{tab:xray}) involve generating images with pathological features like lesions in Chest X-rays (CXR). Lesions vary in shape and size, posing challenges for detection, especially in deep learning methods reliant on extensive training data. Generating X-ray images of COVID-19 patients has also attracted significant interest, with various approaches using different conditional and unconditional methods to enhance the diversity of the COVID-19 patient population, leading to more accurate detection. Consequently, translating between X-rays of normal cases and those showing indications of pneumonia or tumors is frequently performed to improve the performance of anomaly detection algorithms.

Common methods for unconditional synthesis of X-ray images involve WGAN-GPs \cite{x-08, x-09}, DCGANs \cite{x-09, x-31}, and PGGANs \cite{x-31, x-26, x-30}, known for high-resolution synthetic image generation. \cite{x-08} pioneers WGANs for realistic knee joint X-rays, validated by 15 medical experts, showing notable realism and improved osteoarthritis severity classification accuracy. 
\cite{x-09} notes DCGAN's instability during COVID-19 CXR generation. \cite{x-15} stabilizes training with multi-scale gradient flow and self-attention mechanisms. \cite{x-22} explores co-evolutionary learning, and \cite{x-18} leverages a PGGAN for chest X-ray synthesis, useful in cardiomegaly \cite{x-21} and abnormality detection \cite{x-30}. 
While \cite{x-08} and \cite{x-31} investigate the utility of synthetic images for osteoarthritis severity and abnormality classification across various training scenarios, other approaches mainly focus on augmenting with synthetic data alone \cite{x-09, x-30}. \cite{x-21} suggests synthetic data for cardiothoracic ratio calculation in cardiomegaly. Despite using FID scores \cite{x-09, x-31, x-15, x-22},  studies often lack detailed downstream task analysis \cite{x-08, x-09, x-30, x-18, x-21}. \cite{x-31} attempts downstream task analysis, but the evaluation lacks detail, raising concerns about practical applicability in clinical settings.

Recent studies exploring conditional approaches for data generation include employing a DDPM-based model with shared latent noise for semantic consistency in synthesizing image/label pairs \cite{x-24}, as well as latent class optimization for synthesis of diverse pathologies \cite{x-26}. In parallel, \cite{x-32} synthesizes pathology by translating disease-containing medical images into disease-free ones using a cycleGAN and a domain label as a conditional input. In the evaluation process, both \cite{x-24} and \cite{x-26}extensively assess synthetic images, including a thorough examination in downstream segmentation tasks, such as semantic and instance-aware segmentation, for both in-domain and out-of-domain samples. Furthermore, \cite{x-26} stands out as one of the first studies to delve into a comprehensive analysis of the applicability of FID for quality assessment, showcasing its effectiveness in distinguishing not only between various pathologies and normal X-rays but also among different pathologies themselves.

Some studies extend the use of label-conditioned synthetic methods beyond basic label utilization, broadening data generation capabilities. For instance, \cite{x-12} uses a DCGAN-based shape generator with size modulation for precise control of lung nodule shape and diameter. Incorporating PGGAN enhances visual plausibility, adopting a coarse-to-fine generation approach. Notably, they demonstrate that deliberate selection of challenging examples improves augmentation significantly. In a different approach, \cite{x-17} employs StyleGAN2 and Bayesian image reconstruction for ROI-conditioned synthesis of CXR images. It aims to generate synthetic data while preserving clinically relevant features, proposing a potential data sharing strategy for privacy protection. Additionally, \cite{x-20} merges PGGAN with Pix2PixHD GAN to facilitate multi-stage generation from dots, representing various anatomical parts and serving as seeds for subsequent steps that produce high-resolution images. These methods, tailored for specific augmentation tasks, highlight the importance of designing synthesis procedures around downstream challenges, leading to better-performing models and higher-quality generated images.

In the realm of medical text-to-image synthesis, latent diffusion models, as described in \cite{x-04}, employ strategies like reduced image sizes conditioned on textual inputs and incorporate reverse diffusion steps to enhance denoising. Notably, \cite{x-19} introduces semantic diffusion, which involves translating latent variables into image space using a decoder and refining/upscaling through a super-resolution diffusion process. Models aiming to align generated reports with X-ray images, as discussed in \cite{x-01, x-02, x-06}, encounter challenges such as false positives \cite{x-04, x-05}. Joint learning of language embeddings and image generation reveals limitations, prompting \cite{x-19} to propose a cascaded latent diffusion model that includes an autoencoder component. Despite the advancements, comprehensive evaluations of image quality and clinical applicability, as highlighted in studies like \cite{x-19, x-01, x-02}, are still lacking. Furthermore, the potential utility of these models in data augmentation remains largely unexplored, and there is a paucity of comparative analyses against other state-of-the-art methodologies. 

\begin{table*}[h!]
\scriptsize
\begin{tabular}{ccllllcccccccc}
\hline
\multirow{3}{*}{\textbf{Application}} & \multirow{3}{*}{\textbf{Type}} & \multirow{3}{*}{\textbf{Modality}} & \multirow{3}{*}{\textbf{Technology}} & \multirow{3}{*}{\textbf{Paper}} & \multirow{3}{*}{\textbf{DT}} & \multicolumn{7}{c}{\textbf{Evaluation}} & \multirow{3}{*}{\textbf{Code}} \\ \cline{7-13}
 &  &  &  &  &  & \multicolumn{2}{c}{\textbf{U}} & \multirow{2}{*}{\textbf{F}} & \multirow{2}{*}{\textbf{D}} & \multirow{2}{*}{\textbf{Q}} & \multirow{2}{*}{\textbf{C}} & \multirow{2}{*}{\textbf{P}} &  \\ \cline{7-8}
 &  &  &  &  &  & Train & Test &  &  &  &  &  &  \\ \hline
\multirow{8}{*}{Unconditional} & \multirow{8}{*}{GAN} & Knee joint XR & WGAN & \cite{x-08} & Class.* & R,S,A & R & \checkmark &  & \checkmark &  & \checkmark &  \\
 &  & Pneumonia & DCGAN + PGGAN & \cite{x-31} & Class.* & A & R & \checkmark &  &  &  &  &  \\
 &  & \multirow{3}{*}{COVID19} & DCGAN  + WGAN-GP & \cite{x-09} & Class. & A & R & \checkmark &  &  &  &  &  \\
 &  &  & MSG-SAGAN & \cite{x-15} & None &  &  & \checkmark &  &  &  &  &  \\
 &  &  & MLPs, CNNs & \cite{x-22} & None &  &  & \checkmark &  &  &  &  & \checkmark \\
 &  & \multirow{3}{*}{CXR} & \multirow{3}{*}{PGGAN} & \cite{x-18} & None &  &  & \checkmark & \checkmark &  &  &  &  \\
 &  &  &  & \cite{x-30} & Class. & R,A & R &  &  & \checkmark &  &  &  \\
 &  &  &  & \cite{x-21} & Class. & R & S &  &  &  & \checkmark &  &  \\ \hline
\multirow{4}{*}{Class or Label Cond.} & \multirow{4}{*}{GAN} & \multirow{4}{*}{CXR} & DCGAN + patchGAN & \cite{x-12} & Det. & A & R & \checkmark &  &  &  &  & \checkmark \\
 &  &  & StyleGAN2 + Bayesian recon. & \cite{x-17} & Class.* & R,A & R & \checkmark &  &  &  &  &  \\
 &  &  & PGGAN + Pix2PixHD &  \cite{x-20} & Seg.* & R, S & R,S &  &  &  &  &  &  \\ \multicolumn{1}{l}{} & DM &  & DDPMs & \cite{x-24} & Seg. & A & R & \checkmark & \checkmark &  &  &  &  \\ \hline
\multicolumn{1}{l}{\multirow{2}{*}{Attribute conditioning}} & \multicolumn{1}{l}{\multirow{2}{*}{GAN}} & CXR & PGGAN with latent space optim. & \cite{x-26} & Class. & R,S & R & \checkmark &  & \checkmark &  &  &  \\
\multicolumn{1}{l}{} & \multicolumn{1}{l}{} & COVID19 & SD-GAN & \cite{x-32} & Class. & A & R & \checkmark & \checkmark &  &  &  & \checkmark \\
\hline
\multirow{6}{*}{Text-guided} & \multirow{2}{*}{DM} & CXR $\&$ Reports & Semantic DM \& SR diffusion & \cite{x-19} & None &  &  & \checkmark & \checkmark &  &  &  & \checkmark \\
 &  & CXR $\&$  Impressions & Stable diffusion v2 & \cite{x-04} & Det. &  & S & \checkmark & \checkmark &  &  &  & \checkmark \\
 & \multicolumn{1}{l}{\multirow{4}{*}{Hybrid}} & \multirow{4}{*}{CXR $\&$  Reports} & VQ-GAN + GPT-Neox LLM & \cite{x-02} & None &  &  & \checkmark & \checkmark &  & \checkmark &  &  \\
 & \multicolumn{1}{l}{} &  & SDM-VAE + CLIP text-enc. & \cite{x-05} & Class. & R,S,A & R & \checkmark & \checkmark &  & \checkmark &   &  \\
 & \multicolumn{1}{l}{} &  & VQ-GAN + bidirec. transf. & \cite{x-06} & Class. &  & S & \checkmark &  & \checkmark &  \checkmark &  & \checkmark \\
 & \multicolumn{1}{l}{} &  & Cycle GAN + LSTM & \cite{x-01} & None &  &  & \checkmark & \checkmark & \checkmark  &  &  &  \\ \hline
\multirow{6}{*}{Intra-modal trans.} & \multirow{7}{*}{GAN} & \multirow{3}{*}{Normal $\rightarrow$ Abn. CXR} & cond. GAN + spatial transf. & \cite{x-07} & Det. & A & R &  &  &  &  &  &  \\
 &  &  & Cycle GAN + classifier & \cite{x-23} & Class. & R,A & R &  &  &  &  &  &  \\

 &  & \multirow{2}{*}{Normal $\rightarrow$ COVID19} & Adaptive Cycle GAN & \cite{x-13}& None &  &  & \checkmark & & \checkmark &  &  &  \\
 &  &  & Cycle GAN & \cite{x-11} & Det.* & R,A & R &  &  &  &  &  &  \\
 &  & Urinary stones & U-Net cond. GAN & \cite{x-14} & Seg. & R,A & R & \checkmark &  &  &  &  &  \\
 &  & Normal $\rightarrow$ Pneumonia & Cond. + Cycle GAN & \cite{x-28} & Class. &  & S &  &  &  &  &  &  \\ \hline
Inter-modal trans. & Hybrid & Spine US $\rightarrow$ XR & DDPM with ViT & \cite{x-16} & None &  &  & \checkmark &  &  & \checkmark &  &  \\ \hline
\end{tabular}
\caption{Overview of all summarized and discussed studies on \textbf{X-Ray} (XR) synthesis with information about the technology utilized, the downstream task (DT), the evaluation procedure and code availability. Evaluation is split according to the evaluation metrics reported, covering utility (U), whereby automated machine learning models are trained with real (R), synthetic (S) or augmented with synthetic data (A), and tested on real or synthetic data, fidelity (F), diversity (D), visual quality (Q), clinical utility (C) and privacy (P) evaluation. * stands for studies that have utilized synthesized images for pre-training. CXR stands for Chest X-Ray. Papers that specify the downstream task without training data, signify instances where a pre-existing model was directly applied or tested on synthetic data. }
\label{tab:xray}
\end{table*}
In evaluating the quality of generated X-ray images, most studies use downstream tasks and generic fidelity and diversity metrics that potentially overlook nuances specific to medical imaging. However, a few studies, such as \cite{x-08}, conduct human perception quality and clinical usability evaluations. \cite{x-08} performed a Visual Turing test with specialists, revealing that synthetic X-ray images are often mistaken for real ones, suggesting sufficient realism for deceiving experts. Another study, \cite{x-30}, conducted an extensive Turing test with 400 images, demonstrating that radiologists struggled to distinguish between synthetic and real CXRs, especially in abnormal cases, containing lung lesions. In the context of CXRs, \cite{x-26} finds that synthetic data is often mistaken for real, but improved replication of fine anatomical details is required for truly indistinguishable samples. The study also evaluates FID's applicability to synthetic CXRs, noting that small-scale details may not be reproduced at the highest quality. \cite{x-06} conducts human evaluation with board-certified clinicians, revealing that the model generates realistic CXRs but lacks precise alignment with reports. The proposed method excels in generating view-specific X-rays but occasionally struggles with fine details. Furthermore, \cite{x-13} employs the Universal Image Quality Index (UIQI) and Visual Information Fidelity (VIF) to compare synthetic and real images. The quantitative metrics suggest challenges for GANs in translating high-quality synthetic images, possibly due to insufficient training samples. A few studies offer insights into the use of synthetic images for clinical tasks. For instance, \cite{x-06} employs synthetic images for a 14-class diagnosis of anomalies in X-ray images, \cite{x-21} calculates ratios to classify cardiomegaly in CXR images, and \cite{x-02} evaluates findings generated from synthetic images, identifying false positives and misdiagnoses. \cite{x-16} measures the Cobb angle difference between synthetic and original X-ray images, demonstrating effective use in evaluating scoliosis. Additionally, \cite{x-08} explores privacy concerns by assessing the nearest real image neighbors to synthetic images, used as criteria for selecting training data to ensure privacy in downstream tasks; however, a proper evaluation of preserved privacy is lacking.

\subsubsection{OCT}  \label{sec-oct}

\begin{table*}[]
\scriptsize
\begin{tabular}{ccllllcccccccc}

\hline
\multirow{3}{*}{\textbf{Application}} & \multirow{3}{*}{\textbf{Type}} & \multirow{3}{*}{\textbf{Modality}} & \multirow{3}{*}{\textbf{Technology}} & \multirow{3}{*}{\textbf{Paper}} & \multirow{3}{*}{\textbf{Downstream Task}} & \multicolumn{7}{c}{\textbf{Evaluation}} & \multirow{3}{*}{\textbf{Code}} \\ \cline{7-13}
 &  &  &  &  &  & \multicolumn{2}{c}{\textbf{U}} & \multirow{2}{*}{\textbf{F}} & \multirow{2}{*}{\textbf{D}} & \multirow{2}{*}{\textbf{Q}} & \multirow{2}{*}{\textbf{C}} & \multirow{2}{*}{\textbf{P}} &  \\ \cline{7-8}
 &  &  &  &  &  & Train & Test &  &  &  &  &  &  \\ \hline
\multirow{7}{*}{Unconditional} & \multirow{5}{*}{GAN} & AS OCT & PGGAN & \cite{oct-01} & Class. & R,S & R &  &  & $\checkmark$ &  &  &  \\
 &  & AS OCT & Cycle + patch GAN & \cite{oct-17} & Seg. & R,S & R & $\checkmark$ &  &  & $\checkmark$ &  &  \\
 &  & SD DM, NOR, AMD & Vanilla GAN & \cite{oct-02} & Class. & R,S,A & R & $\checkmark$ &  &  &  &  &  \\
 &  & Retinal OCT & StyleGAN2-ADA & \cite{oct-07} & Class.* & A & R &  &  &  &  &  &  \\
 &  & FH OCT & StyleGAN2-ADA & \cite{oct-08} & Class. & R,A & R &  &  & $\checkmark$ & $\checkmark$ &  & $\checkmark$ \\
 & DM & AS OCT & DDPM & \cite{oct-18} & Seg./Det. & R & S & $\checkmark$ &  &  &$\checkmark$  &  &  \\
 & Hybrid & Retinal OCT & DDPM + transf. & \cite{oct-24} & Seg. & A & R & $\checkmark$ &  &  &  &  &  \\ \hline
\multirow{6}{*}{\begin{tabular}[c]{@{}c@{}}Class or Label \\cond.\end{tabular}} & \multirow{5}{*}{GAN} & \multirow{2}{*}{Retinal OCT} & U-Net + patch GAN & \cite{oct-12} & Seg. & R & S & $\checkmark$ & $\checkmark$ & $\checkmark$ &  & $\checkmark$ & $\checkmark$ \\
 &  &  & Modified Pix2Pix & \cite{oct-25} & Class. & R,S,A & R & &  &  &  &  &  \\
 &  & Chorio-retinal OCT & Modified DCGAB &  \cite{oct-20} & Seg. & R,S,A & R & $\checkmark$ & \multicolumn{1}{l}{} & \multicolumn{1}{l}{} & \multicolumn{1}{l}{} & \multicolumn{1}{l}{} & \multicolumn{1}{l}{} \\
 &  & SD OCT & StyleGAN2 & \cite{oct-22} & Seg./Class. & R,S,A & R & $\checkmark$ & \multicolumn{1}{l}{} & \multicolumn{1}{l}{} & \multicolumn{1}{l}{} & \multicolumn{1}{l}{} & \multicolumn{1}{l}{} \\
 &  & Coronary atrial plaques & cond. GAN & \cite{oct-23} & Class. & A & R &  &  &  &  &  &  \\
 & VAE & Retinal (diseased) OCT & Modified cond. VAE & \cite{oct-19} & None &  &  &  &  &  &  &  & $\checkmark$ \\ \hline
\multicolumn{1}{l}{\multirow{6}{*}{Attribute cond.}} & \multirow{6}{*}{GAN} & SD OCT & BAABGAN & \cite{oct-03} & None &  &  & $\checkmark$ &  & $\checkmark$ & $\checkmark$ &  &  \\
\multicolumn{1}{l}{} &  & DME OCT & DDFA-GAN & \cite{oct-14} & Class. & A & R & $\checkmark$ & $\checkmark$ & \multicolumn{1}{l}{} & \multicolumn{1}{l}{} & \multicolumn{1}{l}{} & $\checkmark$ \\
\multicolumn{1}{l}{} &  & AMD OCT and CFP & Pix2PixHD & \cite{oct-09} & Class.* & A & R & \multicolumn{1}{l}{} & \multicolumn{1}{l}{} & \multicolumn{1}{l}{} & \multicolumn{1}{l}{} & \multicolumn{1}{l}{} & $\checkmark$ \\
\multicolumn{1}{l}{} &  & \multirow{3}{*}{Retinal OCT} & Contrastive CNN-based & \cite{oct-06} & None &  &  & \multicolumn{1}{l}{\checkmark} & \multicolumn{1}{l}{} & $\checkmark$ & $\checkmark$ & \multicolumn{1}{l}{} & \multicolumn{1}{l}{} \\
\multicolumn{1}{l}{} &  &  & FOF-GAN & \cite{oct-15} & Class. & R & S & \multicolumn{1}{l}{$\checkmark$} &  & \multicolumn{1}{l}{} & \multicolumn{1}{l}{} & \multicolumn{1}{l}{} & \multicolumn{1}{l}{} \\
\multicolumn{1}{l}{} &  &  & StarGAN & \cite{oct-05} & Seg. & R & S &  &  & $\checkmark$ & $\checkmark$ &  &  \\ \hline
\multirow{3}{*}{Intra-modal trans.} & \multirow{3}{*}{GAN} & Pre-T - Post-T OCT & Pix2PixHD & \cite{oct-10} & None &  &  &  &  & $\checkmark$ & $\checkmark$ &  &  \\
 &  & OCT - PS-OCT & Pix2PixHD & \cite{oct-13} & Class. & R,S & R & $\checkmark$ &  & $\checkmark$ &  &  &  \\
 &  & Inter-device SD-OCT & CycleGAN & \cite{oct-21} & Seg. & R & S &  &  &  & $\checkmark$ &  &  \\ \hline
\multirow{2}{*}{Inter-modal trans.} & \multirow{2}{*}{GAN} & FA - OCT & Pix2Pix & \cite{oct-04} & None &  &  & $\checkmark$ & $\checkmark$ & $\checkmark$ & $\checkmark$ &  &  \\
 &  & Fundus - OCT & U-net-based & \cite{oct-16} & Class.* & R,S,A & R & $\checkmark$ & \multicolumn{1}{l}{} & \multicolumn{1}{l}{} & \multicolumn{1}{l}{} & \multicolumn{1}{l}{} & \multicolumn{1}{l}{} \\ \hline
\end{tabular}
\caption{Overview of all summarized and discussed studies on \textbf{OCT} synthesis with information about the technology utilized, the downstream task (DT), the evaluation procedure and code availability. Evaluation is split according to the evaluation metrics reported, covering utility (U), whereby automated machine learning models are trained with real (R), synthetic (S) or augmented with synthetic data (A), and tested on real or synthetic data, fidelity (F), diversity (D), visual quality (Q), clinical utility (C) and privacy (P) evaluation. AS, SD, DM, NOR, AMD, FH, DME, CFP, Pre-T, Post-T, PS, FA stand for anterior segment, spectral domain, diabetic maculopathy, normal, age-related macular degeneration, foveal hypoplasia, diabetic macular edema, color fundus photography, pre-therapy, post-therapy, polarization sensitive OCT and fluorescein angiography, respectively. * stands for studies that have utilized synthesized images for pre-training. }
\label{tab: oct-table}
\end{table*}

Non-invasive retinal OCT is crucial for diagnosing eye diseases, as features like retinal layer shapes and thicknesses strongly correlate with conditions such as glaucoma, age-related macular degeneration, and diabetic macular edema. Thus, in recent years, there has been an increase in the development of automated methods for identifying macular pathologies \cite{melo2023oct}. However, these approaches face challenges due to limited data availability and imbalanced datasets, given the rarity of pathological findings. Therefore, the primary goal of OCT image synthesis (Table \ref{tab: oct-table}) is to augment data for algorithms, enabling not just disease detection but also enhancing the comprehension of retinal diseases.

The array of unconditional generative methods, as exemplified by \cite{oct-01, oct-17, oct-02, oct-07, oct-08, oct-18, oct-24}, underscores a notable improvement in downstream tasks when the generation process is meticulously tailored to the specific task and the modality of interest, such as various types of OCT images. These studies reveal that the careful consideration of generative architectures significantly enhances the performance of subsequent analyses. Addressing challenges like limited dataset availability, artifact removal, and domain shift issues through the adoption of techniques like PGGANs, structure-consistency GANs, and StyleGAN2-ADA contributes significantly to the generation of diverse and realistic synthetic images. The effectiveness of these unconditional methods is further emphasized in tasks like classification, despeckling, and segmentation, showcasing their versatility in diverse medical image analyses.  The integration of diffusion-based methods, exemplified by \cite{oct-24}, further accentuates this point, showcasing the adaptability of generative architectures on different tasks. These findings collectively emphasize the importance of thoughtfully tailoring generative approaches to the intricacies of medical image analysis, promoting their adoption in the domain of OCT image generation and analysis.

Highlighting the importance of conditioning for accurate representation of retinal pathologies in OCT images, \cite{oct-19} employs a conditional VAE with contrastive learning to discriminate between diseases in the embedding space. Similarly, \cite{oct-15} introduces a FastGAN with a Fourier Attention Block (FAB), improving feature map rescaling based on frequency contributions. Inspired by Spectral Domain OCT (SD-OCT) that utilizes Fourier/spectral detection for improving OCT imaging resolution, the generator uses the Fast Fourier Transform (FFT) and inverse-FFT algorithms to eliminate noise and enhance details. Results demonstrate FAB's effectiveness in noise reduction, but lower accuracies in classifying certain pathologies indicate the need for incorporating more distinctive features. \cite{oct-05} underscores GANs' importance in evaluating subject-specific eye development and tracking subtle retinal changes over time. This study introduces a counterfactual GAN based on starGAN, synthesizing high-resolution counterfactual OCT images and longitudinal time series from retrospective data. These images allow exploration of hypothetical scenarios by altering certain subject characteristics while preserving identity and image acquisition settings. Despite clinical evaluations aligning with previous studies, limitations include potential correlation of disease features with age or sex, which may result in the GAN altering them when generating counterfactual images, as well as the inability to learn relationships for unseen subject groups. Expert evaluations reveal challenges in distinguishing real from synthetic images, highlighting the algorithm's realism in mimicking retinal changes and emphasizing the need for further research into counterfactual approaches in medical imaging for a deeper understanding of subject-specific changes over time.

\cite{oct-03} presents a biomarkers-aware asymmetric bibranch GAN for post-therapeutic (PT) OCT image generation. Using Adaptive Memory Batch Normalization (AMBN), the method transfers knowledge from large-scale data to the target branch, trained on small-scale paired data.  The generated images closely align with clinical requirements, aiding in treatment adjustment and monitoring of neovascular age-related macular degeneration (nAMD) progression, whereby over 95\% of synthetic SD-OCT images effectively convey clinically relevant information and predict biomarker presence accurately. Additionally, the study shows significant advancements in predicting therapeutic effects over longer periods, up to 12 months after initial treatment, compared to existing methods. \cite{oct-14} employs a dual-discriminator Fourier acquisitive GAN to generate realistic OCT images, leveraging Fourier domain similarity. Meanwhile, \cite{oct-09} uses a Pix2PixHD GAN for multi-modal age-related macular degeneration categorization with CAM manipulation to enhance image diversity. Recognizing the limitations of 2D OCT methods, \cite{oct-09} underscores the importance of 3D volumes aligned with routine clinical examinations. Additionally, \cite{oct-06} enhances cross-subsectional resolution using a fully convolutional approach, validated for realism by medical experts. \cite{oct-06} also introduces BRISQUE, a perceptual image quality indicator, effectively evaluating image quality without a reference. These studies contribute nuanced perspectives to OCT image synthesis, emphasizing practical applications and diagnostic advancements.

In the absence of paired data for training, unpaired architectures, like cycle-consistent GAN variations, are employed for image harmonization, ensuring consistency of features across different acquisition sites. However, challenges like instability and contrast inversion limit their reliability in real-world medical imaging. Methods in \cite{oct-12, oct-25, oct-20, oct-22, oct-23} preserve anatomical layout during generation, assuming consistent morphological shape across imaging sites. \cite{oct-12} introduces a segmentation-renormalized image translation approach using a residual U-Net, enhancing image harmonization and robustness against perturbations. It explores Kernel Inception Distance (KID) alongside FID for evaluating visual fidelity, suggesting a multitask autoencoder trained on medical images for feature extraction and FID/KID calculation. Several studies have made significant advancements in medical imaging using GANs. \cite{oct-25} achieved a 94.83\% accuracy in detecting DME fluid with synthetic data using a Pix2Pix U-net-based generator for OCT image synthesis. \cite{oct-20} demonstrated comparable performance between synthetic and real data for OCT patch generation using a DCGAN. \cite{oct-22} enhanced chorio-retinal segmentation by optimizing generator choice with an improved StyleGAN2, leading to substantial improvements in patch classification. Lastly, \cite{oct-23} improved plaque detection accuracy by 15.8\% using conditional GANs with class encoding, aligning well with clinical findings.

Translation-based approaches in OCT imaging address critical clinical needs, like predicting treatment responses through generating post-therapeutic OCT images from pre-therapeutic ones. For example, \cite{oct-10} adapts Pix2PixHD to forecast responses to anti-vascular endothelial growth factors (anti-VEGF) therapy in retinal vein occlusion patients. The generated synthetic OCT images closely match real ones, aiding in treatment response prediction and patient follow-up. Moreover, \cite{oct-13} introduces a method for synthesizing polarization-sensitive (PS) OCT images using a modified Pix2Pix GAN, offering an alternative to complex hardware. Validation demonstrates interchangeability between synthetic and real PS-OCT images, promising applications in cancer diagnosis. Additionally, \cite{oct-21} tackles differences in image intensities and signal-to-noise ratios between instruments using an unpaired cycleGAN-based domain adaptation network. This enables cross-instrumental image analysis, crucial for deep learning model training and clinical segmentation tasks.

\cite{oct-04} presents a method using inter-modality translation to synthesize OCT color-coded macular thickness maps from fluorescein angiography (FA) data and vice versa, aiding in diagnosing diabetic macular edema. By integrating FA's physiological data with OCT's structural data, a more precise assessment of treatment response becomes possible, despite limitations in training data and generator architecture. Similarly, \cite{oct-16} employs GANs, specifically the U-Net generator, for generating OCT images from fundus images to diagnose glaucoma. \cite{oct-16} demonstrates that incremental training improves model accuracy and adaptability across different hospital settings, showcasing the potential of generated images in glaucoma identification.

Synthesized OCT image quality evaluation typically mirrors assessments in other modalities, relying on objective fidelity and diversity metrics. Visual Turing tests, conducted in a limited number of studies \cite{oct-01,oct-06, oct-10,oct-04} with 2-3 experts, indicate that artifacts around crucial regions and unusual noise patterns often hint at synthetic samples. In \cite{oct-01}, specialists noted lower visibility of scleral spurs and anterior segment structures in synthetic AS-OCT ACA images, critical for imaging the anterior chamber angle, aiding identification. \cite{oct-04} revealed the model's ability to represent leakage areas in FA frames with color codes but also generated inconsistent background vessels, signaling synthetic images. In \cite{oct-05}, despite difficulty distinguishing real from fake images in many cases, features like choroid and vitreous areas, pathologic features, and shadowing artifacts hinted at synthetic images. Several studies have assessed the clinical utility of synthetic OCT images, providing insights into image quality and applicability. In \cite{oct-03}, analysis by retinal clinicians of generated post-therapeutic images unveiled predictive accuracy for short and long-term treatment effects, facilitating effective doctor-patient communication and adjustments in treatment strategy. The proposed BAABGAN also demonstrated effective prediction of biomarkers. \cite{oct-05} introduced a neural network generating high-resolution counterfactual OCT images, enabling exploration of hypothetical scenarios for retinal aging research. \cite{oct-08} estimated foveal hypoplasia grade and visual acuity using deep learning, incorporating real and GAN-generated images. \cite{oct-10} focused on calculating central macular thickness, while\cite{oct-21} compared retinal volumes.Additionally, \cite{oct-18} addressed despeckling in AS-OCT images. Despite these advances, the evaluation of synthetic images is lacking, especially in terms of diversity, quality and clinical utility, with privacy preservation mostly unaddressed in current methods.

\subsubsection{Multiple Modalities}  \label{sec-multi}

\begin{table*}[h!]
\centering
\scriptsize
\begin{tabular}{ccllllcccccccc}
\hline
\multirow{3}{*}{\textbf{Application}} & \multirow{3}{*}{\textbf{Type}} & \multirow{3}{*}{\textbf{Modalities}} & \multirow{3}{*}{\textbf{Technology}} & \multirow{3}{*}{\textbf{Paper}} & \multirow{3}{*}{\textbf{DT}} & \multicolumn{7}{c}{\textbf{Evaluation}} & \multirow{3}{*}{\textbf{Code}} \\ \cline{7-13}
 &  &  &  &  &  & \multicolumn{2}{c}{U} & \multirow{2}{*}{F} & \multirow{2}{*}{D} & \multirow{2}{*}{Q} & \multirow{2}{*}{C} & \multirow{2}{*}{P} &  \\ \cline{7-8}
 &  &  &  &  &  & Train & Test &  &  &  &  &  &  \\ \hline
\multirow{3}{*}{Unconditional} & GAN,DM & MR-brain,Chest x-ray & StyleGAN, DDPM & \cite{multi-4} & None &  &  & \checkmark &  &  &  & \checkmark &  \\
 & \begin{tabular}[l]{@{}l@{}}\textbf{H:}VAE,\\DDPM\end{tabular}  & \begin{tabular}[l]{@{}l@{}}3D:CT-lung, MR-brain,\\  MR-breast, MR-knee \end{tabular} & \begin{tabular}[l]{@{}l@{}}VQ-GAN followed\\by DDPM\end{tabular}   & \cite{multi-8} & Seg & A & R &  & \checkmark & \checkmark(2) & \checkmark &  & \checkmark \\
 & DM & \begin{tabular}[l]{@{}l@{}}CT-abdomen, CT-pelvic,\\Chest x-ray, MR-heart \end{tabular}  & MT-DDPM & \cite{multi-1} & Class & S,A & R & \checkmark & \checkmark & \checkmark(3) &  &  &  \\ \hline
\multirow{3}{*}{\begin{tabular}[l]{@{}l@{}}Class or Label \\ cond.\end{tabular} } & GAN & CT-lung, MR-brain & HA-GAN & \cite{multi-14} & Class & \multicolumn{2}{c}{\checkmark} & \checkmark & \checkmark &  & \checkmark &  & \checkmark \\
 & GAN & \begin{tabular}[l]{@{}l@{}}Conditioned on seg\\ mask: cardiac cine-MRI, \\liver CT,  and Fundus images\end{tabular} & Multiple GANs & \cite{multi-12} & Seg & \multicolumn{2}{c}{\checkmark} & \checkmark &  & \checkmark(4) &  &  &  \\
 & DM & \begin{tabular}[c]{@{}c@{}}Chest x-ray, Histopathology,\\ Opthalmology Fundus images\end{tabular}   & \begin{tabular}[l]{@{}l@{}}Medfusion: \\Conditional LDM\end{tabular} & \cite{multi-100} & Class & S & R & \checkmark & \checkmark &  &  &  & \checkmark \\ \hline
Attribute cond. & DM & \begin{tabular}[l]{@{}l@{}}Chest x-ray, Dermatoscopy,\\ Histopathology\end{tabular}  & DDPM & \cite{multi-13} & Class & \multicolumn{2}{c}{\checkmark} &  &  &  &  &  &  \\ \hline
\multirow{2}{*}{Text-guided} & \multirow{2}{*}{DM} & \begin{tabular}[l]{@{}l@{}}breast Ultrasound, spleen CT, \\prostate MR\end{tabular} & EMIT-Diff & \cite{multi-10} & Seg & \multicolumn{2}{c}{\checkmark} & \checkmark &  &  &  &  & ** \\
 &  & \begin{tabular}[l]{@{}l@{}}Chest x-ray, Histopathology,\\ MR-prostate\end{tabular} & \begin{tabular}[l]{@{}l@{}}FineTuned\\ StableDiffusion\end{tabular} & \cite{multi-18} & Class & \multicolumn{2}{c}{\checkmark} & \checkmark &  &  &  &  &  \\ \hline
\multirow{3}{*}{Other cond.} & \multirow{2}{*}{DM} & prostate-MRI, chest x-ray& \begin{tabular}[l]{@{}l@{}}Med-cDiff:\\Conditional DDPM\end{tabular} & \cite{multi-16} & Seg,Class & \multicolumn{2}{c}{\checkmark} & \checkmark &  &  &  &  &  \\
 &  & \begin{tabular}[l]{@{}l@{}}Conditioned on 2perpendicular\\  2d DMs as a 3d prior: MRI, CT\end{tabular} & TPDM & \cite{multi-15} & None &  &  & \checkmark &  &  &  &  & \checkmark \\
 & GAN & \begin{tabular}[l]{@{}l@{}}Combining Chest x-ray\\images and Tabular Data\end{tabular} & αGAN and CTGAN & \cite{multi-20} & Class,Reg & \multicolumn{2}{c}{\checkmark} &  &  & * &  &  &  \\ \hline
\end{tabular}

\caption{Overview of all summarized and discussed studies on \textbf{multiple modalities} synthesis with information about the technology utilized, the downstream task (DT), the evaluation procedure and code availability. Evaluation is split according to the evaluation metrics reported, covering utility (U), fidelity (F), diversity (D), qualitative assessment (Q), clinical utility (C) and privacy (P) evaluation.In the Type column, H stands for hybrid approach combining two different models. In evaluation of utility (U), R stands for real data, S stands for synthetic data, and A stands for augmented data. In qualitative evaluation, the number between parenthesis indicate the number of specialists employed, while * stands for visual inspection of visualization plots.** stands for the availability of code of the original method. Acronyms: Classification (class), Segmentation (seg), regression (reg), conditioning (cond),Denoising Diffusion Probabilistic Model (DDPM), Transformer-based Denoising Diffusion Probabilistic Model (MT-DDPM),Hierarchical Amortized GAN (HA-GAN),Two Perpendicular 2D DMs (TPDM) }
\label{tab:multiple}
\end{table*}

In preceding sections, the focus was primarily on papers centered around generative models targeting a single modality, such as exclusive generation of MRIs or CT scans. This section delves into papers exploring models capable of generating multiple modalities using the same architecture, thereby avoiding significant modifications. For instance, \cite{multi-8} demonstrated that diffusion models are capable of generating realistic 3D synthetic data across four different anatomical regions, two modalities (CT and MRI), and three different resolutions. Remarkably, the diffusion model achieved this without the need for fine-tuning any hyper-parameters to adapt to the different datasets, despite the small training datasets. In addition to the diffusion model, a VQ-GAN architecture was used to compress the 3D CT and MRI images into a latent space, where the DDPM process was performed. In  \cite{multi-4}, a comparison was made between the performance of a GAN and a diffusion model in generating synthetic brain MRI and chest X-rays, with a focus on memorization of training data.\cite{multi-1} utilize Swin-transformer in the design of a diffusion model to enhance the image synthesis quality, applied on multiple imaging modalities. \cite{multi-10} adopts a holistic approach to medical image synthesis, utilizing controllable diffusion with DDPMs incorporating edge information to generate realistic and diverse synthetic ultrasounds, spleen CTs, or prostate MRI images while preserving essential characteristics. Initial pretraining on RadImageNet involves triplets for diverse prompts, while edge information from RadImageNet, guided by a HED algorithm (Holistically-Nested Edge Detection) \cite{xie2015holistically}, enhances realistic image generation. Subsequent fine-tuning on a task-specific dataset adapts the diffusion model to unique characteristics.

\cite{multi-12} employed multiple GANs conditioned on segmentation label masks for different MRI, CT, and retinal images, while \cite{multi-100} utilized a class conditional LDM on various modalities including chest x-ray, histopathology, and fundus images. Notably, a DDPM was employed in \cite{multi-13} conditioned on either the diagnostic label alone or in conjunction with a property such as hospital or a sensitive attribute (e.g.: ethnicity,sex). Augmenting the training data with synthetic data improved both the accuracy and fairness of downstream classification models, particularly for under-represented groups in the training data.


Training diffusion-based models for text-to-image generation from the scratch demands access to extensive datasets with image-caption pairs and significant computational resources. However, in the context of medical image generation, the availability of large, publicly accessible datasets containing text reports is limited. \cite{multi-18} proposes a solution by demonstrating that pre-trained Stable Diffusion models, initially trained on natural images, can be adapted to diverse medical imaging modalities through textual inversion \cite{textualinversion}. To introduce a medical modality as a new concept to a pre-trained diffusion model, textual inversion finds a vector in the text embedding space that best represents the new concept. This is achieved by freezing the entire architecture except for this specific embedding vector, and then performing backpropagation with a similarity loss using a small set of example images.

\cite{multi-20} is the only paper in our survey that generates synthetic hybrid data that combines both images and non-image data, whereby two GANs are used to create synthetic records containing both CXRs and non-image clinical data. This is done by reducing the dimensionality of the images using the pretrained encoder of the GAN.

While the majority of papers focus on utility and fidelity evaluation of the synthetic data, only few papers considered measures of diversity \cite{multi-8,multi-1,multi-100,multi-4}. \cite{multi-1} used a diversity score (DS) to measure how diverse the synthetic images were. The DS, or the nearest SSIM difference, is the KL divergence between the distributions of nearest SSIMs in real and synthetic images. To find the nearest SSIM for the synthetic images, the authors calculated the SSIM of  each synthetic image with all the others and found the closest match. \cite{multi-4} argues that if the  SSIM was calculated between each synthetic image and all training images, then this score would quantify the memorization of the training data by the model. \cite{multi-8} also used SSIM to measure diversity of synthetic images, by calculating the SSIM between pairs of synthetic data to check how similar they are, and thus investigate model's ability to generate diverse images. 

Few studies consider qualitative assessment of the generated data, with a limitation of considering small number of experts. For instance, \cite{multi-8} relied on the evaluation of image quality by only two human experts in terms of realistic image appearance, anatomical correctness, and consistency between slices, while \cite{multi-1} involved three experts, and \cite{multi-12} considered four experts.

Finally, the evaluation of memorization i.e. that generative models can simply generate samples that are copies of the training data, is the focus of \cite{multi-4}. It involved calculating pairwise image correlation between each synthetic image (from a selection of only 100 synthetic images for analysis) and the entire set of training images, and the findings indicated that diffusion models exhibit a higher susceptibility to memorizing the training images than GANs.

\subsection{Medical text}

\label{sec:text}

\begin{table*}[h!]
\centering
\scriptsize
\begin{tabular}{ccllllcccccccc}

\hline
\multirow{2}{*}{\textbf{Application}} &
  \multirow{2}{*}{\textbf{Type}} &
  \multirow{2}{*}{\textbf{Modality}} &
  \multicolumn{1}{c}{\multirow{2}{*}{\textbf{Technology}}} &
  \multicolumn{1}{c}{\multirow{2}{*}{\textbf{Paper}}} &
  \multicolumn{1}{c}{\multirow{2}{*}{\textbf{\begin{tabular}[c]{@{}c@{}}Downstream \\ Task\end{tabular}}}} &
  \multicolumn{6}{c}{\textbf{Evaluation}} &
  \multirow{2}{*}{\textbf{\begin{tabular}[c]{@{}c@{}}Comparison\end{tabular}}} &
  \multirow{2}{*}{\textbf{Code}} \\ \cline{7-12}
 &
   &
   &
  \multicolumn{1}{c}{} &
  \multicolumn{1}{c}{} &
  \multicolumn{1}{c}{} &
  \multicolumn{1}{c}{\textbf{U}} &
  \textbf{F} &
  \textbf{D} &
  \textbf{Q} &
  \textbf{C} &
  \textbf{P} &
   &
   \\ \hline
\begin{tabular}[c]{@{}c@{}}NLP Enhancement\end{tabular} &
  LLM &
  \begin{tabular}[c]{@{}c@{}}Discharge summary\end{tabular} &
  \begin{tabular}[c]{@{}l@{}}LLaMa\\ GPT-3.5, 4\end{tabular} &
  \cite{kweon2023publicly} &
  NLP Tasks &
  \checkmark &
   &
   &
  \checkmark &
  \checkmark &
   &
  \checkmark &
  \checkmark \\ \hline
\begin{tabular}[c]{@{}c@{}}NLP Enhancement\end{tabular} &
  LLM &
  \begin{tabular}[l]{@{}l@{}}Diabetic ketoacidosis\\ guideline\end{tabular} &
  GPT-4 &
  \cite{hamed2023advancing} &
  \begin{tabular}[c]{@{}l@{}}Question \\ Answering\end{tabular} &
     \checkmark &
   &
   &
   &
   &
   &
   &
   \\ \hline
\begin{tabular}[c]{@{}c@{}}NLP Enhancement\end{tabular} &
  \begin{tabular}[c]{@{}c@{}}RNN,\\ GAN,\\ PLM\end{tabular} &
  \begin{tabular}[c]{@{}c@{}}Patient history,\\ present Illness\end{tabular} &
  \begin{tabular}[c]{@{}l@{}}CharRNN, GAN,\\ GPT-2,  CTRL\end{tabular} & \cite{li2021synthetic} &
  \begin{tabular}[c]{@{}l@{}}Name entity\\ recognition\end{tabular} &
  \checkmark &
   &
   &
  \checkmark &
   &
   &
   &
  \checkmark \\ \hline
\begin{tabular}[c]{@{}c@{}}Text augmentation, \\ de-identification\end{tabular} &
  PLM &
  \begin{tabular}[c]{@{}c@{}}Injury records\end{tabular} &
  BERT &
  \cite{zhou2022datasiftertext} &
  \begin{tabular}[c]{@{}l@{}}Replace sensitive \\ data\end{tabular} &
  \checkmark &
   &
   &
   &
   &
  \checkmark &
   &
  \checkmark \\ \hline
\multicolumn{1}{c}{\begin{tabular}[c]{@{}l@{}}Text augmentation\end{tabular}} &
  \multicolumn{1}{c}{PLM} &
  \begin{tabular}[c]{@{}c@{}}Discharge summary\end{tabular} &
  \begin{tabular}[c]{@{}l@{}}Transformer,\\ GPT-2\end{tabular} &
  \cite{al2021differentially} &
  Classification &
  \checkmark &
  \checkmark &
   &
   &
   &
  \checkmark &
  \checkmark &
  \checkmark \\ \hline
\multicolumn{1}{c}{\begin{tabular}[c]{@{}l@{}}De-identification, \\ Text augmentation\end{tabular}} &
  \multicolumn{1}{c}{PLM} &
  \begin{tabular}[c]{@{}c@{}}Discharge summary\end{tabular} &
  \begin{tabular}[c]{@{}l@{}}LSTM, GPT-2\end{tabular} &
  \cite{libbi2021generating} &
  \begin{tabular}[c]{@{}l@{}}Name entity\\ recognition\end{tabular} &
  \checkmark &
   &
   &
   &
   &
  \checkmark &
   &
  \checkmark
  \\ \hline
\end{tabular}
\caption{Reviewed generative models for \textbf{medical text}. Acronyms: Large Language Model (LLM),Large Language Model Meta AI (LLaMa), Generative Pre-Trained Transformer (GPT), Pre-trained Language Models (PLM), Natural Language Processing (NLP), Recurrent Neural Netwrok (RNN), Character Recurrent Neural Network (CharRNN), Conditional Transformer Language (CTRL), Bidirectional Encoder Representations from Transformers (BERT)}
\label{tab:text}
\end{table*}

Different from synthesizing medical imaging or signal data, medical text generation focuses on language models and transformers. As we mentioned before, the main purposes of generating medical text are categorized into (i) enhancing NLP tasks such as Name entity recognition, concept extraction, relation extraction, and question answering, (ii) augmenting medical text data where the high-quality clinical notes are limited and privacy sensitive to use for research, (iii) replacing sensitive information in the real data with generated synthetic data. A variety of medical text data types have been utilized to train the models including 
\begin{enumerate}
    \item discharge summary, a clinical document summarizing the patient's hospitalization from admission to discharge to provide a comprehensive overview of the patient's diagnosis, treatments, condition at discharge, and recommendations for follow-up care; 
    \item disease guideline (a.k.a clinical guideline or practice guideline), a systematically developed documentation to assist medical professionals and patient decisions about appropriate healthcare for specific clinical situations;
    \item case report, a detailed report of the symptoms, diagnosis, treatment, and follow-up of an individual patient;
    \item history of present illness which is a detailed documentation of the development of the patient's illness, including the onset of symptoms, duration, intensity, and any factors that aggravate or relieve them;
    \item chief complaint which describes the primary reason or symptoms that brought the patient to seek medical attention
\end{enumerate}

Some additional input can also be used for medical text generation such as EHR data to generate clinical notes and predict outcomes \cite{al2021differentially}, diagnostic reports where studies take radiology, pathology, and other diagnostic images as input to generate imaging reports with detailed descriptions and diagnosis, clinical trial reports where studies take extensive medical documents as input to generate summarization of the reports. 

In medical text generation, there are no standards or well-accepted evaluation metrics to assess the fidelity, accuracy, and clinical utility for the generators. Moreover, an additional crucial aspect in evaluating synthetic medical text is the truthfulness and correctness of the generated data, ensuring that the information presented adheres to current medical guidelines and practices. If a medical text generator were to generate inaccurate or misleading information (such as hallucinating facts), the implications could be far more serious than with general-purpose text generators. Such errors could mislead healthcare professionals, impacting their decision-making processes and potentially affecting patient care outcomes. Another important aspect is the understand-ability and readability  by humans, mainly medical professionals and patients

Pre-trained and large language models such as BERT \cite{devlin2018bert} and GPT \cite{radford2019language} are widely applied in medical text generation applications. Similarly, there are commonly used training text datasets such as MIMIC clinical notes data \cite{johnson2016mimic,johnson2023mimic}, i2b2 (Informatics for Integrating Biology and the Bedside) data \cite{uzuner2007evaluating}, and n2c2 (National NLP Clinical Challenges) data \cite{henry20202018}. The i2b2 and n2c2 are well-known initiatives that provide datasets for NLP research in the biomedical and clinical domains.

 \cite{kweon2023publicly} developed a generator based on LLaMA \cite{touvron2023llama} but particularly focusing on clinical note generation using case reports, discharge summaries, and clinical notes from multiple sources including MIMIC-III \cite{johnson2016mimic}, MIMIC-IV \cite{johnson2023mimic}, i2b2 \cite{uzuner2007evaluating}, and CASI \cite{moon2014sense}. \cite{kweon2023publicly} first generated 158K synthetic clinical notes from case reports from the PubMed central patients dataset \cite{zhao2022pmc} using GPT-3.5. Then, incorporating the generated clinical notes, the proposed generator was fine-tuned based on LLaMA using discharge summaries from MIMIC-III database to capture the peculiarities in clinical notes such as capturing the relations between two medical terminologies, recognizing the reasoning process in the notes. The proposed model is fully open-source and trained on open-access datasets. The study evaluated the utility of the generator by conducting the NLP tasks using generated text (e.g., summarization, concept extraction, relationship extraction, name entity recognition, question answering) focusing on accuracy, relevancy, and completeness.  They applied GPT-4 to design specific evaluation prompts and also validated by four real clinicians. The study demonstrated their model outperformed other open-source instruction fine-tuned LLMs such as Alpaca \cite{taori2023stanford} and medical domain-specific models such as MedAlpaca \cite{han2023medalpaca} and ChatDoctor \cite{yunxiang2023chatdoctor}. The study demonstrated there is no significant difference between models trained on actual and synthetic clinical notes. However, in the training aspect, the proposed model is limited to dealing with only discharge summaries in one-turn instruction following tasks which means the user can ask one question or give one prompt then generator will provide an answer or generated text based on this single input. 

In the evaluation, the study did not analyze whether and to what extent the model would hallucinate the outcome, which is one of the major concerns for the language model use in the clinical practice. 

\cite{hamed2023advancing} utilized ChatGPT-4 and a Link Retrievel Plug-In to retrieve the medical information, integrate and analyze the collected information, and compile them into a human-understandable answer using three international guidelines on treating diabetic ketoacidosis. The study designed two prompts to 1) retrieve and integrate information focusing on diagnostic criteria, risk factors, signs and symptoms, investigations, and treatments of diabetic ketoacidosis and 2) generate comprehensive and evidence-based answers to the questions. The study aims to enhance the traceability and retrieval accuracy of ChatGPT-4 and claims their work maintained the accuracy, reliability, and quality of the generated answers. However, there are not sufficient systematic and quantitative experiments and evaluations being conducted in the study to support this conclusion. Furthermore, the truthfulness which assesses if the language models hallucinate the facts is not considered and evaluated sufficiently. This reflects many other use cases in using generative models in practical applications in medicine and healthcare domain. 

\cite{li2021synthetic} used a text dataset of 570 patient History and Present Illness (HPI) description from i2b2 and n2c2 challenges to train four different generative models - Character Recurrent Neural Network (CharRNN), Sub-sequence Generative Adversarial Network (SegGAN), GPT-2, and Conditional Transformer Language (CTRL). The four models were constructed based on different architectures and with different focuses. CharRNN is a character-level RNN which has shown its capability to tackle the out-of-vocabulary problem. Out-of-vocabulary challenge occurs when a word or phrase is not in the training data's vocabulary or dictionary used by a language model. The language models learn to understand and generate text only based on the training data, which means they can only recognize and interpret words in their training process and unable to process or understand the out-of-vocabulary words directly. SegGAN follows the traditional GAN framework, adapted to deal with discrete text output and able to adversarially learn on both the entire sequences (i.e. sentences, paragraphs, documents) and sub-sequences (i.e., words, phrases). GPT-2, as a large transformer-based pre-trained language model, has the capability to generate diverse and massive synthetic text data, while CTRL is a parameter conditional transformer language model which is conditioned on control codes that govern the content and task-specific behavior. Control words are specific words or phrases that specify domain, subdomain, entities, relationships between entities to steer or guide the generation process of the model towards a desired outcome. The study fine-tuned the GPT-2 and CTRL models to generate synthetic clinical HPI corpus. The models' performance was assessed using the Bilingual Evaluation Understudy (BLEU) metrics \cite{papineni2002bleu}, to select the best model to create a synthetic corpus of 500 HPI subsections. BLEU examines the similarity of the machine translation and human translation in n-gram representation. An n-gram is a contiguous sequence of n items from a text, where these items can be words for text generation. After manually annotating clinical entities such as problems, treatments, and tests in real and synthetic text, NER models were developed on both real and synthetic datasets to conduct a utility evaluation. The study showed NER models that were trained on synthetic corpus achieved slightly higher performance than that of the real corpus. The NER models that were trained on the augmented corpus (real + synthetic) achieved better performance than that trained on the real corpus only. The study concluded the generated synthetic HPI text can be used to enhanced the development of clinical NLP models.

\cite{zhou2022datasiftertext} proposed a model generating partially synthetic clinical text to protect against re-identification for individual patient and preserve the characteristics information from the population. The proposed model has two key processes - the artificial masking and the content substitution. Masking process utilized the tokenization function in BERT model to break down the sentence into tokens, added special tokens to indicate the beginning and end of sentences, then converted all the tokens into corresponding IDs in the pre-trained BERT vocabulary. Then, the tokenized data was fed into BERT for training. To mask the personally identifiable information in an efficient manner, the study embedded a blacklist and a white list to direct the masking. The blacklist and whitelist are customisable for different use cases. For instance, the blacklist can include words without meaning such as standard punctuation, stop words, while the whitelist contains sensitive attributes such as personally identifiable information. The model will avoid masking words in the blacklist, while the words in the whitelist will be masked with higher probabilities than words that are not in the lists. Masking sensitive words may result in obfuscation in some records. Therefore, the second step in the generative model is to replace defined sensitive information (i.e. words or phrases in the whitelist) of each document with text in similar documents in the dataset using Rapid Automatic Keyword Extraction \cite{rose2010automatic} and TextRank \cite{mihalcea2004textrank} methods. The model was trained on 153K injury reports including textual injury descriptions with gender, age, and injury and illness labels and MIMIC III discharge summaries. The study evaluated data privacy by comparing the similarity between original and generated synthetic texts in word choice, word order, and word frequency. The satisfactory privacy-preserved synthetic text, which is well-obfuscated text, should be significantly different from the original text in word choice, word order, and word frequency. The privacy evaluation utilized BLEU scores, ranging from 0 to 1, measuring the similarity, with lower scores indicating better obfuscation. The data utility of the generated synthetic text was assessed by training a classification model on the original text and applying the trained model to the synthetic text to calculate the data utility loss. Data utility loss was measured by comparing the classification performances between the original and obfuscated text (partially synthetic text). Finally, readability of the generated text was evaluated by a machine evaluator which is a fine-tuned text classifier based on a BERT model to determine if the text input is a real clinical note or generated from a machine. 


\cite{al2021differentially} proposed a differentially private and self-attention generative neural network model to generate synthetic medical texts. The generated text does not contain any identifying information from any individual if the input datasets were private. The study proposed a model as an autoregressive variation of the transformer model introduced by \cite{vaswani2017attention}. In the proposed model, each of the decoder blocks is composed of a multi-headed masked self-attention layer and a simple feed-forward network. The autoregressive feature means when masked self-attention stops the model from seeing tokens that are at the right of the current position which allows the model to take into account the last token when generating the next token. Furthermore, the study applied a differentially private Gaussian mechanism technique by modifying the optimization algorithms in their generator training process. This means in each training iteration, the original weights are adjusted by adding noise values derived from the Gaussian mechanism. This process generates differentially private tokens, resulting in an input layer distinct from the original and ensuring a certain level of privacy guarantee. Similar to \cite{kweon2023publicly} and \cite{li2021synthetic}, this study also used discharge summaries from MIMIC-III and i2b2 datasets. The model performance is evaluated by measuring the data utility on the word-level, document-level, and corpus-level. On the word-level, the study used BLEU score, Jaccard similarity, and G2 Test respectively to analyze the word co-occurrence or similarity, assessing whether the generated text has similar word distributions as the original text. On the document-level, data utility was analyzed by performing a classification task on original and synthetic data to determine whether the underlying document contains a certain disease or not. The study trained classifier on original text then tested on synthetic text, and trained on synthetic text then tested on original. The corpus-level utility, containing all documents, was assessed by an adversarial classification task. The study trained multiple classifiers on combined datasets of original and generated text and predicted whether a document is original or synthetic. The purpose of the classifier is to distinguish between original and synthetic text which is analogous to the Turing test, where a human evaluator predicts the data to be original or synthetic. Regarding data privacy, the study applied different privacy in the model and emphasized on the privacy-preserving feature that proposed model has, but the study did not provide sufficient privacy evaluation on generated text. 

\cite{libbi2021generating} developed two language models - LSTM-based model and GPT-2 (Transformer) based model. 
The study sampled medical documents from the EHRs of 39 healthcare organizations in the Netherlands. Three domains of healthcare are represented within this sample: elderly care, mental care and disabled care. The models were evaluated in terms of the utility and privacy of the synthetic text. The utility was assessed by using synthetic text as a replacement of the real data to protect sensitive personal information and using synthetic text as a data augmentation approach to enrich the real medical documents. To evaluate data privacy, a user study was conducted with participants who were presented with the synthetic documents with high risks of privacy disclosure to judge if the documents contain sensitive information. The study observed the distribution of annotations in the training text and the diversity of the generated text have significant impact on the downstream task performance. The study reported that non-English language training faces additional challenges due to the lack of clear structure and conformity of the languages. Regarding data privacy and utility balance, the study argued that synthetic text does not need to be realistic for utility in downstream tasks so that privacy protection can be priotized. However, it is necessary to provide a mathematical privacy preservation in synthetic text generators such as using differential privacy methods.




\section{Discussion}
\label{sec:discussion}

The survey of the different papers across different data types and modalities has yielded notable findings  that we categorize into goal of synthesis, generation methods, and evaluation methods. For each insight or conclusion, we provide a non-exhaustive list of supporting references, previously discussed in section \ref{sec:results}. For further details, readers are directed to this section.

\subsection{Synthesis applications and purpose of synthesis}

\textbf{Diversity of clinically valid synthesis applications}:
Conditional generative models have demonstrated impressive capabilities in synthesizing missing or corrupted modalities across a diverse range of applications. As depicted in Fig.\ref{fig:synthesis-applications}, these applications vary depending on the data types involved. For example, in signals and images, both intra and inter modal translation between different modalities are common for various clinically valid reasons, such as synthesizing missing or corrupted MRI contrast when acquiring such images is not feasible \cite{qin_style_2022, huang_common_2022,zhan_lr-cgan_2021}, translating low-to-high density mammograms \cite{mamm-6} or DFM to FFDM mammograms \cite{mamm-11}, or B-mode to Elastography US \cite{us-19,us-3}. In the context of EHR, the synthesis and the choice of generative model depends on the specific format and requirements of the EHR data. For instance, the capability to manage high dimensionality is pivotal when handling longitudinal \cite{ehr-15,ehr-38} and aggregated \cite{s-11,ehr-16} EHR data, but its significance diminishes in the context of time-dependent \cite{s-10,s-18} and snapshot \cite{ehr-50,ehr-12} formats. Conversely, the preservation of temporal dependencies among clinical features remains crucial for both longitudinal and time-dependent formats. In medical text generation, the generated data serve multiple purposes beyond augmenting training data \cite{zhou2022datasiftertext,al2021differentially} . They can be also used for NLP enhancement, such as entity extraction \cite{li2021synthetic} and question-answering systems \cite{hamed2023advancing} , as well as text replacement for privacy reasons \cite{libbi2021generating}, as depicted in Table \ref{tab:text}.


\textbf{Limited utilization of synthetic data beyond augmentation:} Despite its potential, synthetic data is underutilized beyond augmentation. This valuable resource, capable of generating diverse datasets varying in quality and anatomy, could serve as a comprehensive validation and testing set during model development \cite{ct-33,ct-15,x-04,oct-15,oct-05,x-03}. It has the potential to uncover algorithm limitations early on, particularly for challenging cases that may not be accessible due to data sharing policies or rarity, facilitating smoother translation of algorithms into clinical practice.



\textbf{Most generative approaches are designed to handle the lack of training data but thorough analysis of how this is achieved is missing:} While numerous methodologies have been proposed to address the scarcity of training data, the specific elements of the generated data that contribute to improved outcomes remain largely unexplored. This is particularly true when the use of synthetic data alone does not yield any enhancement in performance \cite{ct-21, ct-l-02, ct-l-06, ct-l-07, ct-l-08, ct-l-10}. Only a handful of studies have indicated that the type of synthetic data incorporated into the training process can significantly impact results \cite{ct-23,x-12}. For instance, incorporating challenging cases such as dense mammograms \cite{mamm-6},  severe examples of lung nodules in X-ray images or overlapping benign and malignant nodules in breast ultrasound \cite{us-4}, or underrepresented groups \cite{derm-4,derm-10}, has been shown to enhance performance. This underscores the potential benefits of active learning strategies, either during the data generation process or in selecting synthetic samples to be included in the training set.

\subsection{Generation techniques }

\textbf{The leap of textual conditioning opens future directions:}
Recent advancements in text-conditional generative models offer an innovative approach to integrating descriptive data into synthetic medical data generation. This can involve structured text input specifying medical conditions and demographic data or utilizing medical reports as a source of clinically relevant guidance for the model \cite{s-44,derm-3,mamm-4,ct-l-13,x-19,multi-10}.  Readers are directed to the "Text-guided" row in the tables for more information. Text-guided conditioning has been already established  across various modalities, although some modalities are lagging behind and have not yet implemented this approach .

\textbf{Conditional Models- Limited leverage of clinical knowledge, patient context, and text:}
A significant gap exists in the utilization of prior clinical knowledge, the availability of complimentary information accompanying the acquisition, as well as pathological processes,  biomarkers and radiology reports to inform the generation process \cite{ct-10,ct-l-15,ct-14, oct-03, oct-04,ct-07}. Most models fail to take advantage of these valuable resources, leading to potentially sub-optimal results. Similarly, conditioning based on subject characteristics patient and demographics is rarely utilized, except in isolated instances \cite{ehr-59,ehr-25,derm-10,mamm-4}. This highlights the need to focus on more personalized synthesis tailored to specific subject characteristics and patient demographics, prioritizing fairness and inclusivity in medical data generation. Strong evidence suggests that conditional models leveraging such textual information, clinical knowledge, or other images/data inputs consistently produce higher quality and more realistic data compared to their non-conditional counterparts.

\textbf{Limited tailoring of generative approaches to the specifics of medical data:} Most synthetic methods proposed for medical image generation still largely rely on loss functions and architectures initially designed for natural images, without considering the intricate aspects of medical image acquisition, physics and well defined statistics that can guide the generative process \cite{ct-l-13,ct-20,ct-14,ct-07}. Moreover, a significant number of models are designed to generate small patches (especially 3D) and lead to the generation of artificial features that can lead to large errors in anomaly detection and other tasks \cite{ct-11, ct-l-11, ct-16, ct-28}.  Computational limitations also restrict recent transformer-based approaches to slice-based training \cite{ct-11, ct-12, ct-l-21} and diffusion models to low-resolution image generation.
In contrast, EHR generative models draw from established deep learning architectures like CTGANs  \cite{ehr-50,ehr-47} but address unique challenges such as synthesizing mixed-type data with  static (e.g., demographics data) and sequential features (e.g., vital sign time-series) \cite{ehr-69,ehr-38}, numerical features (e.g., blood pressure), and categorical features with many categories (e.g., medical codes) \cite{ehr-22,ehr-15}. Some of these models also consider modeling missing data patterns for informative analyses \cite{ehr-25,s-18}. See Table \ref{tab:ehr} for more information. Similarly, generative models for physiological signals leverage techniques for time-series data, some customizing to signal specifics like morphology  or inter-lead  \cite{r-252,r-98}.
Regarding medical text generation, the application of pre-trained models and large language models has recently become prominent in the field. These models are based on general foundation language models such as BERT and GPT and then adapted for medical applications through training or fine-tuning with specialized datasets, such as PubMed abstracts or clinical notes from the MIMIC database \cite{hamed2023advancing,kweon2023publicly}. See Table \ref{tab:text} for more information. However, this approach has limitations in that these advanced medical text generators may not fully grasp the underlying principles or logic of medical knowledge.

\subsection{Evaluation}


\textbf{Limited benchmarking, comparative studies and the use of small private datasets}
The absence of comparative studies between different methods and the lack of a well-defined benchmark for comparison leaves a gap in understanding the relative strengths and weaknesses of various synthetic data generation approaches and therefore highlights the need for a more systematic approach \cite{ct-l-02, ct-03, x-08,x-05,oct-01,oct-10}. Moreover, the reliance on small, often private datasets for training and testing generative models raises concerns about the generalizability and usability of the generated data in clinical settings  \cite{ ct-20,ct-l-05, ct-l-07,ct-31}.  Most studies report both training and validation of the proposed synthesis approaches on significantly limited datasets - typically up to several hundreds of samples are used for training with a substantially smaller set for validation/testing. In addition, there is a severe lack of investigation of the effects of training data itself \cite{ct-l-17}. 

\textbf{Limited evaluation frameworks}: Most approaches focus on quantitative evaluation using fidelity metrics that are not designed for the specific kind of medical data. For example, in the case of medical images, the fidelity metrics employed are tailored for diverse data types such as natural images, overlooking the distinctive characteristics and statistical properties inherent to typical medical imaging modalities \cite{ct-l-18,ct-l-16,ct-01, x-15, x-17,oct-20}. Similarly, the evaluation metrics used for time-dependent EHR and physiological signals may overlook critical temporal features essential for an accurate evaluation \cite{s-9,s-8,r-233, r-245}. In medical text generation, there are no standards or well-accepted evaluation metrics to assess the fidelity of the generated data.  Sole reliance on such generic metrics, without incorporating additional measures of clinical validation, utility, and human visual perception studies can lead to misleading conclusions about the true quality and suitability of synthetic medical data. 

However, despite the crucial role of clinical validation in making sure that crucial anatomical details are maintained in the generated data, there is a notable lack in its proper implementation, hindering the translation of synthetic data into practical clinical settings. Studies that do perform clinical validation, typically perform it on a small subset of the generated data, signifying a need for evaluation on large internal and external validation datasets \cite{ ct-30,ct-l-12,ct-22,ct-l-05, ct-28,  ct-33}.

Moreover, human assessments are still limited and face significant design shortcomings. Typically, these studies involve a limited number of experts, often ranging from 1 to 3, which raises concerns about the statistical significance of the results \cite{ct-04, ct-05, ct-l-01, ct-l-03, oct-10, oct-11}. Furthermore, the design of these studies is often unclear, especially regarding the selection criteria of synthetic samples for evaluation \cite{ct-11, ct-l-19, ct-l-17, ct-l-14}. To enhance the effectiveness of these studies, they should be designed to gather reports from experts on indications that suggest certain samples are synthetic, providing the ability to correct for these artifacts in future approaches \cite{ct-l-19}. In other instances, human visual assessment is reduced to a mere check of a few image samples for realism or a visual comparison of images from different methods, or merely inspecting visualizations such as PCA or t-SNE plots. 

A more comprehensive evaluation of synthetic data should include measures of diversity, ensuring that the synthetic data encompasses the full range of anatomical and pathological variations encountered in clinical practice \cite{ct-20,ct-l-05,ct-l-07,ct-31}. However, many current methods for generating synthetic data focus on reproducing specific image characteristics rather than generating a truly diverse set of instances. The lack of well-defined diversity metrics adds to the challenge, with researchers employing their own individual definitions \cite{multi-4,multi-8,multi-1,multi-100}.

\textbf{Limited evaluation of recent approaches based on diffusion and transformers:} Current research on diffusion models for data generation indicates they are more stable compared to GANs, with the ability to avoid mode collapse and vanishing gradients and add more control over the sampling process. However, a lot of diffusion models presented so far are yet to be properly tested and evaluated - utility studies are largely missing, as well as clinical validation or detailed qualitative and fidelity analysis of the generated data. Moreover, training is still performed on limited data that may lack adequate variability \cite{ct-l-18, ct-03,ct-l-01, ct-l-12, ct-16,  ct-06}.




\textbf{Privacy disparities in synthetic medical data:}
The assessment of privacy protection in generated medical data, encompassing images, EHR, and signals, faces notable disparities in standardized methodologies. Despite assertions of privacy capabilities and the potential for widespread data sharing, the absence of well-defined evaluation methods for medical images, signals, and text compared to EHR raises concerns. EHR privacy assessments commonly involve testing synthetic data for vulnerabilities against potential adversary attacks and demonstrating resilience against various threats. In contrast, privacy discussions in medical images primarily focus on the memorization of training data, leaving questions about identity disclosure unexplored. The extent to which generators and diffusion models memorize or replicate data, and their impact on tracing back learned distributions to original training data, remains unclear. While EHR benefits from established evaluation practices, addressing the significant gaps in privacy protection evaluation for other data types is crucial for ensuring the responsible deployment of synthetic data in medical contexts.

\section{Recommendations} \label{sec:Recommendations}
Based on the insights collected from the surveyed papers, we propose the following recommendations to be taken into account for future research:

\textbf{Generative approaches should be tailored to the medical data and downstream task or their intended use:} As specified earlier, generative approaches should be more tailored to the specifics of the medical data and also to the specific task at hand, as there are different requirements for each tasks \cite{ct-03,ct-l-06, ct-l-15, ct-01, ct-17, ct-l-21}. For instance, requirements for segmentation/classification are substantial different compared to registration or denoising tasks . 


\textbf{Synthetic data should be utilized beyond data augmentation to train and evaluate medical AI software:}  Synthetic data should not only serve as a tool for data augmentation and training downstream models but also for validation and testing purposes \cite{x-06, x-20, x-21,  oct-18, oct-21}. Moreover, it can be used for the development of explainability mechanisms to improve not only the confidence of the proposed analysis models, but also the confidence of clinicians in utilizing automated methods in clinical practice.

\textbf{Prioritizing fairness and inclusivity in medical data generation:} 
Due to its scarcity, there is a need to focus on more personalized synthesis tailored to specific subject characteristics and patient demographics, prioritizing fairness and inclusivity in medical data generation.

\textbf{Adapting language models to medical tasks should be further approached with more care:} A thorough analysis is needed on the interpretability capabilities of diffusion models when it comes to the understanding and translation of radiological reports to images.

\textbf{A call for benchmarking and comparative studies to promote openness and collaboration:
}
The use of established and well defined benchmarks and open-source datasets would not only accelerate progress in the field but also provide valuable insights into the strengths and weaknesses of various techniques, guiding researchers towards the optimal tools for specific clinical applications.

\textbf{A need for more comprehensive evaluation:} There is a need for holistic assessment covering more specific fidelity metrics, proper clinical validation and human assessment, and well-defined diversity metrics. Moreover, there is a need for well established privacy evaluation practices, especially for data types where there is a significant gap. This is crucial for ensuring the responsible deployment of synthetic data in medical contexts. 








\section{Conclusion} \label{sec:conclsuion}
The advent of AI and ML has a great potential to revolutionize healthcare, with synthetic data generation emerging as a promising solution to the challenges of data availability and diversity. Despite the growing interest in generative models for medical data synthesis, existing reviews often have limited scope, focusing on specific modalities or data types, or overlooking certain types of generative models beyond basic data augmentation techniques. This survey paper aims to fulfill this need by providing a comprehensive examination of the use of generative models, such as GANs, VAEs, LLMs, and DMs, in synthesizing medical data across various modalities, including imaging, text, time-series, and tabular data. Through a meticulous search strategy involving databases such as Scopus, PubMed, and arXiv, this survey paper identifies and analyzes 249 relevant studies published from January 2021 to November 2023, excluding review and perspective papers. The survey aimed to provide a holistic understanding of the applications of these generative models in generating synthetic medical data,shedding light on the current practices, potential, and challenges in this field, delving into three key aspects: the purpose of synthesis, generation techniques, and evaluation methods.

Our findings reveal a diversity of clinically valid synthesis applications, yet also highlight a limited exploration of synthetic data’s potential and its utilization beyond augmentation. Most generative approaches are designed to handle the lack of training data, but a thorough analysis of how this is achieved is often missing.

In terms of generation techniques, the advent of textual conditioning opens up exciting future directions. However, there is limited leverage of clinical knowledge, patient context, and text in conditional models, and generative approaches are often not tailored to the specifics of medical data.

Evaluation methods also present challenges, with limited benchmarking, comparative studies, and the use of small private datasets. Evaluation frameworks are often limited, and recent approaches based on diffusion and transformers are not thoroughly evaluated. Privacy disparities in synthetic medical data also pose a significant concern.

In essence, while synthetic data generation holds immense potential in transforming healthcare research and practice, there are notable gaps and challenges that need to be addressed. There is a need for more personalized synthesis approaches, standardized evaluation methodologies, and in-depth evaluation approaches relevant to clinical applications. The field would also benefit from more benchmarking and comparative studies to promote openness and collaboration.

This survey paper serves as a valuable resource for researchers and practitioners interested in leveraging generative models for synthesizing medical data. By providing a holistic understanding of the field, we hope to spur further research and innovation in this critical area of healthcare AI and ML. The insights gained from this survey underscore the importance of continued exploration and innovation in synthetic medical data generation, with the ultimate goal of advancing patient care and clinical practice.
\section{Acknowledgements}
This work was funded by the European Union under the Horizon Europe grant 101095435. Views and opinions expressed are however those of the author(s) only and do not necessarily reflect those of the European Union. Neither the European Union nor the granting authority can be held responsible for them.

\begin{figure}[h!]
    \centering
    \includegraphics[width=0.5\linewidth]{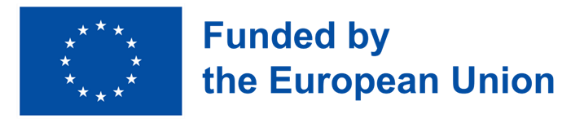}
\end{figure}
\section{Statements and Declarations}

The authors have no competing interests to declare that are relevant to the content of this article.

\section{CRediT author statement}

\textbf{Mahmoud Ibrahim:} Conceptualization, Methodology, Investigation, Data Curation, Writing, Visualization, Project Administration.\textbf{Yasmina Al Khalil:} Methodology, Investigation, Writing, Visualization. \textbf{Sina Amirrajab}: Methodology, Investigation, Writing. \textbf{Chang Sun:} Methodology, Investigation, Writing, Funding Acquisition. \textbf{Bart Elen:} Writing - Review and Editing, Funding Acquisition, Supervision.  \textbf{Gokhan Ertaylan:} Funding Acquisition, Supervision. \textbf{Michel Dumontier:} Funding Acquisition, Supervision.

\bibliography{references.bib}
\bibliographystyle{vancouver} 
\onecolumn
\section{Appendix}

\subsection{Evolution of the usage of generation techniques over years}
\begin{figure}
    \centering
    \includegraphics[width=0.7\linewidth]{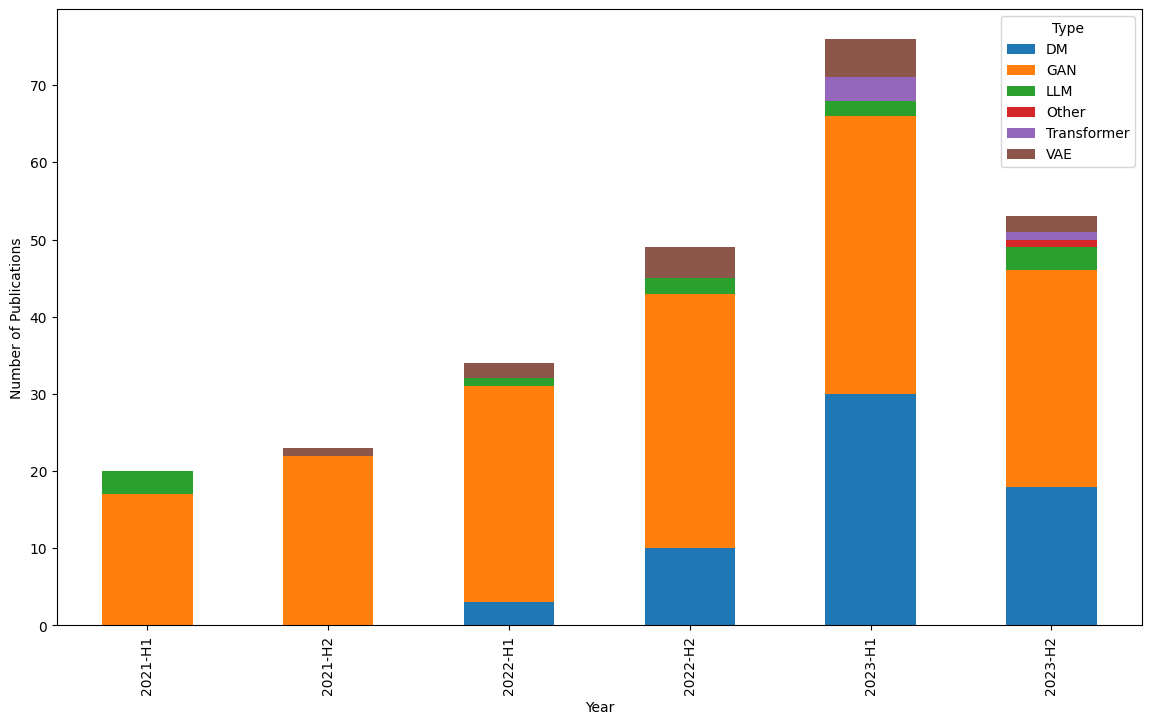}
    \caption{Evolution of used techniques (GANs, DM, VAE, etc.) over years}
    \label{app-years}
\end{figure}
Based on Fig.\ref{app-years}, we notice the following:
\begin{itemize}
    \item \textbf{Dominance of GANs:} GANs consistently dominate the number of publications across all half-year periods. This indicates a strong and sustained interest in GANs within the research community,
    \item \textbf{Rising interest in Diffusion Models (DM):} There is a notable increase in publications related to Diffusion Models (DM), particularly in the latter half of 2022 and 2023. This suggests growing recognition of the potential of DM in generating high-quality synthetic data.
    \item \textbf{Emergence of novel synthetic data generation techniques:} Other model types, including LLMs and Transformers, show periodic spikes in publication numbers. This reflects the expanding scope of this field.
\end{itemize}

\begin{table}[ht]
\centering

\begin{tabular}{p{0.45\columnwidth} p{0.45\columnwidth}}
\hline
\textbf{Privacy Risk} & \textbf{References} \\
\hline
Membership Inference & \cite{s-14,s-41,s-24,s-11,s-12,s-36,ehr-38,s-10,s-18,ehr-69,s-34}, Presence Disclosure  \cite{ehr-15}\\
\hline
Attribute Inference Attack & \cite{s-41,s-11,s-36,s-18,ehr-69} \\
\hline
Nearest Neighbor Adversarial Accuracy Risk (nnaa) & \cite{ehr-58,ehr-59,ehr-26} \\
\hline
Re-identification (or De-anonymization) & \cite{s-18} \\
\hline
Other & Distance to Closest Record  \cite{s-12}, Counting Identical Rows \cite{ehr-52}, Latent Space Nearest Neighbor  \cite{ehr-25}, Re-identification Attack \cite{s-18}, Differential Privacy  \cite{s-10}, KNN Estimation  \cite{ehr-21}\\

\hline
\end{tabular}
\caption{Summary of Privacy Risk Metrics}
\label{tab:privacy_risks}

\end{table}

\subsection{Synthetic Data Evaluation}
As mentioned in section \ref{subsec-evaluation}, synthetic data evaluation involves multiple dimensions. Below, more information on some of these dimensions is provided.

 \subsubsection{Privacy Metrics}Table \ref{tab:privacy_risks} in the appendix displays commonly used metrics and risks in evaluating privacy for EHR data.

 \subsubsection{Fidelity Metrics}
The different fidelity metrics used in the surveyed papers are displayed in Tables \ref{app-imaging} and \ref{app-ehr}. For the imaging data type, only top frequently used metrics are displayed. 

\paragraph{Fidelity Metrics for EHR}

The different metrics used for evaluating EHR are categorized according to the following:

\begin{itemize}
    \item \textbf{Dimension Wise distributional similarity:} This category of metrics focuses on the resemblance of the distribution of individual features between the real and synthetic datasets. \textbf{Dimension-wise probability} is a family of tests that compare the distribution of features of synthetic data to those of real data. Depending on the type of the feature, different tests and metrics are used. \textit{Bernoulli Success probability} \cite{s-14,s-15,s-12,s-10} ,\textit{Pearson Chi-square test} \cite{ehr-52,ehr-12}, and \textit{analysis of variance F-test} \cite{s-34} are commonly used for binary features (which is the case when ICD-9 codes are present), while \textit{Student T-test} \cite{s-34,ehr-12} and \textit{Snedecor’s F-test} \cite{s-34}. is used for continuous features. \textbf{Dimension-wise distribution} is another term used in the literature \cite{s-11, s-24} that focuses on evaluating the resemblance between synthetic and real data at the feature level. Although these publications use the same term, they refer to slightly different aspects and methodologies. Both descriptions use the means to compare the real and synthetic binary features. However, \cite{s-24} extends the evaluation by using the 1-Wasserstein distance for continuous values. The \textit{Kolmogorov–Smirnov Test} \cite{s-18,ehr-50,s-34,ehr-52} is also commonly used to check if the real and synthetic data come from the same distribution. \cite{s-34} use \textit{three sigma rule test}\cite{s-34}  to check if the values of the synthetic data are within realistic ranges of the real data, especially when the KS-test fail. The \textit{Kernel Density Estimation }\cite{s-15,s-34} is used as well to compare the probability density functions of the real and synthetic data. The \textit{Kullback Leibler Divergence} (KLD) \cite{ehr-50,s-34} is a divergence measure that is also used to measure the divergence of the synthetic distribution from the real one. \textbf{Dimension-wise statistics}  \cite{s-41,s-36,ehr-38,s-18} is widely used, where the univariate statistics (mean, standard deviations..) of the synthetic and real features are compared.
    
    \item \textbf{Joint Distribution Similarity}: assesses how well the synthetic data replicates the real data in terms of the overall distribution that encompasses all features simultaneously. It’s about the global structure of the data, ensuring that the synthetic data captures the complex, multi-dimensional relationships as they exist in the real data. In this sense, multiple metrics are used, like the \textit{Jensen-Shannon Divergence} \cite{ehr-12,ehr-22} which is a symmetrical extension of the KLD. \textit{Maximum-Mean Discrepancy} measures the dissimilarity between two probability distributions and is widely used \cite{ehr-16,s-12,ehr-50,s-10}. Measures like the \textit{Wasserstein distance} \cite{ehr-12}, the \textit{Cross-type Conditional Distribution} \cite{s-41}, and\textit{ first-order proximity} \cite{s-41} are also used, each with a different way to measure the joint distributional similarity.  The discriminative score, usually called propensity score as well, measures the overall resemblance of the synthetic data to the real data without explicitly calculating statistical distances. It involves a post-hoc classifier that tries to discriminate the real samples from the synthetic ones. 
    \item \textbf{Inter-dimensional Relationship Similarity}: focuses on the relationships and correlations between pairs or groups of features within the data. It evaluates how well the synthetic data maintains the dependencies and interactions between different features, as observed in the real data. Correlation-based metrics like \textit{Pearson pairwise correlation} \cite{ehr-15,s-10}, \textit{pairwise correlation difference} \cite{s-10,ehr-50,s-24,s-11}, \textit{Kendall’s tau  rank correlation} \cite{s-34} tend to capture if the inter-dimensional correlations of the real-data are maintained in the synthetic data. \textit{Dimension-wise prediction test }is a widely common used method \cite{s-14,s-11,s-12,ehr-16,s-41,s-18}  that is said to catch the quality and inter-dimensional relationships of features of the synthetic data. This metric is calculated by systematically selecting each feature as the target in turn, while utilizing the remaining features to train a classifier.  Such an approach offers a more holistic evaluation compared to the conventional prediction utility metric, which is usually limited to specific downstream tasks. This metric is more comprehensive than the standard one, as it can assess the utility of all features, including those that may not significantly influence the task-specific outcome, thus providing a broader insight into the data's overall quality and utility. \textit{Frequent Association Rules}, as used in \cite{s-41}, are employed to verify whether the association rules identified in the real data remain applicable in the synthetic data. It ensures that the underlying relationships and associations of the real-data are maintained.
    \item \textbf{Latent Distribution similarity:} This family of metrics tend to evaluate if the distribution of synthetic and real data is similar as well in the latent space. The \textit{log-cluster metric} \cite{s-24,s-34,s-11} measures the clustering pattern similarities between the real and synthetic data.  \textit{Latent space representation}\cite{s-36} and latent space evaluation \cite{s-41} metrics also fall into this family.
    \item \textbf{Special Fidelity metrics}: Other metrics that fall under the fidelity category tend to evaluate special features of the synthesis process, like the {diversity} of the synthetic data. Diversity is the ability to cover the variability of the synthetic data. For example, \textit{category coverage} \cite{s-34} and \textit{Generated Disease types}\cite{ehr-22}  are concerned with the presence and representation of all possible categories, usually diseases and medical codes, in the synthetic data. \cite{s-11} focuses on generating uncommon diseases, and as a result, introduced (1) \textit{Medical Concept Abundance} \cite{s-11} that evaluates how diverse and representative these medical codes are in the synthetic data compared to the real data., (2) \textit{Required sample number to generate all diseases (RN)} \cite{s-11} to evaluate the ability to generate uncommon diseases, and (3) \textit{Normalized distance} \cite{s-11} to further evaluate the distribution of uncommon diseases.  Some other metrics that focus on the quality of generating time-series synthetic data, are available, like \textit{autocorrelation} function \cite{s-10} and \textit{patient trajectories} \cite{s-10}. 

\end{itemize}



\begin{table*}[ht]
\centering
    \rotatebox{90}{
\begin{tabular}{p{0.1\textwidth}p{0.2\textwidth}p{0.3\textwidth}p{0.65\textwidth}p{0.05\textwidth}}
\toprule
Modality & Purpose of Evaluation & Fidelity Metric & Usage & Count \\ \midrule
\multirow{21}{*}{Imaging} & \multirow{11}{*}{\parbox{0.18\textwidth}
{Pixel-wise similarity (image based)}} & Structural Similarity Index   (SSIM) & \cite{us-11,us-19,us-3,us-2,us-23,us-22,mamm-12,mamm-7,mamm-8,amirrajab_label-informed_2022,fei_deep_2021,hu_domain-adaptive_2022,huang_common_2022,huo_brain_2022,jiang_cola-diff_2023,la_rosa_mprage_2021,meng_novel_2022,moshe_handling_2023,qin_style_2022,teixeira_adversarial_2021,yoon_sadm_2023,yurt_mustgan_2021,zhan_lr-cgan_2021,zhu_dualmmp-gan_2022,zhu_make--volume_2023,multi-15,ct-01,ct-03,ct-06,ct-10,ct-13,ct-17,ct-18,ct-19,ct-23,ct-26,ct-29,ct-30,ct-33,ct-l-01,ct-l-05,ct-l-12,ct-l-16,ct-32,oct-02,oct-04,oct-12,oct-13,oct-16,x-05,x-12,x-14,x-16,x-17,x-18,x-19,x-24,x-32} & 58 \\
 &  & Peak   Signal-to-Noise Ratio (PSNR) & \cite{us-3,us-23,us-22,mamm-12,mamm-7,mamm-8,amirrajab_label-informed_2022,fan_tr-gan_2022,fei_deep_2021,huang_common_2022,huo_brain_2022,jiang_cola-diff_2023,kim_diffusion_2022,la_rosa_mprage_2021,meng_novel_2022,moshe_handling_2023,qin_style_2022,yoon_sadm_2023,yurt_mustgan_2021,zhan_lr-cgan_2021,zhu_dualmmp-gan_2022,zhu_make--volume_2023,multi-15,ct-06,ct-10,ct-13,ct-16,ct-17,ct-18,ct-19,ct-23,ct-24,ct-26,ct-27,ct-29,ct-30,ct-33,ct-39,ct-l-12,ct-l-16,ct-32,oct-02,oct-04,oct-16,x-12,x-13,x-14,x-16,x-18,x-19,x-32,ct-03} & 52 \\
 &  & Mean Absolute Error   (MAE) & \cite{meng_novel_2022,zhu_make--volume_2023,multi-10,r-229,ct-10,ct-17,ct-23,ct-26,ct-27,ct-29,ct-30,ct-33,ct-39,ct-41,ct-l-05,oct-20,oct-22,x-12} & 18 \\
 &  & Multi-Scale   Structural Similarity Index (MS-SSIM) & \cite{us-6,us-2,us-17,mamm-4,dorjsembe_conditional_2023,dorjsembe_three-dimensional_2022,fan_tr-gan_2022,wolleb_swiss_2022,multi-10,eeg-1,ct-24,oct-14,oct-16,x-15,x-31} & 15 \\
 &  & Mean Squared Error   (MSE) & \cite{us-22,us-21,dorjsembe_conditional_2023,multi-10,r-229,ct-13,ct-18,ct-l-16,oct-12,x-13,x-14,x-19} & 12 \\
 &  & Root Mean Square   Error (RMSE) & \cite{zhu_dualmmp-gan_2022,multi-10,r-98,r-229,r-245,r-251,ct-01,ct-03,ct-10,ct-41,ct-l-05,x-13,ct-20} & 13 \\
 &  & Normalized Mean   Squared Error (NMSE) & \cite{fan_tr-gan_2022,huang_common_2022,huo_brain_2022,kim_diffusion_2022,ct-l-12} & 5 \\
 &  & Normalized Root   Mean Square Error (NRMSE) & \cite{amirrajab_label-informed_2022,fei_deep_2021,la_rosa_mprage_2021,yoon_sadm_2023,zhan_lr-cgan_2021} & 5 \\
 &  & Mean Error (ME) & \cite{ct-27,ct-39,ct-41} & 3 \\
 &  & Universal Quality   Index (UQI) & \cite{ct-32,x-24,x-13} & 3 \\
 &  & contrast noise   ratio  (CNR) & \cite{oct-17,x-04} & 2 \\ \cmidrule(l){2-5} 
 & \multirow{7}{*}{\parbox{0.18\textwidth}
{Feature-wise similarity (distrubution based)}} & Frechet Inception Distance   (FID) & \cite{derm-8,derm-1,derm-5,derm-4,derm-13,derm-7,us-18,us-13,us-11,us-14,us-3,us-2,us-17,us-22,mamm-12,mamm-4,mamm-6,pinaya_brain_2022,teixeira_adversarial_2021,thermos_controllable_2021,wolleb_swiss_2022,multi-4,multi-1,multi-18,multi-16,multi-12,multi-14,multi-100,s-8,s-9,r-229,r-245,s-44,r-235,r-251,eeg-1,ct-01,ct-04,ct-08,ct-12,ct-38,ct-l-01,ct-l-03,ct-l-11,ct-l-12,ct-l-13,ct-l-14,ct-l-15,ct-l-16,ct-l-18,ct-l-02,oct-02,oct-03,oct-04,oct-12,oct-14,oct-15,oct-20,oct-24,x-02,x-04,x-05,x-06,x-08,x-09,x-12,x-13,x-15,x-19,x-22,x-24,x-26,x-31,x-32,ct-l-19} & 75 \\
 &  & Inception Score   (IS) & \cite{derm-11,derm-5,derm-4,us-14,multi-4,multi-1,multi-14,ct-04,ct-l-02,x-22} & 10 \\
 &  & Kernel Inception   Distance (KID) & \cite{derm-8,derm-13,derm-7,us-17,us-22,multi-100,ct-08,oct-12,x-19,x-24} & 10 \\
 &  & Maximum Mean   Discrepancy (MMD) & \cite{us-6,dorjsembe_conditional_2023,qin_style_2022,wolleb_swiss_2022,multi-14,r-256,r-74,r-233,r-229,r-245,r-77,eeg-5,ct-l-13,ct-l-19} & 14 \\
 &  & Learned Perceptual   Image Patch Similarity (LPIPS) & \cite{us-17,us-23,mamm-18,multi-16,ct-08,ct-38,ct-l-12,x-32} & 8 \\
 &  & Feature Similarity   Index (FSIM) & \cite{mamm-12,mamm-7,qin_style_2022} & 3 \\
 &  & feature   distribution similarity (FDS) & \cite{ct-04} & 1 \\ \cmidrule(l){2-5} 
 & \multirow{3}{*}{\parbox{0.18\textwidth}
{Image-text alignment/accuracy}} & Bilingual Evaluation   Understudy (BLEU) & \cite{x-01,x-06} & 2 \\
 &  & Recall-Oriented   Understudy for Gisting Evaluation (ROUGE) & \cite{x-01,x-05} & 2 \\
 &  & Contrastive   Language–Image Pre-training score (CLIP) & \cite{ct-12} & 1 \\ \midrule

\end{tabular}
}
\caption{Fidelity Metrics for Imaging Data}
\label{app-imaging}
\end{table*}

\begin{table}[]
    \rotatebox{90}{

\begin{tabular}{lllll}
\hline
Modality & Purpose of Evaluation & Fidelity Metric & Usage & Count \\ \hline
\multirow{27}{*}{EHR} & \multirow{10}{*}{Dimension   Wise distributional similarity} & Bernoulli success probability & \cite{s-14,s-15,s-12,s-10} & 4 \\
 &  & Pearson Chi-Squared   Test & \cite{ehr-52,ehr-12} & 2 \\
 &  & Analysis of   variance F-test & \cite{s-34} & 1 \\
 &  & Student’s t-test & \cite{s-34,ehr-12} & 2 \\
 &  & Dimension-wise   distribution & \cite{s-24,s-11} & 2 \\
 &  & dimension-wise   statistics & \cite{s-41,s-36,ehr-38,s-18} & 4 \\
 &  & Kolmogorov–Smirnov   Test & \cite{s-18,ehr-50,s-34,ehr-52} & 4 \\
 &  & three sigma rule   test & \cite{s-34} & 1 \\
 &  & Kernel Density   Estimation & \cite{s-15,s-34} & 2 \\
 &  & Kullback Leibler   Divergence & \cite{ehr-50,s-34} & 2 \\ \cline{2-5} 
 & \multirow{6}{*}{inter-dimensional similarity} & Pearson pairwise   correlation & \cite{ehr-15,s-10} & 2 \\
 &  & pairwise correlation difference & \cite{s-10,ehr-50,s-24,s-11} & 4 \\
 &  & Kendall’s $\tau$ rank   correlation & \cite{s-34} & 1 \\
 &  & Dimension-wise   prediction & \cite{s-14,s-11,s-12,ehr-16,s-41,s-18} & 6 \\
 &  & Frequent   Association Rules & \cite{s-41} & 1 \\
 &  & Dimension-wise   correlation & \cite{s-11} & 1 \\ \cline{2-5} 
 & \multirow{5}{*}{Joint distribution similarity} & Jensen-Shannon Divergence & \cite{ehr-12,ehr-22} & 2 \\
 &  & Maximum-Mean   Discrepancy & \cite{ehr-16,s-12,ehr-50,s-10} & 4 \\
 &  & Wasserstein   distance & \cite{ehr-12} & 1 \\
 &  & Cross-type   Conditional Distribution,first-order proximity & \cite{s-41} & 1 \\
 &  & discriminatve score & \cite{s-10,s-18,ehr-58,ehr-52} & 5 \\ \cline{2-5} 
 & \multirow{3}{*}{Latent distribtuion similarity} & Log-cluster & \cite{s-24,s-34,s-11} & 3 \\
 &  & latent space   evaluation metrics & \cite{s-36} & 1 \\
 &  & latent space   representation & \cite{s-41} & 1 \\ \cline{2-5} 
 & specific-language   model & PERPLEXITY & \cite{ehr-38,ehr-69} & 2 \\
 & specific-time-series & autocorrelation function & \cite{s-10} & 1 \\
 & specific-time-series & patient trajectories & \cite{s-10} & 1 \\ \hline
\multirow{2}{*}{Signals} & \multirow{2}{*}{Time} & Dynamic Time Warping (DTW) & \cite{s-8,s-9,r-233,r-245} & 4 \\
 &  & Time Warp Edit   Distance (TWED) & \cite{r-245} & 1 \\ \hline
\multirow{2}{*}{Text} & Similarity between machine   and human translation & Bilingual Evaluation   Understudy (BLEU) & \cite{li2021synthetic,zhou2022datasiftertext,al2021differentially} & 3 \\
 & Similarity between   real and generated text & ROUGE-N recall & \cite{libbi2021generating} & 1 \\ \hline
\end{tabular}

}
\caption{Fidelity Metrics for EHR, Signals, and Text}
\label{app-ehr}

\end{table}

\end{document}